\def\informs{}
\def\operationsresearch{}
\documentclass[mnsc,nonblindrev]{informs3}

\RequirePackage{tgtermes}
\RequirePackage{newtxtext}
\RequirePackage{newtxmath}
\RequirePackage{endnotes}
\usepackage{changepage}

\OneAndAHalfSpacedXI % Current default line spacing
\usepackage{tikz}
\usepackage{algorithm, algpseudocode}

\usepackage{natbib}
 \bibpunct[, ]{(}{)}{,}{a}{}{,}%
 \def\bibfont{\small}%

\newcommand{\zzedit}[1]{\textcolor{violet}{[Note: #1]}}
\newcommand{\srscomment}[1]{\textcolor{red}{[Note: #1]}}
\newcommand{\opre}{}

\newcommand{\myparagraph}[1]{\paragraph{#1}}

\usepackage{amsmath,amsfonts,cancel}
\usepackage{thmtools}
\usepackage{adjustbox}
\usepackage[shortlabels]{enumitem}
\usepackage[table]{xcolor}
\definecolor{edits}{rgb}{0,0,0}
\usepackage{mathtools}
\usepackage{dsfont}
\usepackage{tcolorbox}
\usepackage{booktabs}
\usepackage{bbm}
\usepackage{xspace}
\usepackage{subcaption}
\usepackage{tensor}
\usepackage{bm}
\usepackage{tikz}
\usetikzlibrary{patterns}
\usepackage{natbib}
\usepackage[mathlines]{lineno}

\usepackage{thm-restate}

\usepackage{placeins}
\usepackage[colorlinks=true,breaklinks=true,bookmarks=true,urlcolor=magenta,citecolor=magenta,linkcolor=magenta,bookmarksopen=false,draft=false]{hyperref}
\usepackage[capitalize]{cleveref}
\usepackage{tabularx}
\newenvironment{rproofof}[1]{%
    \ifdefined\opre 
        \begin{proof}{\it Proof of #1.}%
    \else 
        \begin{proof}[Proof of #1]%
    \fi
}{%
    \ifdefined\opre
        \hfill\Halmos\end{proof}  
    \else
        \end{proof}%
    \fi
}

\newenvironment{rproof}{%
    \ifdefined\opre 
        \proof{\it Proof.}%
    \else 
        \begin{proof}%
    \fi
}{%
    \ifdefined\opre
        \hfill\Halmos\endproof  
    \else
        \end{proof}
    \fi
}

\newenvironment{rproofsketchof}[1]{%
    \ifdefined\opre 
        \begin{proof}{\it Proof Sketch of~#1.}%
    \else 
        \begin{proof}[{\it Proof Sketch of~#1}]%
    \fi
}{%
    \ifdefined\opre
        \unskip\nobreak\hfill\Halmos\end{proof}  
    \else
        \unskip\nobreak\hfill\qed\end{proof}%
    \fi
}

\usepackage[dvipsnames]{xcolor}
\usepackage{color-edits}
\addauthor{srs}{orange}
\addauthor{zz}{OliveGreen}
\addauthor{cy}{cyan}
\addauthor{maj}{blue}

\newcommand{\RR}{\mathbb{R}}
\renewcommand{\S}{\mathcal{S}}
\newcommand{\Regret}{\textsf{Regret}}
\newcommand{\A}{\mathcal{A}}
\newcommand{\C}{\mathcal{C}}
\newcommand{\V}{\mathcal{V}}
\renewcommand{\H}{\mathcal{H}}
\newcommand{\E}{\mathcal{E}}
\newcommand{\Ex}{\mathbb{E}}
\newcommand{\M}{\mathcal{M}}

\newcommand{\Lip}{L_{\Theta}}
\newcommand{\ALG}{\textsf IOPEA\xspace}
\newcommand{\diam}{U}
\newcommand{\Thetamax}{\Theta^{\textsf{max}}}
\newcommand{\Inventory}{I}
\newcommand{\Order}{Q}
\newcommand{\F}{\mathcal{F}}
\renewcommand{\Pr}{\mathbb{P}}
\newcommand{\Gbar}{G_\theta}
\newcommand{\Unif}{\textsf{Unif}}
\newcommand{\Gbarprime}{G_{\theta'}}

\newcommand{\CGbarprime}{\tilde{G}_{\theta'|\theta}}
\newcommand{\CGbarprimez}{\tilde{G}_{\theta'|\tilde{\theta}_k(\theta')}}
\newcommand{\CGbarz}{\tilde{G}_{\theta|\tilde{\theta}_k(\theta)}}
\newcommand{\ermthetap}{\hat{\theta}}
\newcommand{\Th}{\kappa}
\newcommand{\deltalocal}{\delta_{\text{local}}}
\newcommand{\Cupperbound}{\bar{C}}

\newcommand{\KL}{\mathrm{D_{KL}}}
\newcommand{\Bern}{\operatorname{Bernoulli}}

\newcommand{\Ind}[1]{\mathbb{I}( #1 )}

\newcommand{\Exp}[1]{\mathbb E \left[ #1 \right]} % Variance

\usepackage{multirow}

\crefname{assumption}{assumption}{assumptions}
\crefname{algocf}{algorithm}{algorithm}

\usepackage{mathtools}

\DeclarePairedDelimiter{\abs}{\lvert}{\rvert}

\DeclareMathOperator*{\tsum}{\sum}

\mathchardef\mhyphen="2D % Define a "math hyphen"
\ifdefined \argmin
\else \DeclareMathOperator*{\argmin}{arg\,min}
\fi
\ifdefined \argmax
\else \DeclareMathOperator*{\argmax}{arg\,max}
\fi
\DeclarePairedDelimiter{\norm}{\lVert}{\rVert}

\let\originalleft\left
\let\originalright\right
\renewcommand{\left}{\mathopen{}\mathclose\bgroup\originalleft}
\renewcommand{\right}{\aftergroup\egroup\originalright}

\renewcommand{\paragraph}[1]{\noindent\textbf{#1}.}

\renewcommand{\paragraph}[1]{\subsubsection*{#1}}

\EquationsNumberedThrough    % Default: (1), (2), ...
\TheoremsNumberedThrough     % Preferred (Theorem 1, Lemma 1, Theorem 2)
\ECRepeatTheorems  % 

\MANUSCRIPTNO{MNSC-0001-2024.00}

\begin{document}
%%%%%%%%%%%%%%%%

% Outcomment only when entries are known. Otherwise leave as is and
%   default values will be used.
%\setcounter{page}{1}
%\VOLUME{00}%
%\NO{0}%
%\MONTH{Xxxxx}% (month or a similar seasonal id)
%\YEAR{0000}% e.g., 2005
%\FIRSTPAGE{000}%
%\LASTPAGE{000}%
%\SHORTYEAR{00}% shortened year (two-digit)
%\ISSUE{0000} %
%\LONGFIRSTPAGE{0001} %
%\DOI{10.1287/xxxx.0000.0000}%

% Author's names for the running heads
% Sample depending on the number of authors;
% \RUNAUTHOR{Jones}
% \RUNAUTHOR{Jones and Wilson}
% \RUNAUTHOR{Jones, Miller, and Wilson}
% \RUNAUTHOR{Jones et al.} % for four or more authors
% Enter authors following the given pattern:
%\RUNAUTHOR{}
\RUNAUTHOR{Zhongjun et al.}

% Title or shortened title suitable for running heads. Sample:
% \RUNTITLE{Predictive Maintenance in Manufacturing}
% Enter the (shortened) title:
\RUNTITLE{RL in Information Ordered MDPs}

% Full title. Sample:
% \TITLE{Optimal Resource Allocation in Humanitarian Logistics: A Stochastic Programming Approach}
% Enter the full title:
\TITLE{Reinforcement Learning in MDPs with Information-Ordered Policies}

% Block of authors and their affiliations starts here:
% NOTE: Authors with same affiliation, if the order of authors allows,
%   should be entered in ONE field, separated by a comma.
%   \EMAIL field can be repeated if more than one author
\ARTICLEAUTHORS{%
%\AUTHOR{John Doe,\textsuperscript{a} Jane Smith,\textsuperscript{b}}
%\AFF{\textsuperscript{a}Department of Industrial Engineering, University of XYZ, \EMAIL{john.doe@xyz.edu; \textsuperscript{b}Department of Computer Science, University of ABC, \EMAIL{jane.smith@abc.edu}} 

\AUTHOR{Zhongjun Zhang}
\AFF{Department of Industrial Engineering and Management Sciences,
Northwestern University, \EMAIL{zhongjun@u.northwestern.edu}}

\AUTHOR{Shipra Agrawal}
\AFF{Department of Industrial Engineering and Operations Research,
Columbia University, \EMAIL{sa3305@columbia.edu}}

\AUTHOR{Ilan Lobel}
\AFF{Technology, Operations, and Statistics,
Stern School of Business, New York University, \EMAIL{il26@stern.nyu.edu}}

\AUTHOR{Sean R. Sinclair}
\AFF{Department of Industrial Engineering and Management Sciences,
Northwestern University, \EMAIL{sean.sinclair@northwestern.edu}}

\AUTHOR{Christina Lee Yu}
\AFF{School of Operations Research and Information Engineering,
Cornell University, 
\EMAIL{cleeyu@cornell.edu}}
} % end of the block

\ABSTRACT{%
% We introduce an epoch-based reinforcement learning algorithm for infinite-horizon average-cost MDPs \majedit{with continuous state and action spaces} that exploits an \majedit{information order} over a policy class. In this structure, $\pi_{\theta’} \preceq \pi_\theta$ if data collected under $\pi_\theta$ can be used to estimate the performance of $\pi_{\theta’}$, enabling counterfactual inference without additional environment interaction. Leveraging this information order, \majedit{we show that our algorithm achieves a regret bound of $\tilde{O}(\sqrt{wdT})$, where $w$ is the width of the information order and $d$ is the dimension of the policy class.}
% We illustrate the applicability of these information orders in many domains in operations research, including inventory control and queuing systems. For each, we apply our framework to that problem, yielding new theoretical guarantees and strong empirical results without imposing extra assumptions such as convexity in the inventory model or specialized arrival‐rate structure in the queuing model.
%
Many operational Markov decision processes provide structured information where a trajectory generated under one policy may also reveal how other policies would have performed. We formalize this through an \emph{information order} over a parameterized policy class, writing $\pi_{\theta'} \preceq \pi_\theta$ when data collected under $\pi_\theta$ can be used to estimate the long-run average cost of $\pi_{\theta'}$. We develop an epoch-based reinforcement learning algorithm for infinite-horizon average-cost MDPs with continuous state and action spaces that exploits this order through counterfactual inference. For a $d$-dimensional policy class whose information order has width $w$, the algorithm achieves $\tilde O(H\sqrt{wdT/\alpha})$ regret relative to the optimal policy in the class, where $H$ is a bias span upper bound and $\alpha$ denotes an ``information ratio''. Thus, the width captures the effective exploration complexity of the problem, interpolating between full counterfactual feedback and unstructured policy learning. We instantiate the framework in lost-sales inventory systems with positive lead times, dual-sourcing systems governed by dual-index policies, and queues with state-dependent service rates. These applications yield new regret guarantees without requiring global convexity of the average-cost function or specialized arrival-rate structure. Numerical experiments show that exploiting the information order substantially improves learning performance over existing general-purpose and problem-specific benchmarks.

}%

% \FUNDING{}

%Supplemental Material:
%Data Ethics & Reproducibility Note:

% Sample
%\KEYWORDS{Stochastic programming, Decision support,Uncertainty, Disaster response, Optimization}

% Fill in data. If unknown, outcomment the field
%\HISTORY{Received: Month DD, YYYY; Accepted: Month DD, YYYY; Published Online: Month DD, YYYY}

\maketitle
%%%%%%%%%%%%%%%%%%%%%%%%%%%%%%%%%%%%%%%%%%%%%%%%%%%%%%%%%%%%%%%%%%%%%%

% Text of your paper here

\section{Introduction}
\label{sec:introduction}

Recent breakthroughs in reinforcement learning (RL) have demonstrated superhuman-level performance in complex board and video games, such as Go \citep{silver2016mastering,silver2017mastering}, Atari \citep{mnih2015human}, and StarCraft \citep{vinyals2017starcraft,alphastarblog}.
While these advances open exciting new avenues for controlling complex and unknown systems, their success relied on access to millions or even billions of data points \citep{vinyals2017starcraft, alphastarblog}, collected via interactions with high-fidelity simulators such as the StarCraft game engine. However, despite this progress, the deployment of RL in real-world operational systems remains nascent. While there have been promising developments in inventory management~\citep{madeka2022deep,alvo2023neural,eisenach2024neural}, queuing systems~\citep{anselmi2022reinforcement}, supply chain optimization~\citep{chong2022optimization, rolf2023review}, and algorithmic pricing~\citep{bertsimas2017data, zhang2022data}, wide-scale adoption has been hindered by persistent challenges related to the required sample complexity.
One critical reason is that many operational systems are inherently non-tabular, with {\em continuous} state and action spaces~\citep{madeka2022deep,alvo2023deep,besbes2009dynamic,han2025optimal}.
At the same time, these systems often provide structured side information where data collected under one policy may also reveal how other policies would have performed. Our central idea is to formalize and exploit this structure to improve learning efficiency.

\myparagraph{Motivation from Inventory Control.} As a concrete example, let us consider the classical perishable newsvendor model from supply chain management~\citep{goldberg2021survey}. Consider operations in a retail store where the goal is to manage inventory (products in the store) by making ordering decisions subject to stochastic demand.  The seller decides on an ordering quantity $q$, faces i.i.d. demand $D$, and incurs a (random) cost $C(q,D) = h(q-D)^+ + p(D-q)^+$, where $(\cdot)^+ = \max\{\cdot, 0\}$ and $h$ and $p$ are the holding and lost-sales cost coefficients respectively.
If demand $D$ is fully observed, then a single demand realization can be used to calculate $C(q,D)$ for every ordering quantity $q$. This corresponds to the {\em full-feedback} setting, and existing work leverages this structure to obtain improved sample-complexity guarantees~\citep{levi2007provably}. In practice, however, the seller typically observes only sales $N(q,D) = \min\{q,D\}$ rather than the underlying demand~\citep{besbes2013implications,hssaine2024data}.
Our insight is that the lost-sales feedback induces a natural {\em order} over ordering levels $q$, since observed sales $N(q,D)$ can be deterministically used to infer $N(q',D)$ for any $q' \leq q$. This structure induces an ordering $\pi_{q'} \preceq \pi_q$ over policies (characterized by their ordering level) with {\em width} one. 
We show below that the same principle extends to substantially more complex MDPs, including systems with positive lead times.

This example suggests an alternative approach to learning in non-tabular MDPs. In tabular settings, $\sqrt{T}$ scaling regret bounds are attainable relative to the globally optimal policy, but they also depend polynomially on the sizes of the state and action spaces~\citep{auer2008near}. For MDPs with continuous state and action spaces, however, Lipschitzness combined with discretization generally yields regret bounds whose dependence on $T$ is slower and worsens with the dimension~\citep{sinclair2023adaptive,kar2024provably}. Recovering $\sqrt{T}$ dependence typically requires stronger global assumptions, such as linearity~\citep{wang2019optimism}.  Such assumptions may not hold in operational settings, and adopting them as approximations can introduce errors that are difficult to characterize, particularly when the globally optimal policy is itself poorly understood.

Rather than imposing global structure assumptions on value functions or transition dynamics~\citep{wang2019optimism}, we focus on well-studied, problem-specific policy classes from the operations research literature. These classes are often motivated by practice and supported by separate guarantees such as asymptotic optimality. Although they need not satisfy conventional assumptions such as linearity or convexity, they may exhibit structured side information analogous to that in the newsvendor example where trajectories generated by some policies can be used to infer how other policies would have performed.  The idea of exploiting such side information, commonly referred to as ``rich feedback'', has previously been studied~\citep{dann2020reinforcement,gong2024bandits}. Existing approaches, however, do not directly capture the policy-level information structures arising in the non-tabular operational models considered here. This motivates a unified framework for describing rich feedback directly over policies in general MDPs. Beyond characterizing which policies provide information about others, we seek a structural measure that quantifies how much independent exploration is required for learning. We therefore ask:

\smallskip
\begin{adjustwidth}{4em}{4em}
{\em \begin{itemize}
    \item How can we formalize rich feedback over policies in general MDPs?
    \item What property of this information structure determines the difficulty of learning? 
    \item Can it be leveraged to improve sample efficiency? 
    \item What real-world problems admit favorable information structures?
\end{itemize}}
\end{adjustwidth}

\subsection{Contributions}

We consider infinite-horizon average-cost MDPs and a class of parameterized stationary policies $\Pi_\Theta = \{\pi_\theta : \theta \in \Theta\}$. Our objective is to learn the policy in this class that minimizes the long-run average cost. Within this setting, we answer the preceding questions affirmatively by introducing an \emph{information order} over policies grounded in $\sigma$-algebra measurability. The primary contributions of this paper are as follows.

\myparagraph{Information Orders and Width as a Learning-Complexity Measure.}
Our main conceptual contribution is to identify and formalize a new type of information structure over policy classes in general MDPs. For two policies $\pi_{\theta'}$ and $\pi_\theta$, we write $\pi_{\theta'} \preceq \pi_\theta$ if data generated by $\pi_\theta$ can be used to construct an estimator of the long-run average cost under $\pi_{\theta'}$. Thus, the order describes which policies can be counterfactually evaluated from trajectories generated by other policies.
We formalize two versions of this idea. First, the {\em Sample‐Path Information Order} (\cref{definition_sample_path_order}) holds when the {\em empirical average cost} (\cref{definition_el}) under \(\pi_{\theta'}\) can be computed directly from a trajectory generated by \(\pi_{\theta}\), which can occur when the counterfactual trajectory under \(\pi_{\theta'}\) is fully recoverable. Second, the {\em Distributional Information Order} (\cref{definition_general_order}), which is a generalization of the Sample-Path Information Order, applies when the empirical average cost under \(\pi_{\theta'}\) is not directly obtainable, but there exists a random variable measurable with respect to the \(\pi_{\theta}\) trajectory, whose distribution lies within a small total variation distance of the empirical average cost under \(\pi_{\theta'}\). 

Beyond describing which policies inform others, we introduce the {\em width} $w$ of an information order as a measure of the learning complexity. Formally, $w$ is the smallest integer such that, for every subset $\Theta'\subset\Theta$, implementing at most $w$ selected policies in $\Theta'$ is sufficient to counterfactually estimate the performance of every policy in $\Theta'$. Thus, width captures the amount of {\em exploration} required by the information structure; $w=1$ corresponds to a ``full-feedback'' regime in which one sufficiently informative policy can evaluate the entire candidate set, whereas under the trivial order every policy must be explored independently.

\myparagraph{Exploiting Low-Width Information Orders with $\tilde{O}(H\sqrt{wdT/\alpha})$ Regret.}The definition of width $w$ immediately suggests a natural learning strategy.  At each stage, rather than implementing every active policy (set of estimated near-optimal policies), it suffices to explore a collection of at most $w$ maximal policies under the information order; trajectories generated by these policies can then be used to construct counterfactual estimates for all remaining active policies. Combining this structural observation with classical action-elimination algorithms leads to our Information-Ordered Epoch-based Policy Elimination Algorithm (\ALG; \Cref{alg_policy_elimination}).
Because we do not impose global assumptions such as convexity or linearity on the long-run average-cost function, we first discretize the compact policy class $\Theta$ using a finite cover $\Theta_1$. The algorithm then searches for the best policy in $\Theta_1$, with the discretization radius chosen to balance the discretization and learning errors.  In each epoch, \ALG implements only the maximal policies in the current active set, uses their trajectories to evaluate all active policies through the information order, and eliminates policies certified to be suboptimal.
Although \ALG builds on classical action elimination, the information order fundamentally changes how exploration is allocated. Since all policies in the active set are treated equally from the perspective of regret control (see~\Cref{lem_a.5}), exploration naturally focuses on the maximal policies, which are precisely the most informative under the information order. This is reminiscent of information-directed sampling~\citep{russo2018learning}, but here the relevant information structure is identified directly by the information order. 
Notably, this exploration mechanism differs from many existing rich-feedback algorithms that passively exploit available side observations without explicitly using the information structure to guide exploration decisions~\citep{dann2020reinforcement,lykouris2020feedback}; see~\Cref{discussion_feedback_graph}.

Our main result (\Cref{thm:regret_bound}) shows that \ALG achieves regret $\tilde{O}(H\sqrt{wdT/\alpha})$, where $w$ is the information order width, $d$ is the dimension of the policy class, and $H$ is the standard bias-span bound for infinite-horizon MDPs. $\alpha$ denotes an ``information ratio'' that captures the potential information loss when the information order is used for counterfactual estimation; see~\Cref{definition_general_order}.
In contrast, discretizing a policy class that admits only the trivial information order, under which each implemented policy provides information only about itself and hence $w = |\Theta_1|$, yields regret scaling as $T^{\frac{d + 1}{d + 2}}$. The width then interpolates between the no-feedback regime ($w = |\Theta_1|$), where every policy must be explored separately, and the full-feedback regime ($w = 1$), where a single policy provides information about the entire class. 

We also establish a nearly matching lower bound of $\Omega(\sqrt{HwdT/\alpha})$ in~\Cref{sec_lower_bound}, showing that the dependence on the width $w$ is fundamental. The remaining $\sqrt{H}$ gap is standard and can be closed through a variance-aware analysis~\citep{boone2024achieving}.
We further note that these guarantees require only the information order and mild regularity conditions on the MDP. In particular, they do not rely on global properties of the long-run average-cost function such as convexity~\citep{agrawal2019learning} or on linear function approximation~\citep{jin2020provably}. 
Finally, in \cref{propo_complexity_1} we consider the computational complexity of \ALG.  This result highlights a trade-off between {\em statistical} and {\em computational} complexity, where leveraging the information order achieves {\em exponentially} better regret, with {\em mildly} worse computational complexity due to maintaining a finer discretization of the policy class $\Theta$.

\myparagraph{Application in Real-World MDPs.}
In \Cref{sec_case_studies}, we demonstrate the versatility of information orders and \ALG across several canonical problems in operations research. These include inventory control with positive lead times and lost sales~\citep{goldberg2021survey}, as well as extensions incorporating product returns~\citep{chen2025managing} and multiple products; lost-sales dual-sourcing systems governed by dual-index policies~\citep{whittemore1977optimal}; and queues with state-dependent service rates~\citep{anselmi2022reinforcement}. In each case, we identify natural policy classes that admit a low-width information order, yielding new regret bounds.

In particular, the dual index policy case study highlights that our framework captures structural properties of MDPs that go beyond those traditionally explored in the literature, such as convexity \citep{agrawal2019learning,gong2024bandits, chen2024tailored} and feedback graph connectivity \citep{dann2020reinforcement}. The lost-sales dual index policy setting lacks convexity~\citep{veeraraghavan2008now}, and applying existing learning algorithms yields regret scaling as $T^{2/3}$, whereas our approach yields a regret upper bound that scales as $\sqrt{T}$. We also obtain a $\sqrt{T}$ scaling regret upper bound for the queuing problem studied in \Cref{case_study:queuing}, which is independent of the action space size without any additional structural assumptions on the system dynamics. In contrast, the \(\tilde{O}(\sqrt{|\A|T})\) regret bound in \citet{anselmi2022reinforcement}, although independent of the state space size, scales with the size of the action space $|\A|$ and depends critically on assuming a state‐dependent decay in the arrival rate to limit visits to high‐occupancy states. Moreover, our approach extends naturally to more complex scenarios, such as state‐dependent arrival rates in queuing systems.

We also perform numerical simulations on all three case studies in \Cref{sec:simulations}, and show that~\ALG consistently matches or exceeds the performance of specialized, problem-specific baselines. These results confirm the practical effectiveness and scalability of our information‐order‐based approach.

\subsection{Literature Review}

\myparagraph{Reinforcement Learning for Tabular and Non-Tabular MDPs.}
Reinforcement learning algorithms for tabular MDPs have been extensively studied in both model-based~\citep{auer2008near, osband2016generalization} and model-free settings~\citep{strehl2006pac, jin2018q,  wei2020model}. 
In the sequential interaction setting with finite horizon \(H\) and without access to a generative model, the best-known regret scales as \(\tilde{O}(H^{1/2}\sqrt{|\S||\A|T})\) in the model-based setting~\citep{azar2017minimax} and \(\tilde{O}(H\sqrt{|\S||\A|T})\) in the model-free setting~\citep{zhang2020almost}. Similar bounds hold in infinite-horizon average-cost settings, with \(H\) replaced by related problem primitives such as the diameter, mixing time, or bias span~\citep{auer2008near,boone2024achieving}. These bounds become prohibitive as the state and action spaces grow, motivating a growing body of research on MDPs with large or continuous state and action spaces that exploits problem-specific structure to improve learning efficiency. Examples include linear function approximation~\citep{wang2019optimism,jin2020provably}, kernel approximation~\citep{chowdhury2019online}, Lipschitz continuity~\citep{sinclair2023adaptive,kar2024provably}, eluder dimension~\citep{osband2014model,mou2020sample}, Bellman rank~\citep{dong2020root}, and convexity~\citep{agrawal2019learning}.
Specifically, unlike stronger assumptions listed above that permit $\sqrt{T}$ scaling regret bounds, Lipschitzness alone cannot yield a regret bound with a better dependence on $T$ than $T^{2/3}$~\citep{sinclair2023adaptive,kar2024provably}.
Alongside these works, we explore a new type of structural property for MDPs with continuous state and action spaces, i.e., information orders over the policy class.

\myparagraph{MDPs with Rich Feedback.}
The ordering structure we investigate corresponds to the broader notion of rich feedback in both bandit and RL problems. In the bandit literature, rich feedback has been extensively analyzed~\citep{mannor2011bandits, alon2015online, wen2024stochastic}, with foundational results on feedback graphs and side‐observations. In contrast, for RL and MDPs, the study of rich feedback remains relatively nascent. \citet{dann2020reinforcement} analyze the regret of a confidence set elimination algorithm under general feedback graphs, and show that in tabular MDPs, certain structural properties of the graph can be used to obtain regret bounds that do not depend on $|\S|$ and $|\A|$. While the setting studied in \citet{dann2020reinforcement} is similar to ours, their method yields a $T^{2/3}$ scaling regret upper bound in several important operations research domains due to the tabular assumption, including the lost-sales inventory control and the dual sourcing problem we study (see \Cref{discussion_feedback_graph} for details). In contrast, our approach directly applies to continuous state-action MDPs and achieves a tighter $\sqrt{T}$ scaling regret bound in these settings. Furthermore, \citet{wan2024exploiting} analyze two extreme feedback regimes in exogenous MDPs—no feedback and full feedback (their “no observation” and “full observation”), and point out a \(\sqrt{d}\) gap in the regret of these two cases, where \(d\) is the size of the exogenous state space.  Our work differs from \citet{wan2024exploiting} in two key respects. First, we handle general MDPs rather than only exogenous ones, thus their analysis, which is based on linear mixture MDPs, would fail; Second, we introduce an information order framework that interpolates between these extremes, yielding greater generality than either no‐feedback or full‐feedback analyses.

\myparagraph{MDPs with One-Sided Information.}
Our concept of information order is motivated by (and indeed, builds upon) one-sided information, which represents a special case of rich feedback. 
%The notion of one-sided information has also been studied in the literature across various contexts.
\citet{zhao2019stochastic} and \citet{yuan2021marrying} investigate one-sided information in the context of bandit problems. Similarly, \citet{gong2020provably, gong2024bandits} study a general Q-learning algorithm under both full-feedback and one-sided-feedback settings, establishing a learning framework that achieves a $\sqrt{T}$ scaling regret bound independent of the state and action spaces. They later instantiate this framework in a zero lead-time inventory problem with cyclic demand. The feedback structure in \citet{gong2020provably, gong2024bandits} is defined at the action-space level, and does not apply to inventory control models with positive lead times.  We extend this idea to define information order over {\em policies} instead of {\em actions}, and show how our results apply to a larger class of inventory models.
% \zzdelete{Similarly, \citet{gong2024bandits, gong2020provably} study an information order in inventory control with cyclic demand. However, a key difference is that \citep{gong2024bandits} focus exclusively on inventory settings with zero lead time, thus implicitly assumes convexity. In contrast, the inventory model we consider extends naturally to include positive lead times.}

\myparagraph{Algorithms Leveraging Problem‐Specific Structure.}
From an application perspective, several prior works have investigated leveraging problem-specific structure for real-world MDPs. A closely related work is \citet{agrawal2019learning}, which establishes a $\sqrt{T}$ scaling regret bound for the same lost-sales inventory problem we consider in~\cref{case_study:inventory} by exploiting convexity. However, this convexity‐based approach cannot be extended to the dual‐sourcing problem in \cref{case_study:dual_index_policy}, since dual index policies lack the requisite convexity property~\citep{veeraraghavan2008now}. In contrast, our policy‐ordering framework applies naturally to both problems. \citet{chen2024tailored} derive a $\sqrt{T}$ scaling regret bound for tailored base‐surge policies in the dual‐sourcing setting by relying on convexity over tailored base‐surge policies, an assumption that fails for dual index policies and thus prevents a direct extension of their method. \citet{tang2024online} analyze the dual index policy under the backlog formulation, which is a simpler setting {with full feedback} compared to the lost-sales setting considered in this paper.  
%Lastly, we emphasize that our results do not rely on any {\em strong} assumptions on the underlying distributions and cost functions.

%Lastly, our results do not rely on any assumptions on the underlying distributions.   %distributional modeling, since you just evaluate on the sampled trajectory itself

%\srscomment{Confused by this - since you just made the point that this set-up is nonconvex?}\zzcomment{Two different policies. We worked on dual index policy and theirs is base-surge policy (basically fix the long-lead replenishment at each period and use the short-lead channel to deal with the stochasticity).}  Additionally, \citet{tang2024online} analyze the dual index policy under a backlog formulation, which is a simpler setting compared to the lost-sales setting considered in this paper.

Note that for real‐world MDPs with finite state and action spaces, standard RL algorithms apply directly~\citep{strehl2006pac, auer2008near,osband2016generalization,jin2018q,wei2020model, khodadadian2021finite, chen2021lyapunov}. \citet{anselmi2022reinforcement} study the service‐rate control queuing problem in \Cref{case_study:queuing}, using a model‐based estimator that exploits problem‐specific structure to obtain a $\sqrt{T}$ scaling regret bound that is independent of the state space size but includes terms that scale linearly with the action-space size $|\A|$, assuming a state‐dependent decay in the arrival rate. In contrast, our information‐ordering method achieves a $\sqrt{T}$ scaling regret bound independent of the action space size, without explicit estimation of transition dynamics or rewards and without requiring any arrival‐rate decay assumptions.

Beyond inventory and queuing models, numerous works exploit domain-specific structure for improved online learning algorithms in various operations research applications, including assortment optimization~\citep{cao2019dynamic}, first-price auction~\citep{han2025optimal}, joint pricing and inventory control~\citep{chen2022dynamic}, dynamic pricing with product returns~\citep{chen2025managing}, reward programs in revenue management~\citep{hssaine2025learning}, and pricing for service systems with reusable resources~\citep{jia2024online}.
This growing body of work complements our focus on leveraging information structures in real-world MDPs for improved guarantees.

% Beyond the above works, recent studies in operations research have examined leveraging domain-specific structure for improved algorithms. 
% For instance, \citet{cao2019dynamic} study an assortment learning problem under a multinomial logit model, where a tiered product structure is introduced to accommodate frequent new product launches. 
% \citet{chen2022dynamic} investigate a joint pricing and inventory control problem in which pricing can be adjusted more frequently than inventory replenishments. 
% \citet{chen2025managing} consider dynamic pricing with product returns, where the return mechanism introduces an additional operational structure. 
% \citet{jia2024online} study online learning and pricing for service systems with reusable resources, where resource reusability creates a distinctive operational constraint. 
% These studies reflect a growing interest in online learning driven by domain-specific business structures, complementing our focus on informational structures in real-world MDPs.

\myparagraph{Empirical RL Applications.}  
We note that many empirical algorithms have been applied to a variety of real‐world systems without explicitly exploiting their MDP structure. For example, \citet{feng2021scalable} use deep learning methods to optimize ride‐sharing dispatch, \citet{dai2021queueing} apply policy‐gradient methods to dynamic queuing networks, and \citet{fang2019reinforcement} explore empirical approaches for jitter‐buffer management in streaming. While these methods have shown empirical success in many applications, they typically incur high sample complexity because they do not exploit the underlying problem structure (see comparison with \textsf{PPO} in \Cref{sec:simulations}).  By incorporating information ordering, our method offers the potential to improve sample efficiency and overall performance in these domains.

\myparagraph{Paper Organization.}
In \Cref{sec:preliminary}, we introduce the necessary notation and formally define the MDP problem, along with the information ordering over the policy class that underpins our framework. \Cref{sec_main_result} presents our main theoretical contributions: we derive upper and lower regret bounds under information orders over policies, and show that these bounds are nearly tight. In \Cref{sec_case_studies}, we demonstrate the applicability of our framework through three case studies: single-retailer inventory control with positive lead time in \Cref{case_study:inventory}, {the dual index policy for the lost-sales dual sourcing problem in \Cref{case_study:dual_index_policy}}, and an M/M/1/L queue with service-rate control in \Cref{case_study:queuing}. In each case, we compare the theoretical performance of our algorithm against existing approaches (when available). \Cref{sec:simulations} presents detailed numerical simulations that validate our theoretical guarantees and empirically examine the impact of model parameters in each case study. Finally, \Cref{sec_conclusion} concludes with a discussion of potential extensions, and all technical proofs and supplementary details are deferred to the Appendix.

\section{Preliminary}\label{sec:preliminary}

% \srscomment{Think about putting polciy order before the assumptions (main idea, so you want to lead with that).}

% \srscomment{Maybe use different notation for the distributional policy order?}

% \srscomment{I also thought you should emphasize somewhere the fact that your approach is more robust to distributional assumptions? I feel like you don't really highlight that anywhere but the great thing about these information orders is that you don't need to do any distributional modeling, since you just evaluate on the sampled trajectory itself (I guess you do need just minimally something that guarantees mixing, and for really generic distributions you might not get mixing?) but I imagined that previous methods might rely on more distributional assumptions? }

% \srscomment{For spacing can change the myparagraph to noindent and then just bold the header. myparagraph adds a little bit of white space between the myparagraphs.}

\myparagraph{Technical Notation.} 
We use \( [T] = \{1, \ldots, T\} \). For random variables \( X \) and $Y$, we write \( X \in Y \) to indicate that \( X \) is measurable with respect to the $\sigma$-algebra generated by $Y$. When \( X \in Y \), we sometimes write \( X(Y) \) to emphasize this measurability. 
The total variation distance between \( X \) and \( Y \), defined as the
total variation distance between their pushforward measures (i.e., their
induced distributions), is denoted by \( d_{\mathrm{TV}}(X, Y) \).
Unless otherwise specified, all norms over the policy class \( \Theta \) refer to the \( \ell_\infty \) norm. We use $O(\cdot)$ and $\Omega(\cdot)$ to denote rates that omit absolute constants and, when summarizing leading-order rates, use $\tilde O(\cdot)$ and $\tilde\Omega(\cdot)$ to additionally omit problem-dependent polynomial factors not displayed in their arguments and polylogarithmic factors in $T$.

\myparagraph{MDP and Policies.} We consider an agent interacting with an underlying infinite-horizon average-cost Markov Decision Process (MDP) over $T$ sequential periods.  The underlying MDP is given by a tuple $(\S, \A, P, C, s_1)$, where $\S$ denotes the state space, $\A$ the action space, $P : \S \times \A \rightarrow \Delta(\S)$ the state transition kernel, $C: \S \times \A \to \Delta([-\Cupperbound, \Cupperbound])$ the stochastic cost function, and $s_1$ the initial state~\citep{puterman2014markov}.
\(\Cupperbound \in \mathbb{R}\) denotes a constant upper bound on the absolute value of the one-step cost. We allow costs to be negative in order to accommodate several operations research case studies; see \Cref{case_study:inventory,case_study:dual_index_policy}. We assume that the agent has access to a class of stationary deterministic policies \(\pi_\theta:\S\to\A\) parameterized by \(\theta\in\Theta\).
We use $\theta$ and $\pi_\theta$ interchangeably to denote both the parameter and the policy. 
We note that the state and action spaces, as well as the policy class $\Theta$, could be continuous.

The system evolves as follows. The initial state is given by \( s_1 \). At each time step \( t \), the agent observes the current state \( S_t \in \mathcal{S} \), selects a policy parameter \( \theta_t \in \Theta \), and applies the corresponding action \( A_t = \pi_{\theta_t}(S_t) \). The agent then observes a realized cost \( C(S_t, A_t) \), and the state transitions to \( S_{t+1} \) according to the transition kernel \( P(\cdot \mid S_t, A_t) \).

\myparagraph{Loss and Bias.} For any fixed policy $\theta$, the {\em long-run average cost} \( g_\theta(s) \) is the cumulative average cost starting from state $s$. The {\em bias} $v_\theta(s)$ is the total difference in the average cost from the asymptotic average cost starting from state $s$.  More formally we have~\citep{puterman2014markov}:
\begin{align*}
    g_\theta(s) = \lim_{T \rightarrow \infty} \frac{1}{T}\mathbb{E} \Bigl[ \sum_{t \leq T} C(S_t, A_t) ~\Big|~ S_1 = s\Bigr], \quad
    v_\theta(s) =  \lim_{T \rightarrow \infty}\mathbb{E} \Bigl[ \sum_{t \leq T} \left(C(S_t, A_t) - g_\theta(S_t)\right) ~\Big|~ S_1 = s \Bigr].
\end{align*}
More technical details, such as the existence of the limits, are deferred to~\Cref{section_modelling_assumptions}.
We denote by $\theta^* = \argmin_{\theta \in \Theta} g_\theta(s_1)$ the optimal parameter within the policy class $\Theta$, and by $\pi_{\theta^*}$ the corresponding optimal policy within the class.

\myparagraph{Online Learning Structure.} We assume that over a series of rounds $t \in [T]$ the agent needs to make sequential decisions using one of the policies $\pi_{\theta_t}$ for $\theta_t \in \Theta$.  Their goal is to minimize the total cost, $\sum_{t} C(S_t, A_t)$ starting from $S_1 = s_1$ and following their decisions $A_t \sim \pi_{\theta_t}(S_t)$.  We benchmark the agent on their {\em regret}: the additive loss over all periods the agent experiences using their policy instead of the optimal in-class one.  In particular, the (expected) regret is defined as:
\begin{align}
\Regret(T) = \Ex\left[\sum_{t = 1}^T C(S_t, A_t) - T g_{\theta^*}(s_1)\right],
\end{align}
where $A_t = \pi_{\theta_t}(S_t)$ and $S_{t+1} \sim P(\cdot \mid S_t, A_t)$.
We consider in-class policy learning, since in many practical reinforcement learning and operations research settings, the policy class $\Theta$ is chosen to represent a family of structured or heuristic decision rules that are computationally and operationally tractable~\citep{karlin1958inventory, veeraraghavan2008now, huh2009asymptotic, xin2023dual} (including those examined in \cref{sec_case_studies}). The quality of our regret metric therefore depends on the expressiveness of $\Theta$, i.e., how close the in-class optimal policy $\pi_{\theta^*}$ is to the global optimal.  We discuss this in more detail in \cref{section_modelling_assumptions}.

\subsection{Information Orders over Policies}\label{section_policy_ordering}

In many problem domains, data collected from executing one policy can be used to infer the performance of others.  In the newsvendor example above, parameterizing policies according to their ordering level $q$ yields that samples collected under policy $\pi_q$ can be used to estimate the performance of $\pi_{q'}$ for $q' \leq q$.  We introduce an information order over policies capturing this phenomenon represented via $\pi_{q'} \preceq \pi_q$.  Before defining this we start with some necessary technical notation.

\begin{definition}\label{definition_el}
    For any policy $\theta \in \Theta$ and $T \in \mathbb{N}^+$, we denote $\mathcal{H}_\theta^T$ to refer to a random sample path of the form $ \{(S_t, A_t, C(S_t, A_t), S_{t+1})\}_{t \in [T]},$ collected under policy $\pi_\theta$ where the starting state $S_1 = s_1$, $A_t = \pi_\theta(S_t)$, and $S_{t+1} \sim P(\cdot \mid S_t, A_t)$.
    We define $\Gbar(\H_\theta^T)$ as the empirical average cost over a length $T$ trajectory $\H_\theta^T$ starting from $s_1$:
    \begin{align}
    \Gbar(\H_\theta^T) = \frac{1}{T} \tsum_{t=1}^{T} C(S_t, A_t), \quad \text{ where } S_1 = s_1. \label{eq:erm}
    \end{align}
\end{definition}
The dependence of $\Gbar(\H_\theta^T)$ on \( s_1 \) is implicit and omitted for brevity. Note that the random variable $\Gbar(\H_\theta^T)$ is measurable with respect to $\mathcal{H}_\theta^T$, indicating that by sampling a trajectory from $\theta$ one can evaluate $\Gbar(\H_\theta^T)$ to estimate its long-run average cost $g_\theta(s_1)$. 
For example, in the newsvendor example from the introduction, $\theta$ corresponds to the order quantity $q$ and a sample path $\H_\theta^T$ corresponds to $\{(\cdot, q, C(q, D_t))\}_{t \in [T]}$, since there is no inventory carry-over across periods.  (We note that later in \cref{case_study:inventory} we consider a stateful setting with inventory carry-over and positive lead times).

\ifdefined\informs \subsubsection{Sample-Path Information Ordering}\label{sec_policy_ordering_2}
\else \myparagraph{Sample-Path Policy Ordering} \fi
The most straightforward way to perform counterfactual estimation is when \( \Gbarprime(\mathcal{H}_{\theta'}^{T}) \in \mathcal{H}_\theta^T \), i.e., the empirical average cost under $\theta'$ is estimatable using data collected from $\theta$. This is exemplified in the newsvendor discussion above.
% In this case, setting \( \CGbarprime = \Gbarprime(\mathcal{H}_{\theta'}^{T}) \) trivially ensures measurability, and \( \CGbarprime \) approximates \( \Gbarprime(\mathcal{H}_{\theta'}^{T}) \) exactly, as they share the same distribution. 
While this condition may appear restrictive, we show that it holds in some cases, like the inventory control problem with positive lead time and lost-sales in~\cref{case_study:inventory}. The corresponding information order (referred to as the {\em sample-path} information order) is defined as follows:

%We start off with the most straightforward but also restrictive case of policy ordering case, where \( \Gbarprime(\mathcal{H}_{\theta'}^{T}) \in \mathcal{H}_\theta^T \).

%a simple example of our information structures, where we set $\CGbarprime = G_{\theta'}(T, \mathcal{H}_{\theta'}^{T}$.  Comment on how this is the most straightforward (but also restrictive), because we are assuming that the full data is useful to estimate $\theta'$ from $\theta$.  i.e. $T' = T$, and the full distribution is satisfied.

%Based on conditions (i) and (ii) above, if \( \Gbarprime(T', \mathcal{H}_{\theta'}^{T'}) \in \mathcal{H}_\theta^T \) for some \( T' \in \mathbb{N}^+ \), then setting \( \CGbarprime = \Gbarprime(T', \mathcal{H}_{\theta'}^{T'}) \) allows direct counterfactual estimation of \( \theta' \) using the trajectory \( \mathcal{H}_\theta^T \). We begin with the special case where \( \Gbarprime(T, \mathcal{H}_{\theta'}^T) \in \mathcal{H}_\theta^T \) (i.e., \( T' = T \) in this case), referred to as {\em sample-path-wise} policy ordering. We then extend the analysis to more general settings where \( \Gbarprime(T', \mathcal{H}_{\theta'}^{T'}) \notin \mathcal{H}_\theta^T \) for any \( T' \in \mathbb{N}^+ \), which we refer to as {\em distribution-wise} policy ordering. \srscomment{Feels a bit repetitive of above}

%The {\em sample-path-wise} policy order relation $\preceq$ on the set $\Theta$ is defined as follows:
\begin{definition}
\label{definition_sample_path_order}
For two policies $\theta', \theta \in \Theta$, we write $\pi_{\theta'} \preceq \pi_{\theta}$ if, for all $T \in \mathbb{N}^+$, it holds that
\begin{align}
G_{\theta'}(\mathcal{H}_{\theta'}^{T}) \in \mathcal{H}_{\theta}^T. \label{eq:sample_path_measurable}
\end{align}
\end{definition}
%
% \srscomment{Why not $G_{\theta'}(T) \in \mathcal{H}_{\theta}^T$? Then state after the formal definition that this is implied whenever $\mathcal{H}_{\theta'}^T \in \mathcal{H}_\theta^T$ holds.}
%
\noindent The term {\em sample-path} arises from the observation that \cref{definition_sample_path_order} is satisfied whenever \( \mathcal{H}_{\theta'}^T \in \mathcal{H}_\theta^T\), since $G_{\theta'}(\H_{\theta'}^T)$ is measurable with respect to $\mathcal{H}_{\theta'}^T$ trivially.

We revisit the newsvendor problem discussed above to illustrate the sample-path ordering. With zero lead time, the state space is trivial, and observed sales are given by $N(q,D) = \min\{q,D\}$ where $D$ denotes the stochastic demand. To establish the sample-path ordering, we couple the two policies $q'$ and $q$ on the same realized sequence of demands.  Under this coupling, if $q' \leq q$, then the sales observed under $q'$ is measurable with respect to sales under $q$, since $N(q', D)$ can be recovered from $N(q,D)$.
The only subtlety is that the original lost-sales cost $C(q',D)$ is not necessarily measurable with respect to $N(q,D)$, since when $D > q$, the demand $D$ is censored.  As is standard, we therefore use the pseudo-cost (see \cref{eq_pseudo_cost}):
\[
\tilde{C}(q',D) = C(q',D) - pD = h(q'-D)^+ + p(D-q')^+ - pD,
\]
which differs from the original cost by a policy-independent term $pD$.  This transformation preserves regret by an additive constant independent of the policy. Moreover, $\tilde C(q',D)$ is measurable with respect to $N(q', D)$, since when $D > q'$ then $\tilde C(q',D) = p(D-q') - pD = - pq'$.  Therefore, $\tilde C(q',D) \in N(q',D) \in N(q,D)$ whenever $q' \leq q$.  It follows that, for every $T$, the counterfactual cost generated under $q'$ is measurable with respect to the trajectory observed under $q$, so $\mathcal H_{q'}^T \in \mathcal H_q^T$, and thus $G_{q'}(\mathcal H_{q'}^T) \in \mathcal H_q^T$.

\ifdefined\informs \subsubsection{Distributional Information Ordering} \label{sec_policy_ordering_3} \else \myparagraph{Distributional Policy Ordering} \fi
While the sample-path order captures exact measurability of one policy's cost under trajectories collected under another, many other domains only permit {\em approximate} inference. \ifdefined\informs {In such cases, it suffices to construct a random variable $\CGbarprime$ such that (i) $\CGbarprime \in \mathcal{H}_\theta^T$, and (ii) $\CGbarprime$ approximates the distribution of $\Gbarprime(\H_{\theta'}^{T'})$ for some $T' \in \mathbb{N}^+$. Measurability implies that we can construct $\CGbarprime$ from $\H_\theta^T$ to estimate the performance of $\theta'$ using samples collected from $\theta$.  The second condition ensures that the estimate is ``sufficiently good''.  Moreover, $T'$ may need to be strictly less than $T$ if the information collected under $\theta$ is insufficient to support counterfactual inference over the full horizon $T$. However, as long as $T'$ is a {\em constant fraction} (denoted with the ``information ratio'' $\alpha \in (0, 1]$)~\footnote{For notational convenience, we treat $\alpha T$ as an integer; otherwise, one may replace $\alpha T$ with $\lfloor \alpha T \rfloor$ without affecting our algorithm or the regret bounds up to logarithmic factors.} of $T$, then using $\CGbarprime$ in lieu of $G_{\theta'}(\H_{\theta'}^T)$ is sufficient.} \else To accommodate this broader class, we introduced a relaxed notion of policy ordering based on (i) using fewer than $T$ samples, {i.e., $T' = \alpha T$ where $0 < \alpha \le 1$}, and (ii) approximating the distribution of {$\Gbarprime(\H_{\theta'}^{T'})$} instead of exactly. \fi  The formal definition of this {\em distributional} ordering is given by:

\begin{definition}\label{definition_general_order}
Fix \(\alpha\in(0,1]\) and a function \(\Th:(0,1)\to\mathbb N^+\). For two policies \(\theta,\theta'\in\Theta\), we write \(\pi_{\theta'}\preceq \pi_\theta\) if, for every \(\delta \in (0,1)\), there exists a random variable $\CGbarprime$ such that
\begin{align}
d_{\mathrm{TV}}(\CGbarprime(\mathcal{H}_{\theta}^T), \Gbarprime(\H_{\theta'}^{\alpha T})) \leq \delta \quad \text{for all} \quad T \geq \Th(\delta).
\label{eq:distribution_order}
\end{align}
\end{definition}

Note that \cref{definition_sample_path_order} implies \cref{definition_general_order} under $\alpha = 1$ and $\Th(\delta) = 1$ by setting $\CGbarprime(\mathcal{H}_{\theta}^T) = \Gbarprime(\mathcal{H}_{\theta'}^{T})$. We note that our algorithms require knowledge of $\CGbarprime(\mathcal{H}_{\theta}^T)$ in order to perform counterfactual inference. 
However, these orders allow us to capture settings where exact sample-path measurability fails, by permitting approximate estimation over a fraction of the trajectory and tolerating small distributional discrepancies. 
In \Cref{sec_case_studies}, we show that the distributional order is powerful in some cases (\Cref{case_study:dual_index_policy} and \Cref{case_study:queuing}), where the sample-path policy order does not hold.

\subsubsection{Width of the Information Order and Maximal Policies}\label{section_width}
The information orders defined above do not, in general, form a partial order~\citep{dushnik1941partially}, as antisymmetry may fail.
Moreover, the distributional order need not be transitive.
Hence, technically, ``$\preceq$'' should be regarded simply as a binary relation. We define the {\em width} of such a relation as follows.
\begin{definition}
\label{def_width}
    Given a policy class $\Theta'$, the width $w$ of $\Theta'$ following the information order $\preceq$ is defined as 
    $$
        w = \max_{\Theta'' \subseteq \Theta'} \min\left\{|A| \mid A \subseteq \Theta'', \, \forall \theta \in \Theta'', \, \exists \bar{\theta} \in A \text{ s.t. } \theta \preceq \bar{\theta}\right\}.
    $$
\end{definition}
When ``$\preceq$'' is a partial order, as can occur in settings with full trajectory recovery (see the discussion of 
\(
\mathcal{H}_{\theta'}^T \in \mathcal{H}_\theta^T
\)
in~\Cref{sec_policy_ordering_2}), \Cref{def_width} coincides with the classical notion of width for a partial order defined via maximum antichains~\citep{dushnik1941partially}.
Furthermore, \Cref{def_width} implies that for every subset \(\Theta_A \subseteq \Theta'\), there exists a subset \(\Thetamax_A \subseteq \Theta_A\) with \(|\Thetamax_A| \le w\) such that for every policy \(\theta' \in \Theta_A\), there exists \(\theta \in \Thetamax_A\) satisfying \(\theta' \preceq \theta\). For notational convenience, we let \(|\Thetamax_A| = w\) by augmenting the set if necessary. We refer to each element in \(\Thetamax_A\) as a {\em maximal policy} of \(\Theta_A\), by analogy with the terminology in partial orders~\citep{dushnik1941partially}.

Note that any finite policy class $\Theta'$ admits a trivial information order, where no two policies are related. In particular, one has $\pi_{\theta'} \not\preceq \pi_\theta$ for all $\theta' \neq \theta$, resulting in an information order with width $w = |\Theta'|$.
While such orderings always exist, they offer no benefit in improving sample efficiency. In contrast, when $\Theta'$ has an information order with small width, policies can share more information with one another, leading to faster learning and lower regret. {Note that \(w=1\) implies the existence of a policy capable of counterfactual estimation for every policy in \(\Theta'\).  The full‐feedback bandit setting is a special case of the \(w=1\) full ordering~\citep{alon2015online}.} More surprisingly, several classic problems in operations have natural policy classes with information orders of small width. 

\subsection{Assumptions on the MDP} 
Our algorithm and results rely on several mild assumptions on the underlying MDP and parameterized policy class $\Theta$.  % The first requires that $\Theta$ is a compact subset of $\mathbb{R}^d$.
\begin{assumption}\label{assumption_continuity}
    The parameterized policy class $\Theta \subset \mathbb{R}^d$ is bounded by $\diam$, i.e. $\sup_{\theta \in \Theta} \norm{\theta} \leq \diam.$ 
\end{assumption}

The boundedness assumption on \( \Theta \) is mild and commonly adopted in the literature on policy learning problems~\citep{agrawal2019learning, gong2024bandits, chen2024tailored}. 
We recall that \(\Theta\) can be continuous.
The next assumption requires that there is a policy which can be used to return to the starting state $s_1$ in a bounded number of timesteps.
\begin{assumption}\label{assumption_restart}
    For any state $s\in \S$, there exists a policy $\theta_R(s)\in\Theta$ such that, under policy $\pi_{\theta_R(s)}$, the expected time to reach state $s_1$ from $s$ is bounded by $D_\Theta \in \mathbb{N}^+$.
\end{assumption}

The MDP diameter introduced by \citet{auer2008near} can serve the same purpose as \Cref{assumption_restart}, in which case the initial state $s_1$ may be chosen arbitrarily.
The last assumption requires that the loss function is uniform and the span of the bias is bounded.

\begin{assumption}\label{assumption_MRP}
For any policy $\theta \in \Theta$, let $\S_{\theta}$ denote the set of states that can arise when the system starts from \(s_1\) and follows policy \(\pi_\theta\). The following properties are satisfied:
    \begin{enumerate}
    \item The average loss function $g_\theta(s)$ is uniform and independent of the initial state $s$ (i.e. $g_\theta(s) = g_\theta$ for all $s \in \S_\theta$).
    \item The span of the bias is bounded by $H$ (i.e. $\max_{s, s' \in \S_{\theta}^2} |v_\theta(s) - v_\theta(s')| \leq H$).
    \item The cost function $g_\theta$ is Lipschitz continuous with respect to $\theta$, with Lipschitz constant $\Lip$.
    \end{enumerate}
\end{assumption}

The uniformity of $g_\theta(s)$ is closely related to coupling (see~\Cref{cor:base_stock_regret} in~\Cref{case_study:inventory} for an example) and regeneration properties (see~\Cref{lem:dual_index_bias} in~\Cref{case_study:dual_index_policy} for an example). It is also closely connected to positive Harris recurrence in stochastic processes with general state spaces; see Section~17 of~\citet{meyn2012markov} for further discussion.
Beyond \Cref{assumption_restart} and~\Cref{assumption_MRP}, the literature has also utilized mixing time~\citep{kearns2002near} and MDP diameter~\citep{auer2008near} as alternative regularity conditions. We discuss this more in \cref{section_modelling_assumptions}.

\subsection{Discussion of Modeling Assumptions}
\label{section_modelling_assumptions}
We conclude the section with a discussion of our modeling assumptions.

\paragraph{In-Class Learning within $\Theta$.} Our regret guarantees are stated with respect to the best policy within the policy class $\Theta$, namely $\theta^* = \argmin_{\theta \in \Theta} g_\theta(s_1)$.  In general, the in-class optimal policy need not coincide with the globally optimal policy.  Indeed, if $\pi^*$ denotes the {\em globally} optimal policy, then the regret relative to $\pi^*$ would incur an additional approximation error term on the order of $T(g_{\theta^*}(s_1) - g_{\pi^*}(s_1))$, which reflects the expressiveness of the policy class $\Theta$.  We focus on in-class regret for two reasons: 
(i) A large body of work in operations research has established the near‐optimality and practical relevance of many widely used heuristic policy classes~\citep{karlin1958inventory, veeraraghavan2008now, huh2009asymptotic, xin2023dual}, including those examined in \cref{sec_case_studies}.
(ii) From a theoretical perspective, sample-efficient learning in continuous state and action spaces requires some form of model complexity control. Existing work imposes such model complexity in different ways, including policy class~\citep{mou2020sample}, linearity~\citep{wan2024exploiting}, Lipschitz continuity (see~\Cref{discussion_complexity_2}), convexity~\citep{agrawal2019learning}, function class complexity (e.g., in Eluder dimension~\citep{russo2013eluder} and  Bellman rank~\citep{pmlr-v70-jiang17c}). 
Our approach places the complexity directly on the policy class, inspired by many operational settings discussed above where there are well-known heuristics.

\paragraph{Technical Details on $g_{\theta}$ and $v_\theta$.}
To employ the long-run average cost $g_{\theta}$ and the bias $v_\theta$ in the analysis, we require that the corresponding limits exist. A sufficient condition ensuring the existence of $g_{\theta}$ is that the state space is finite (see Proposition 8.1.1 of~\citet{puterman2014markov}).
Technically, for an MDP with a continuous state space satisfying \Cref{assumption_continuity}--\ref{assumption_MRP}, we can discretize the state space to ensure that $g_\theta$ is well-defined, while allowing the discretization radius to be {\em arbitrarily small}. This has no impact on either the algorithm or the regret bound (see \Cref{alg_policy_elimination} and \Cref{thm:regret_bound}). 
In particular, although the discretization radius may appear to affect the span of the bias function, since the span is defined over all $s, s' \in \S_\theta$, this span is uniformly bounded by $H$ under \Cref{assumption_MRP}, and this bound is independent of the discretization radius above. We provide two examples in \Cref{sec_case_studies} (\Cref{case_study:inventory} and~\Cref{case_study:dual_index_policy}) to illustrate how such a bound on $H$ can be established without relying on the cardinality of the state space.  Accordingly, we treat the discretization as a purely technical device and view our algorithm as operating effectively over compact spaces. See Remark~3 of~\citet{agrawal2019learning} for more discussion. 

We also note that $v_\theta$ is well-defined under finiteness above, i.e., the corresponding limit exists for all $\theta \in \Theta$; see Section 8.2 of~\citet{puterman2014markov}. It is worth mentioning that when the chain is periodic, the definition of the bias should be interpreted in the Cesaro sense. All results continue to hold regardless of whether the chain is aperiodic. We do not pursue this point further and refer to Section 8.2.1 of~\citet{puterman2014markov} for additional discussion.
Additional technical properties, such as the interchange of limits and expectations, are addressed as needed in subsequent sections.

\paragraph{Ergodic Regularity Parameters and $D_{\Theta}$.}
Beyond the bounded-bias-span condition in \Cref{assumption_MRP}, two other ergodicity regularity parameters are studied in the infinite-horizon MDP literature, the diameter and the mixing time.  Let
\(
D_{\mathrm{diam}} := \sup_{\pi} \max_{s, s' \in \S} \Ex_{\pi}[\tau(s') \mid S_1 = s]
\)
denote the diameter of the MDP, where $\tau(s')$ is the first hitting time of state $s'$; see the definition of $D_{\mathrm{worst}}$ in~\citet{bartlett2012regal}.  We also let $t_{\mathrm{mix}}$ denote a uniform mixing-time parameter; see Section~5 of \citet{wang2022near}.  Under the arbitrarily fine state-space discretization discussed above, the relationships among $H$, $D_{\mathrm{diam}}$, and $t_{\mathrm{mix}}$ are well-established in the literature. In particular, the bias span satisfies $H \lesssim D_{\mathrm{diam}}$ and $H \lesssim t_{\mathrm{mix}}$ up to problem-dependent constants; see~\citet{bartlett2012regal,wang2022near}.  Moreover, the restart parameter in~\Cref{assumption_restart} can be chosen to satisfy $D_\Theta \leq D_{\mathrm{diam}}$, as follows directly from the definitions. The diameter and mixing time are not comparable in general~\citep{wang2022near}.  However, these implications do not reverse in general; a bounded bias span $H$ does not imply a bounded diameter, and hence does not imply a bound on $D_\Theta$. This is why \Cref{assumption_restart} is imposed separately from \Cref{assumption_MRP}.

We also emphasize that these constants may depend implicitly on geometric properties of the state and action spaces, such as the {\em diameter} of \(\S\); e.g., see~\Cref{cor:base_stock_regret} in~\Cref{case_study:inventory}. However, they need not depend on the cardinality induced by an arbitrarily fine discretization of the state space as discussed above, and can remain bounded in terms of problem-specific parameters. Consequently, such dependence may affect the leading constant, but does not alter the $\sqrt{T}$-type dependence on $T$ of the regret bounds studied in this work.

\section{Main Results}\label{sec_main_result}

%Goal of this section:
%\begin{itemize}
%    \item Introduce the two main algorithms for distribution-wise partial order and sample-path-wise partial order, respectively. (section 4.1)
%    \item Proof sketch of Theorem 1 (section 4.2)
%\end{itemize}

%\subsection{Information-Ordered Policy Elimination Algorithm}
%The several things of the algorithm we want to discuss:
%\begin{itemize}
%    \item The evolvement of the algorithm: At first we do discretization following $\epsilon = O(1/\sqrt{T})$, then we divide $T$ into $K$ epochs, and in each epoch we employ one specific policy. After each epoch we eliminate the suboptimal policies. 
%    \item Detailed analysis of the policy elimination. The key observation is that we can bound the probability of the event, that the optimal policy is within the confidence set of each epoch, with high probability. 
%    \item Give the pseudo-code of the algorithm here.
%\end{itemize}
%\srscomment{Rewrote this section substantially}
%\srscomment{Need to emphasize that $w$ is the width of $\Theta_1$.....so it is somehow a function of the discretization. That isn't coming out in the current discussion.}

\begin{algorithm}[!t]
\caption{Information-Ordered Epoch Based Policy Elimination Algorithm (\ALG)}
\label{alg_policy_elimination}
\begin{algorithmic}[1]
    % \State \textbf{Input:} Starting state $s_1$, policy set $\Theta \subset \mathbb{R}^d$, partial order $\preceq$ with width $w$, confidence terms $\beta_k$, radius of $\epsilon$-net $r$,  time horizon $T$
    \State Discretize $\Theta$: $\Theta_1 = \mathcal{N}_{r}(\Theta)$ 
    \hfill // \textit{Discretization}
    \For{epoch $k = 1, 2, \dots$}
        \State Compute the maximal policies of \( \Theta_k \): \( \Thetamax_k = \{\pi_{\theta_{k1}}, \ldots, \pi_{\theta_{kw}}\} \) following order $\preceq$ (see \eqref{eq:theta_k_max}) \\ \hfill // \textit{Compute most informative policies}
        \For {each $j \in [w]$}
            \State play $\pi_{\theta_{kj}}$ for $N_k$ timesteps to obtain $\mathcal{H}^{N_k}_{\theta_{kj}}$
            \State Play $\pi_{\theta_R}$ until returning to state $s_1$ \hfill // \textit{Return to starting state}
        \EndFor
        \State Use $\{\mathcal{H}^{N_k}_{\theta_{k1}}, \ldots, \mathcal{H}^{N_k}_{\theta_{kw}}\}$ to estimate $\CGbarprimez$ for all $\theta' \in \Theta_k$
        \hfill // \textit{Derive counterfactual estimates}
        %with policy order following \Cref{definition_sample_path_order} or \ref{definition_general_order} \\
        \State Update $\Theta_{k+1} = \{\theta' \in \Theta_k \mid \CGbarprimez - \min_{\theta \in \Theta_k} \CGbarz \leq 2 \beta_k \}$
    \EndFor
\end{algorithmic}
\end{algorithm}

In this section, we present our Information-Ordered Epoch-Based Policy Elimination Algorithm (\ALG), as well as its regret guarantee. 
We then complement it with a nearly-matching lower bound.

\subsection{Algorithm Design}

At a high level, our algorithm builds on the insight that policy learning can be accelerated by systematically leveraging the information order over the policy class for counterfactual inference.  Across a series of epochs we maintain a confidence set $\Theta_k$ of near-optimal policies.  This set is refined over time throughout the course of learning as more data is collected. Rather than exploring each policy within $\Theta_k$ across each epoch, we sample the {\em most informative policies} (the ``maximal'' policies in the information order, those that enable counterfactual evaluation of all other policies in $\Theta_k$), thereby improving sample efficiency and regret guarantees.  We now describe each stage in detail (see \cref{alg_policy_elimination} for full pseudocode).

\myparagraph{Discretizing the Policy Set $\Theta$.} We begin by discretizing the policy space \( \Theta \subset \mathbb{R}^d \) with an \( r \)-net \(\Theta_1 = \mathcal{N}_r(\Theta) \) under the \( \ell_\infty \)-norm.  Note that by \Cref{assumption_continuity} we know that $\Theta$ is bounded, and so we will later set {\(r = (1/T)^{1/2}\)}.  This balances the loss due to learning and the loss due to discretization (handled since $g_\theta$ is Lipschitz under \Cref{assumption_MRP}).

\myparagraph{Epochs and Policy Sampling.} 
The learning process proceeds over a series of epochs $k = 1, 2, \ldots$, where within each epoch we maintain a set $\Theta_k \subseteq \Theta_1$ of near-optimal policies.  
In order to collect samples to further estimate policy performance and refine \(\Theta_k\), we instantiate the notation of {\em maximal} policies in~\Cref{section_width} with \(\Theta_A = \Theta_k\). Concretely, for each \(\Theta_k\), we define
\begin{equation}
\label{eq:theta_k_max}
\Thetamax_k = \{\pi_{\theta_{k1}}, \ldots, \pi_{\theta_{kw}}\} {\subset \Theta_k},
\end{equation}
based on the information order \(\preceq\), where \(w\) is the corresponding width of $\Theta$.
Importantly, \(\Thetamax_k\) is chosen with cardinality \(w\) such that for every policy $\theta \in \Theta_k$, there exists $\tilde{\theta} \in \Thetamax_k$ with $\pi_\theta \preceq \pi_{\tilde{\theta}}$. This property ensures that by sampling only the policies in $\Thetamax_k$, we obtain sufficient counterfactual information to evaluate the performance of every policy in $\Theta_k$.

For each policy in $\Thetamax_k$, we sample a trajectory of length $N_k =  \frac{4^k}{\alpha}\Th\left(\frac{1}{|\mathcal{N}_{r}(\Theta)| T\log(T)}\right)$. We assume without loss of generality that $\frac{1}{\alpha}\Th\left(\frac{1}{|\mathcal{N}_{r}(\Theta)| T\log(T)}\right)$ is an integer.~\footnote{Otherwise, replacing $N_k$ by $\left\lceil \frac{4^k}{\alpha} \Th\left(\frac{1}{|\mathcal{N}_{r}(\Theta)| T\log(T)}\right) \right\rceil$ affects the regret by at most a constant factor. This is because the rounding increases the total length up to epoch $K$ by at most $O(K)=O(\log (T))$ (see~\cref{eq_main_proof_2} in Appendix). Hence, the maximal feasible number of epochs $K$ changes by at most a constant.
}
Note that this scales exponentially with the epoch index, ensuring sufficient samples for reliable estimation.  Lastly, after playing each of the $w$ policies, we require a ``restart'' step to ensure that we start collecting our trajectories from the fixed initial state $s_1$.  The cumulative regret incurred by these restart steps is bounded due to \Cref{assumption_restart}. 

\myparagraph{Updating Confidence Set.} 
Due to the information order and the construction of $\Thetamax_k$, for each $\theta' \in \Theta_k$, there exists $\tilde{\theta}_k(\theta') \in \Thetamax_k$ such that $\pi_{\theta'} \preceq \pi_{\tilde{\theta}_k(\theta')}$. Using this ordering, we update the confidence set at the end of each epoch:
    \begin{align}
{\Theta_{k+1} = \left\{\theta' \in \Theta_k \mid \CGbarprimez - \min_{\theta \in \Theta_k} \CGbarz \leq 2 \beta_k \right\}},
    \end{align}
where \( \beta_k \) is a time-dependent confidence parameter that decreases with rate $O(\frac{1}{\sqrt{N_k}})$. This reflects the intuition that, as more data is accumulated, the true optimal policy lies in a progressively smaller region of the parameter space. 
Such policy-elimination approaches are standard in several related settings, including bandits~\citep{gong2024bandits} and linear mixture MDPs~\citep{jia2020model}.
Appropriate tuning of \( \beta_k \) ensures that the confidence set contains the optimal policy with high probability throughout the learning process.

\subsection{Regret Upper Bound}
\label{sec_regret_upper_bound}
We are now ready to state our main theorem; its proof is deferred to~\Cref{app:regret_bound_proof}.~\footnote{While we assume $T$ is known, the standard doubling trick can be applied to obtain an anytime version of the algorithm without affecting the regret guarantees~\citep{lattimore2020bandit}.
}

% \srscomment{The $w$ isn't an assumption, it's a definition for a partial order.  State the theorem once. Then in a myparagraph just say what it gives for the sample-path policy order (no need to state the same theorem twice).  Also noting that you use different definitions for $\preceq$ depending on the ``type'' of order. Just use the same one, since one subsumes the other.}
% \srscomment{Technically $\Th(\delta)$ here right?  Feels like we need another requirement that $\Th(\delta)$ is not too large? Or that $T$ is large relative to $\Th(\delta)$? For instance as $\delta \rightarrow 0$ then $\Th(\delta) \rightarrow \infty$ probably, and so you only get one epoch actually.}

% \srscomment{TODO: Add the dependence on $U$, likely linear, also Lipschitz, and then also $\diam$? And maybe $D_\Theta$?}
% \zzcomment{Haven't finished updating all constants in all theorems/corollaries/lemmas}

% \srscomment{Need to specify $r$ in the Theorem statement.  Just do expected regret to deal with the $D_\Theta$ terms. $r = (U / T)^{1/2}$?}

\begin{restatable}[Regret Upper Bound]{theorem}{RegretBound} 
\label{thm:regret_bound}
Let $w$ denote the width of the information order $\preceq$ restricted to $\Theta_1 = \mathcal{N}_{r}(\Theta)$ for $r = T^{-1/2}$.  Then under \Cref{assumption_continuity,assumption_restart,assumption_MRP}, by setting $\beta_k = \frac{ H}{\alpha N_k} + (H + 2\Cupperbound)\sqrt{\frac{2 \log(4|\mathcal{N}_{r}(\Theta)| T\log(T))}{\alpha N_k}}$ and $N_k =  \frac{4^k}{\alpha}\Th\left(\frac{1}{|\mathcal{N}_{r}(\Theta)| T\log(T)}\right)$, \Cref{alg_policy_elimination} ensures that, for any $T \ge \frac{4w}{\alpha}\Th\left(\frac{1}{|\mathcal{N}_{r}(\Theta)| T\log(T)}\right)$,
\begin{align}
\Regret(T) &= O\Biggl(\left( (H+2\Cupperbound) \sqrt{ \frac{w d}{\alpha} \log\left(  UT^{\frac{3}{2}}\log(T) \right)} + L_{\Theta} \right) \sqrt{T} \nonumber \\
& \quad + \left( \frac{w H}{\alpha} + w \Cupperbound D_{\Theta} \right) \log(T) + \frac{w\Cupperbound}{\alpha}\Th\left(\frac{1}{|\mathcal{N}_{r}(\Theta)| T\log(T)}\right)\Biggr),
\end{align}
where $O(\cdot)$ hides absolute constants.
\end{restatable}

% \srscomment{Just thinking - should we put the explicit value for $N_r(\Theta)$ throughout here? Since we are assuming compact and bounded by $U$ we know that $N_r(\theta) = (U / r)^{1/d}$ or whatever?}

%\srscomment{Maybe put the bounded width as an assumption in the policy order section vs in the statement of this Theorem?}\zzcomment{I tried, but as in section 3.2 we need to discretize $\Theta \subset R^d$, we should define dimension $d$ before section 3.2? To discuss.}

%\srscomment{Hmm - $T_h$ depends on $\delta$ right?}\zzcomment{yes. but since $T_h$ cancels out in the regret, should we mention that $T_h$ depends on $\delta$?}

Since the sample-path information order is a special case of the distributional order (with $\alpha = 1$ and $\Th(\cdot) = 1$), a direct corollary of \cref{thm:regret_bound} yields a regret bound of $\tilde{O}(H \sqrt{wdT})$ in this case.  
Note that the regret bound in~\Cref{thm:regret_bound} also applies to settings in which the policy class \(\Theta\) is discrete, in which case \(d\) is replaced by \(\log(|\Theta|)\); see~\cref{eq_main_proof_3} in the appendix.

Our analysis assumes knowledge of the information order, including the parameters \(w\), \(\alpha\), and \(\Th(\cdot)\), the Lipschitz constant \(\Lip\), as well as the bias upper bound $H$. These quantities can often be bounded using known structural properties of the problem studied~\citep{fruit2018efficient}.
We illustrate in \cref{sec_case_studies} how these quantities can be bounded for each case study we investigated individually.
Also, in some cases, conservative bounds may be used for these unknown quantities when they are difficult to estimate, while preserving the regret guarantee in~\Cref{thm:regret_bound} up to additional logarithmic factors; see \Cref{discussion_unknown_parameters}.
Furthermore, when $\alpha \rightarrow 0$ the regret bound deteriorates. This is a scenario where the distributional order is {\em uninformative}, and one can obtain better regret guarantees with the uninformative order (no two policies are related) with width $w = |\Theta_1|$ and $\alpha = 1$. This further implies that whenever a low‐width information order can be established, the regret bound depends on \(w\) rather than \(|\Theta_1|\), offering a sharper characterization of learning complexity under our information order framework. 

The regret bound in~\Cref{thm:regret_bound} enables a unified comparison across different feedback regimes within our framework. When no side observations are available, \(w=|\Theta_1|\), and our bound reduces to the standard $T^{\frac{d+1}{d+2}}$ scaling regret bound; see~\Cref{propo_complexity_2}. At the other extreme, when each implemented policy provides sufficient information to evaluate all policies, we have \(w=1\), yielding a \(\tilde O\left(H\sqrt{dT/\alpha}\right)\) regret bound. This matches the \(\sqrt{T}\)-type dependence on regret obtained by \citet{dann2020reinforcement} under full feedback in episodic tabular MDPs, which is independent of the state and action space sizes, up to the distinction between the infinite-horizon average-cost and finite-horizon settings.
While their setting is restricted to tabular MDPs and feedback graphs over state-action pairs, our framework applies to continuous state and action spaces and captures information relationships directly over the policy class.
Beyond the no-feedback and full-feedback settings above, our regret guarantee (i) interpolates between these two extremes, and (ii) allows for more complex information structures within the distributional order.  In \cref{sec_case_studies} we provide several case studies in operations research and specialize \cref{thm:regret_bound} to those case studies.

\myparagraph{Complexity Analysis of~\ALG.}

We next study the computational and storage requirements of~\ALG. The following proposition gives the complexity under~\Cref{assumption_complexity_oracle}, which assumes that maximal policies can be identified in $O(w)$ time and that each counterfactual estimator over an epoch of length $N_k$ can be constructed in $O(N_k)$ time (see~\Cref{discussion_complexity} for details).

\begin{restatable}{proposition}{ComplexityBoundforALG}
    \label{propo_complexity_1}
    Under~\Cref{assumption_complexity_oracle}, the time complexity of \ALG is $O\bigl(w \log(T) + |\Theta_1| \frac{T}{w} \bigr)$ almost surely, while the space complexity is $O(|\Theta_1|)$.  
    In particular, if $\Theta \subset \RR^d$ and $\Theta_1 = \mathcal N_r(\Theta)$ with $r = T^{-1/2}$ as in~\Cref{thm:regret_bound}, the time complexity of~\ALG is $O(w\log(T)+T^{1 + \frac{d}{2}})$ almost surely, while the space complexity is $O(T^{\frac{d}{2}})$ almost surely.
\end{restatable}

\cref{propo_complexity_1} highlights a computational price for the statistical gains obtained from leveraging the information order. When the width $w$ is small, \ALG obtains the $\tilde O\left(H\sqrt{\frac{wd}{\alpha}T}\right)$ regret guarantee in \Cref{thm:regret_bound}. This improvement comes from implementing only a small set of maximal policies in each epoch, while using their trajectories to form counterfactual estimates for all remaining policies.  However, because \ALG does not assume global structure such as convexity or linearity of $g_\theta$, it still maintains and updates estimates for every candidate policy in the discretized class $\Theta_1$. Consequently, the time and storage requirements continue to scale with $|\Theta_1|$.  In order to get the improved regret guarantees, one needs a {\em finer} discretization of $\Theta$ where $r = T^{-1/2}$.  This finer grid increases $|\Theta_1|$, causing the larger time and space bounds.
In contrast, under the trivial order where no policy provides information about any other policy, one optimizes the discretization radius at $r = T^{-1/(d+2)}$, yielding a $T^{\frac{d+1}{d+2}}$ scaling regret bound.  While this has slightly better time and space complexity, the regret is {\em exponentially worse}.

Thus, the information order primarily improves statistical efficiency rather than eliminating the computational dependence on the policy class. 
Further discussion of this trade-off is deferred to~\Cref{discussion_complexity} and~\Cref{tab_complexity_comparison}, where we compare \ALG with the convexity-based algorithm of~\citet{agrawal2019learning} and the feedback-graph framework of~\citet{dann2020reinforcement}.

\ifdefined\informs
\myparagraph{Choice of Information Order.}
When the policy class \(\Theta\) admits multiple valid orderings, it is essential to compare and select among them. Among \emph{sample-path} orders (\(\alpha=1\)), Theorem~\ref{thm:regret_bound} asserts that the ordering with minimal width \(w\) yields the tightest bound. For \emph{distributional} orders, minimizing the ratio \(w/\alpha\) directly reduces the leading constant in~\cref{thm:regret_bound}; nevertheless, this criterion does not always lead to the optimal information order. The distributional order only yields regret bounds for horizons \(T \ge \Th(\delta)\) at fixed confidence \(\delta\), where the threshold \(\Th(\delta)\) itself depends on the chosen ordering, making order selection ambiguous in non-asymptotic \(T\) settings. Empirically, in our dual‐sourcing experiments (\Cref{case_study:dual_index_policy}) where the worst‐case bound on \(\alpha\) decays exponentially in the lead time \(L_r\), we still observe rapid convergence (Figure~\ref{fig_convergence_curve}), indicating that the \emph{effective} mixing rate is typically much higher (often \(O(1)\)) than the pessimistic estimate.  Thus, a pragmatic heuristic is to choose the order that minimizes \(w\) first, regardless of \(\alpha\).  A systematic study of policy‐order selection across diverse MDPs is left for future work.

\fi

\ifdefined\informs
\begin{rproofsketchof}{\Cref{thm:regret_bound}}
For each policy $\theta' \in \Theta_k$ we evaluate at epoch $k$, we first establish the concentration of $\CGbarprimez$ to its true long-run average cost $g_{\theta'}$ by combining a martingale concentration argument and the bounded total variation distance from the distributional order.  Thus we establish that with high probability, $|\CGbarprimez - g_{\theta'}| \leq \beta_k$.

% % both empirical average (\Cref{lem_concentration_lem_shipra}) and counterfactual estimates (\Cref{main_lem1}) around the true long-run average cost by combining a martingale concentration argument and the bounded total variation distance in~\Cref{definition_general_order}. Thus with union bound, we establish that for any $\delta > 0$, if denoting
% \[
% \E_0 = \left\{ \forall k, \theta' \in \Theta_k \mid \abs{\CGbarprimez - g_{\theta'}} \leq \frac{ H}{\alpha N_k} + (H+2)\sqrt{\frac{2 \log(4 |\mathcal{N}_{r}(\Theta)| K / \delta)}{\alpha N_k}}\right\},
% \]
% then $\E_0$ occurs with probability at least $1 - 2\delta$.

Next, we show by 
%
%Next, conditioned on the concentration events (based on $\E_0$), we show by 
induction (Lemma~\ref{lem_a.5}) that with high probability, the discretized optimal policy $\theta^*_r = \argmin_{\theta \in \mathcal{N}_{r}(\Theta)} g_\theta$
is never eliminated for each epoch $k$.  Specifically, at each epoch \(k\) with $k \ge 2$, any surviving policies \(\theta, \theta'\in\Theta_k\) satisfy \(\bigl|g_\theta - g_{\theta'}\bigr|\le 16\beta_k\). 

Finally, we decompose the total regret over \(T\) steps by summing over each epoch.  In epoch \(k\) with $k \ge 2$, at most \(w\) policies are implemented for \(N_k\) steps each, incurring regret at most \(w N_k\cdot O(\beta_k)\).  Returning to the initial state adds at most \(wD_{\Theta}\) cost per epoch in expectation.  Choosing \(N_k = O\bigl(4^k\Th\left(\frac{1}{|\mathcal{N}_{r}(\Theta)| T\log(T)}\right)\alpha^{-1}\bigr)\) forces the number of epochs \(K = O(\log (T))\). 
Summing $w N_k \cdot O(\beta_k)$ over \(k\ge 2\), and incorporating the first-epoch term, the return cost, and the discretization error $O(T r L_{\Theta})$, together with the bound $|\mathcal{N}_{r}(\Theta)| = O((U\sqrt{T})^d)$ for $r = T^{-1/2}$, completes the proof.
\end{rproofsketchof}

\else 
\ifdefined\informs

\subsection{Proof Sketch of \Cref{thm:regret_bound}}
We outline the main ideas of the proof of \Cref{thm:regret_bound} in three steps.

\myparagraph{1. Concentration for Counterfactual Estimates via Total‐Variation Bounds.}  
For any fixed policy \(\theta'\), recall that $\Gbarprime(\H_{\theta'}^{T})$ denotes its empirical average cost over \(T\) samples.  By a standard martingale concentration argument (\Cref{lem_concentration_lem_shipra}), we have that with probability at least \(1-\delta\),
\[
\bigl|\Gbarprime(\H_{\theta'}^{T})  - g_{\theta'}\bigr| \le \tilde{O}\left(\frac{H}{T} + (H+2)\sqrt{\frac{2\log(4/\delta)}{T}}\right).
\]
Next, for each epoch \(k\) and each policy \(\theta'\in\Theta_k\), the counterfactual estimate $\CGbarprimez$ is derived from a trajectory under a “maximal” policy \(\tilde\theta_k(\theta')\).  Policy‐ordering (either sample‐path or distributional) guarantees that the total‐variation distance between $\CGbarprimez$ and the empirical estimate of $g_{\theta'}$, $\Gbarprime(\H_{\theta'}^{\alpha N_k})$, is at most \(\delta\).  Therefore, by union bound, we obtain that for any fixed $\delta > 0$, if define the following event:
\[
\E_0 = \left\{ \forall k, \theta' \in \Theta_k \mid \abs{\CGbarprimez - g_{\theta'}} \leq \frac{ H}{\alpha N_k} + (H+2)\sqrt{\frac{2 \log(4 |\mathcal{N}_{r}(\Theta)| K / \delta)}{\alpha N_k}}\right\},
\]
then we have that $\E_0$ occurs with probability at least $1 - 2\delta$.

\myparagraph{2. Maintaining the Optimal Policy in the Active Set.}  
Conditioned on the concentration events (the exact definition of the concentration event is based on $\E_0$ but more complicated thus deferred to \Cref{app:regret_bound_proof}), one shows by induction (Lemma~\ref{lem_a.5}) that the discretized optimal policy 
$$
    \theta^*_r = \argmin_{\theta \in \mathcal{N}_{r}(\Theta)} g_\theta
$$
is never eliminated for each epoch $k$.  Specifically, at each epoch \(k\), any surviving policies \(\theta, \theta'\in\Theta_k\) satisfy \(\bigl|g_\theta - g_{\theta'}\bigr|\le 6\beta_k\).

\myparagraph{3. Regret Decomposition and Epoch Count.}  
Conditioned on the concentration events, we decompose the total regret over \(T\) steps by summing over each epoch.  In epoch \(k\), at most \(w\) policies are implemented for \(N_k\) steps each, incurring regret at most \(w N_k\cdot \tilde{O}(\beta_k)\).  Returning to the initial state adds at most \(wD_{\Theta}\) cost per epoch.  Choosing \(N_k = \tilde{O}\bigl(4^k \alpha^{-1}\log T\bigr)\) forces the number of epochs \(K = \tilde{O}(\log T)\).  Summing \(w N_k\cdot \tilde{O}(\beta_k)\) over \(k=1,\dots,K\) and adding the returning cost and discretization error \(O(T\,r\,L_{\Theta})\) yields
\[
\Regret(T) \leq O\left(\frac{wH}{\alpha} + wD_{\Theta}\right)\log_4 \frac{\alpha T}{w \Th(\delta)} + O\left((H+2) \sqrt{\frac{w}{\alpha} \log(4 |\mathcal{N}_{r}(\Theta)| K / \delta)T}\log(T)\right) +  TL_{\Theta}r.
\]
By taking $r =(1/T)^{1/2}$, we have $|\mathcal{N}_{r}(\Theta)| = O((U\sqrt{T})^d)$, which completes the proof.
\fi
\fi

\subsection{Regret Lower Bound}
\label{sec_lower_bound}
Lastly, we complement this regret upper bound with a nearly matching lower bound. 
For simplicity, we state the lower bound under the normalization \(C(s,a) \in \Delta[0,1]\). The proof and discussion are deferred to~\cref{app_proof_lower_bound_new}.

\begin{restatable}[Regret Lower Bound]{theorem}{RegretLowerBoundNew} 
\label{theorem_lower_bound_new}
For any fixed integers $d$, $w$, and $H$, for any $T \ge \max\{128\log(2)Hd, \frac{2560\log(40)}{3}H\}$, for any algorithm, there exists an MDP with continuous state and action spaces and a compact policy class $\Theta \subset \RR^d$ satisfying the sample-path information order of width $w$ and \Cref{assumption_continuity,assumption_restart,assumption_MRP}, such that
\[
\Regret(T) = \Omega\left(\min\left\{\sqrt{HwdT}, H^{\frac{1}{d+2}}T^{\frac{d+1}{d+2}}\right\}\right),
\]
where $\Omega(\cdot)$ only hides absolute constants.
\end{restatable}

Note that~\Cref{theorem_lower_bound_new} also covers the trivial information order discussed in~\Cref{sec_regret_upper_bound}. In particular, for a compact policy class with $|\Theta|=\infty$, one may
have an unbounded $w$. This regime gives rise to the second term,
\(
\Omega\left(H^{\frac{1}{d+2}}T^{\frac{d+1}{d+2}}\right),
\)
which is also tight compared with~\Cref{tab_complexity_comparison} (see the row $w = |\Theta_1|$). Furthermore, when $w$ and $d$ are bounded, the lower bound in~\Cref{theorem_lower_bound_new} is $\Omega(\sqrt{HwdT})$, since this term is smaller than $\Omega\left(H^{\frac{1}{d+2}}T^{\frac{d+1}{d+2}}\right)$ for sufficiently large $T$. 

Thus, \Cref{theorem_lower_bound_new} shows that the dependence of the regret bound in~\Cref{thm:regret_bound} on $T$, $d$, and $w$ is optimal up to logarithmic factors. \ifdefined\informs There is a $\sqrt{H}$ gap between the lower bound and the upper bound in~\Cref{thm:regret_bound}; such a gap is common in the literature (see~\Cref{remark_sqrtH_discrepancy}). \Cref{theorem_lower_bound_new} also ignores the potential dependence on $\alpha$. 
Note that for any compact policy class $\Theta$, the trivial order with width $w=|\Theta_1|$ after discretization yields a $T^{\frac{d+1}{d+2}}$ scaling regret bound under the choice $r=T^{-\frac{1}{d+2}}$; see~\Cref{tab_complexity_comparison}. This rules out a general lower bound of order \(\Omega\bigl(\sqrt{HwdT/\alpha}\bigr)\), since, as $\alpha\to 0$, the naive \(T^{\frac{d+1}{d+2}}\) scaling upper bound outperforms it. However, our notion of distributional order, as well as \ALG, does not assume that \(C_t(S_t,A_t)\) is observed at every step. To capture the \(\alpha\) dependence, we therefore analyze a family of MDPs where costs are revealed only on an \(\alpha\)-fraction of time steps.  We show that in this case, any algorithm incurs \(\Omega(\sqrt{HwdT/\alpha})\) regret. Thus our results are tight in $\alpha$ as well. More details are provided in \Cref{theorem_lower_bound_new_alpha_dependence}.
\section{Case Studies}\label{sec_case_studies}

To demonstrate the practical relevance of our framework, we instantiate it in several canonical problems in operations research, including inventory control with positive lead time~\citep{goldberg2021survey}, dual sourcing~\citep{whittemore1977optimal}, and queuing models with state-dependent service rates~\citep{puterman2014markov}. In each case, we identify natural policy classes that admit a low-width information order, enabling new algorithms and novel regret guarantees. We further complement our theoretical findings with numerical simulations comparing our method to existing baselines in~\cref{sec:simulations}. 
Additional proofs and details, together with two extensions of the case study in~\Cref{case_study:inventory}, one incorporating returns under zero lead time~\citep{chen2025managing}, detailed in~\Cref{case_study_return}, and the other extending the model to multiple products, detailed in~\Cref{case_study_multiple_products}, are provided in~\Cref{appendix_case_studies}. The information order framework also applies to several bandit-like models studied in the literature, such as the first-price auction model in~\citet{han2025optimal}. We omit further discussion of these models because the corresponding information orders are more straightforward in bandit-like settings.

\subsection{Single-Retailer Inventory Control with Positive Lead Time}
\label{case_study:inventory}

Our first case study considers a single-retailer inventory control problem with positive lead time, a canonical supply chain management problem in the operations research literature (see \citet{goldberg2021survey} for a survey of recent structural results on this problem).

\subsubsection{Model}

A retailer is faced with making ordering decisions $\Order_t$ online over a period of rounds $t = 1, \ldots, T$.  At the beginning of each step $t$, the inventory manager observes the current inventory level $\Inventory_t$ as well as the $L$ previous unfulfilled orders in the pipeline, denoted as $\Order_{t-L}, \ldots, \Order_{t-1}$.  Here, the integer $L \geq 1$ is the so-called {\em lead time}, or delay in the number of steps between placing an order and receiving it.  The system evolves according to the following dynamics.  At the beginning of each stage $t$, the inventory manager picks an order $\Order_t$ to arrive at stage $t+L$.  Then, the order $\Order_{t-L}$ that was made $L$ time steps earlier arrives. Next, an unobserved demand $D_t \geq 0$ is generated independently from an unknown distribution $\F$, which we assume is supported on $[0, U]$ for simplicity; however, our results extend to more general demand distributions. We assume that $\Pr(D_t = 0) > 0$.  The number of products sold is the minimum of on-hand inventory and demand, i.e. $\min\{ \Inventory_t + \Order_{t-L}, D_t\}$.  Note that the decision maker only observes the sales, and not the actual demand $D_t$.

\myparagraph{MDP Formulation.}  To formulate this as an MDP, the state space is $\S = [0,(L+1)U]^{L+1}$, where each state $s$ consists of the current inventory level $\Inventory_t$ along with the previous $L$ unfulfilled orders $\Order_{t-L}, \ldots, \Order_{t-1}$ in the pipeline.  The action space is given by $\A = [0,U]$ where action $A_t = \Order_t$ denotes the order placed at time $t$.  The transition and cost dynamics are:
\begin{align}
    S_{t+1} & = ((\Inventory_t + \Order_{t-L} - D_t)^+, \Order_{t-L+1}, \ldots, \Order_t) \label{eq_ic_dynamics}\\
    C_t(S_t) & = h(\Inventory_t + \Order_{t-L} - D_t)^+ + p(D_t - (\Inventory_t + \Order_{t-L}))^+,
\end{align}
where $h$ and $p$ are the holding and lost-sales cost coefficients. 
We assume a fixed starting state of {$s_1 = (0, \ldots, 0)$}, corresponding to no on-hand inventory and no outstanding orders in the pipeline.
We emphasize that all model variables are continuous, including the states (inventory positions), the actions (order quantities), policies (order-up-to levels), and demand.

\myparagraph{Modifications to Cost Function.}
As written, $C_t(S_t)$ is not observed, since it depends on the realized demand $D_t$ whereas the algorithm only has access to {\em sales} data $N_t = \min\{D_t, \Inventory_t + \Order_{t-L}\}$.  However, one can transform the cost to a so-called {\em pseudo-cost} which maintains the same average cost up to a constant independent of the policy~\citep{agrawal2019learning}:
\begin{align}
\label{eq_pseudo_cost}
\tilde{C}(S_t) = C(S_t) - pD_t = h(\Inventory_t + \Order_{t-L} - \min\{\Inventory_t + \Order_{t-L}, D_t\}) - p \min\{\Inventory_t + \Order_{t-L}, D_t\}.
\end{align}
As is common in the literature, our results on the information order leverage this pseudo-cost, but when reporting the performance of the algorithms, we report the true cost $C_t(S_t)$. We emphasize that our results apply regardless of the choice of cost function, since unlike existing literature, we do not rely on convexity~\citep{agrawal2019learning}.  Indeed, there are several well-studied models in operations (including step-dependent holding and lost-sales costs) where the cost function is no longer convex~\citep{chen2021eoq}.

\subsubsection{Policies and Information Order}
We consider the class of {\em base-stock} policies parameterized by their base stock level $\theta \in [0,U] = \Theta$~\citep{goldberg2021survey}.  Intuitively, these policies order a quantity that brings the sum of leftover inventory and outstanding orders to $\theta$.  Formally, fixing the base-stock level $\theta$, the action at step $t$ is given by:
\begin{equation}
\pi_\theta(\Inventory_t, \Order_{t-L}, \ldots, \Order_{t-1}) = \Order_t = (\theta - \Inventory_t - \tsum_{i \in [L]} \Order_{t-i})^+.
\end{equation}
Prior work such as~\citet{huh2009asymptotic,zipkin2008old} show that base-stock policies are optimal when either $L = 0$ or the {lost-sales cost $p \rightarrow \infty$.} See~\citet{goldberg2021survey} for more discussion.  Next up, we show that the base-stock policies satisfy a sample-path information order, where $\pi_{\theta'} \preceq \pi_\theta$ whenever $\theta' \leq \theta$.  We further note that computing the counterfactual estimates for $G_{\theta'}(\mathcal{H}_{\theta'}^{T})$ from $\H_{\theta}^T$ can be done in $O(T)$ time (as exemplified in the proof of the result).

\begin{restatable}{lemma}{BaseStockInformationOrder} 
\label{lem:base_stock_policy_orders}
The set of base stock policies satisfies the sample-path information order of width one, where $\pi_{\theta'} \preceq \pi_\theta$ whenever $\theta' \leq \theta$.
\end{restatable}

\subsubsection{Performance Guarantee}

Combining \cref{lem:base_stock_policy_orders} with the fact that the resulting problem satisfies \Cref{assumption_continuity,assumption_MRP,assumption_restart} establishes:
\begin{restatable}{corollary}{BaseStockRegretCor}
\label{cor:base_stock_regret}
Applying \cref{alg_policy_elimination} to the inventory control with positive lead time $L$ yields that $\Regret(T) = \tilde{O}\left(L\max\{h,p\}U\sqrt{\log(UT^{3/2})T}\right)$.
\end{restatable}
\Cref{cor:base_stock_regret} recovers the regret bound of \citet{agrawal2019learning}; see Theorem~1 and Theorem~3 therein. Note that the absence of a \(\sqrt{\log(U)}\) term in their Theorem~1 is because it is absorbed into the \(\tilde{O}(\cdot)\) notation, whereas our bound states this dependence explicitly. Moreover, our result does not require the long-run average cost \(g_\theta\) to be convex, and therefore also applies to nonconvex cost structures such as the stepwise ordering costs considered in \citet{chen2021eoq}.

\begin{rproof}
In order to apply \cref{thm:regret_bound}, it suffices to show that the set of base stock policies satisfies \Cref{assumption_continuity,assumption_restart,assumption_MRP}, and to provide values for $\diam$, $D_\Theta$, $H$, $L_\Theta$, $\alpha$, $w$ ,$d$, and $\Cupperbound$.  Using Lemma C.1 in \citet{agrawal2019learning} we know $g_\theta$ is Lipschitz with $L_\Theta = \max\{h,p\}$.  Similarly, \Cref{assumption_continuity,assumption_MRP} are satisfied under $\diam = U$ and $H = 36 \max\{h,p\}LU$ (Lemma 2.8 of \citet{agrawal2019learning}). We also have $\Cupperbound = \max\{h,p\}U$. \Cref{assumption_restart} holds under ``restart'' policy $\theta_R = 0$ with $D_\Theta = U \Ex[\tau] + L$ where $\tau = \inf\{t \geq 0 : \sum_{t'=1}^t D_{t'} \geq 1\}$. Regardless of the state before implementing $\theta_R = 0$, it takes at most $L$ periods to clear the inventory pipeline, and the on-hand inventory is bounded by $U$ since $\Theta = [0, U]$. 
The term $U \Ex[\tau]$ follows since each unit of on-hand inventory requires at most $\Ex[\tau]$ periods to deplete in expectation. 
Together, these imply that the system returns to $s_1 = (0, \ldots, 0)$ within $D_\Theta$ periods in expectation.
We have $\alpha = 1$ by the sample-path information order established in \Cref{lem:base_stock_policy_orders}. We have $d = 1$ since $\theta \in [0, U]$. Furthermore, the information order has width $w = 1$, as it is fully determined by the natural order on $\mathbb{R}_{\geq 0}$.
\end{rproof}

\begin{remark}
\label{remark_induced_state_space}
A natural concern one may have is that the ambient state space \(\S=[0,(L+1)U]^{L+1}\) may contain states with very large inventory levels, such as
\(
s'=((L+1)U,\ldots,(L+1)U)\in\S .
\)
If the bias span were taken over the entire ambient space, such states could lead to a large span of bias, since \(v_\theta(s')\) would reflect the transient cost of draining excess inventory. This is not an issue under \Cref{assumption_MRP}, because \(H\) is defined only on the policy-induced support set \(\S_\theta\), and \(s'\notin\S_\theta\) for any $\theta \in \Theta = [0, U]$.
\end{remark}

\subsection{Dual Index Policy for the Dual Sourcing Problem}\label{case_study:dual_index_policy}
Dual sourcing extends the single-channel inventory control model in \Cref{case_study:inventory} to a setting with two replenishment channels: a regular channel with lower cost and longer lead time, and an expedited channel with higher cost and shorter lead time. The optimal control of the dual sourcing problem with lost-sales is notoriously challenging~\citep{xin2023dual}. Consequently, numerous heuristic policies have been proposed in the literature, such as dual index policies and tailored base-surge policies~\citep{xin2023dual}. In this section, we focus on the dual index policies of \citet{veeraraghavan2008now} for two reasons. First, the superior performance of dual index policies in dual‐sourcing is well‐documented~\citep{li2014multimodularity, hua2015structural}, making them a preferred approach in practice. Second, despite their popularity, providing regret guarantees for dual index policies in the lost‐sales context remains an open problem in the literature.

\subsubsection{Model}

The system dynamics mirror those in \Cref{case_study:inventory} with two key extensions: (i) the retailer places two orders $(\Order^r_t, \Order^e_t)$ from regular and expedited channels with lead times $L_r > L_e$, and (ii) outstanding orders from both channels are tracked separately.

At time $t$, the retailer observes the current on-hand inventory $\Inventory_t$, along with the pending regular orders $(\Order^r_{t-L_r}, \ldots, \Order^r_{t-1})$ and expedited orders $(\Order^e_{t-L_e}, \ldots, \Order^e_{t-1})$. The retailer then selects actions $(\Order^r_t, \Order^e_t)$ and fulfills demand $D_t$ using available inventory. Let the sales be $N_t = \min\{\Inventory_t + \Order^r_{t-L_r} + \Order^e_{t-L_e}, D_t\}$. Only sales $N_t$ are observed instead of the true demand $D_t$.
The full state is defined as
\[
S_t = (I_t, \Order^r_{t-L_r}, \ldots, \Order^r_{t-1}, \Order^e_{t-L_e}, \ldots, \Order^e_{t-1}) \in [0, (L_r+L_e+1)U]^{L_r + L_e + 1},
\]
and the action space is $\A = [0, U]^2$. The system evolves according to the following transition and cost dynamics:
\begin{align*}
    \Inventory_{t+1} &= \left( \Inventory_t + \Order^r_{t-L_r} + \Order^e_{t-L_e} - D_t \right)^+, \\
    C_t(S_t) &= h \left( \Inventory_t + \Order^r_{t-L_r} + \Order^e_{t-L_e} - D_t \right)^+ + p(D_t - \Inventory_t - \Order^r_{t-L_r} - \Order^e_{t-L_e})^+ + c_r \Order^r_t + c_e \Order^e_t,
\end{align*}
% \srscomment{repeat here in writing out the cost used for the evaluation in the table. Make sure to switch to $b,h$ for the overage and underage costs, to align with earlier discussions (sorry, we will need to update the appendix accordingly with this).  Guess the alternative is to keep $(h,p)$ but then rewrite the newsvendor example in the introduction to use that}
where $h$ and $p$ are the holding and lost-sales cost coefficients, and $c_r$ and $c_e$ are the unit costs for regular and expedited orders. The next state $S_{t+1}$ is obtained by shifting the regular and expedited pipelines one step forward, appending the newly placed orders, and updating the inventory level:
\[
S_{t+1} = \left( \Inventory_{t+1}, \Order^r_{t-L_r+1}, \ldots, \Order^r_t, \Order^e_{t-L_e+1}, \ldots, \Order^e_t \right).
\]
We assume a fixed initial state $s_1 = (0, \ldots, 0)$. 
Again, we emphasize that all model variables are continuous, including the states, the actions, policies, and demand.

\myparagraph{Modifications to Cost Function.} As in \cref{case_study:inventory}, $C_t(S_t)$ is not observed since it depends on the realized demand $D_t$.  However, one can again transform the cost to a pseudo-cost maintaining the same average cost via:
\begin{align}
\tilde{C}(S_t) = C(S_t) - pD_t. \label{eq_pseudo_cost_dual_sourcing}
\end{align}
Our results in this section still apply regardless of the choice of cost function, as in~\cref{case_study:inventory}.  When indicating the performance of the algorithms we measure the true cost $C(S_t)$ instead of the pseudocost $\tilde{C}$. 
Moreover,
\(
\Cupperbound = (\max\{h,p\} + c_r + c_e)U
\)
is a valid upper bound on the absolute value of the one-step cost.

\subsubsection{Policies and Information Order}
We consider a class of heuristic policies known as dual index policies~\citep{veeraraghavan2008now}, which are known to perform well in practice despite the complexity of the optimal policy~\citep{janakiraman2017dual}. Each policy $\theta = (z_e^\theta, z_r^\theta) \in \Theta \subset \mathbb{R}^2$ is parameterized by two base-stock levels: $z_e^\theta$ for the expedited channel and $z_r^\theta$ for the total inventory position. At each time step $t$, the expedited and regular orders are computed sequentially as:
\begin{align*}
    \Order^e_t &= \left( z_e^\theta - \Inventory_t - \sum_{i=1}^{L_e} \Order^e_{t-i} - \sum_{i=L_r-L_e}^{L_r} \Order^r_{t-i} \right)^+ \\
    \Order^r_t &= \left( z_r^\theta - \Inventory_t - \sum_{i=1}^{L_e} \Order^e_{t-i} - \sum_{i=1}^{L_r} \Order^r_{t-i} - \Order^e_t \right)^+.
\end{align*}
The policy first brings the {\em expedited inventory position} (on-hand inventory plus expedited pipeline and truncated regular pipeline) up to $z_e^\theta$ (if possible), and then tops off the total inventory position to $z_r^\theta$ using regular orders. 
%This sequential adjustment structure allows for tractable control while capturing key features of the dual-sourcing setting.
%
%
Although dual index policies do not satisfy a sample-path order, a distributional information order holds. 

More concretely, we construct our counterfactual estimator $\CGbarprime$ based on the following key insight: \textit{regardless of which policy $\theta$ is executed}, it is always possible to obtain an estimate of $g_{\theta'}$ for any $\theta' \in \Theta$, provided that enough uncensored demand samples are available. This follows from the fact that demand realizations are independent of both the state and the policy, together with the fact that observed sales are capped by the on-hand inventory {(after order arrivals)} at each step (which is policy dependent). Thus any demand sample censored under a higher inventory level remains valid for estimating counterfactual sales under any lower inventory level.

As a result, for any pair of policies $\theta$ and $\theta'$ with $z_r^{\theta'} \le z_r^{\theta}$, the segments of the trajectory $\H_{\theta}^T$ where the on-hand inventory equals $z_r^{\theta}$ (after order arrivals) can be used to construct an estimate of $g_{\theta'}$, since under $\theta'$, the inventory level never exceeds $z_r^{\theta'} \le z_r^{\theta}$.
Moreover, trajectories induced by $\theta$ revisit high-inventory states with constant probability. This ensures that, with high probability, a sufficient number of usable samples are collected under $\H_{\theta}^T$ to accurately estimate $g_{\theta'}$ for all $\theta'$ such that $z_r^{\theta'} \le z_r^{\theta}$. We formalize this idea in what follows, beginning with the definition of the estimator $\CGbarprime$.

\begin{restatable}{definition}{DefDipEstimator}
\label{def_dip_ERM}
Consider a fixed \(\alpha\in(0,1]\) and a trajectory \(\H_{\theta}^T\) under policy \(\theta\).  Define the hitting times
\[
\tau_i = \inf\{t> \tau_{i-1} : I_t^\theta + \Order^{\theta, r}_{t-L_r} + \Order^{\theta, e}_{t-L_e} = z_r^\theta\},  \quad \tau_0=0,
\]
where the superscript $\theta$ denotes quantities under policy $\pi_\theta$,
and let \(\mathcal I_\theta = \{\tau_1,\tau_2,\dots\}\). Denote \(\tilde D_t\) as the sales observed at time \(t\). For any \(\theta'\in\Theta\), if \(\lvert\mathcal I_\theta\rvert\ge\alpha T\), we define the counterfactual trajectory \(\tilde\H_{\theta'}^{\alpha T}\) by simulating policy \(\theta'\) from the initial state \(s_1\) assuming the true demand sequence is \(\{\tilde D_{\tau_i}\}_{i=1}^{\alpha T}\). Then the counterfactual estimate $\CGbarprime(\H_{\theta}^T)$ is defined as:
\begin{equation}
    \CGbarprime(\H_{\theta}^T) = 
    \begin{cases} 
    G_{\theta'}(\tilde{\H}_{\theta'}^{\alpha T}), & \text{if } \left|\mathcal{I}_\theta\right| \ge \alpha T, \\
    0, & \text{otherwise.}
    \end{cases}
    \label{eq_estimator_dual_index}
    \end{equation}
\end{restatable}

Note that the dual‐sourcing problem constitutes an exogenous MDP (Exo‐MDP): demand realizations are independent of the MDP’s actions and states~\citep{sinclair2023hindsight,wan2024exploiting}. This exogeneity directly enables counterfactual estimation, allowing us to infer the performance of one policy using data collected under another. Finally, we have:

\begin{restatable}{lemma}{DualIndexOrder}
\label{lem:dual_index_partial_orders}
Set $\gamma = \Pr(D_t = 0)$ and let $\alpha = \frac{1 - \gamma}{2} \gamma^{L_r}$. Fix any policy $\theta \in \Theta$, confidence level $\delta > 0$, and time horizon $T \ge \Th(\delta)$, where $\Th(\delta) = \Theta(\frac{\log(1 / \delta)}{(1 - \gamma)^4 \gamma^{4L_r}})$. Let $\CGbarprime$ be as specified in \Cref{def_dip_ERM}. Then, for any policy $\theta' \in \Theta$ such that
\(
    z^{\theta'}_r \le z^{\theta}_r,
\)
it follows that
\(
    \pi_{\theta'} \preceq \pi_\theta.
\)
\end{restatable}

\subsubsection{Performance Guarantee}

We now state our regret bound for dual index policies in the lost‐sales dual‐sourcing problem.  Notably, (i) the lost‐sales dual index policy setting lacks convexity~\citep{veeraraghavan2008now}, and (ii) we achieve a $\sqrt{T}$ scaling regret bound, improving on the $T^{2/3}$ dependence in \citet{dann2020reinforcement}.  To our knowledge, this is the first \(\sqrt T\)‐regret result for dual index policies under lost‐sales. Related work includes \citet{tang2024online}, who obtain a $\sqrt{T}$ scaling regret bound for dual index policies in a backlog model, and \citet{chen2024tailored}, who achieve $\sqrt{T}$-type regret for tailored base‐surge policies which rely on convexity.

The proof follows the same structure as that in \Cref{case_study:inventory}. We first verify \Cref{assumption_continuity,assumption_restart,assumption_MRP}:

\begin{restatable}{lemma}{DipGainSpan}
\label{lem:dual_index_bias}
For each dual index policy $\theta$ in \Cref{case_study:dual_index_policy}, the gain is uniform, and the span of the bias is upper bounded by $\frac{2\Cupperbound}{(1 - \gamma) \gamma^{L_r}}$ on the policy-induced support set $\S_{\theta}$.
\end{restatable}

The exponential dependence on \(L_r\) for the span $H$ matches the existing reinforcement learning literature~\citep{agrawal2019learning}, but it remains open whether one can obtain \(H=O(L_r)\) in the dual‐sourcing setting.  While \citet{agrawal2019learning} shows \(H=O(L_r)\) for the single‐channel case, no such linear bound is known here.  Although our policy‐order proof incurs \(H=O((\gamma^{L_r})^{-1})\), empirical evidence suggests this exponential factor may be avoidable with a sharper argument, see~\Cref{appendix_case_study_dual_index_empirical_evidence}. We leave this refinement to future work.

In order to apply \cref{thm:regret_bound} it suffices to show that $g_\theta$ is Lipschitz continuous in the base-stock levels $\theta$.  However, this property has not been studied in the literature.  Here we present a proof for the case of ``slow-moving'' items, i.e. that the demand in each period is zero with high probability~\citep{hahn2015managing}. This assumption is realistic, since slow-moving products account for the bulk of modern inventories~\citep{snyder2012forecasting,bi2023taming}.

\begin{restatable}{lemma}{DipLipschitz}
\label{case_study_dual_index_policy_lem1} \label{dip_lem1}
    For any dual sourcing system, there exists $c_\gamma \in (0,1)$ such that if $\gamma > c_\gamma$, the cost function $g_\theta$ is Lipschitz continuous with respect to $\theta$.
\end{restatable}

% \srscomment{This is the only place you refer to slow moving items. State something along the liens of ``Lipschitz properties of this policy class has not been studied before in the literature. Here we show that the average cost function is Lipschitz under a slow moving item assumption. Motivate (you have some text that does that earlier). Empirically, we believe that this is not required, and our results hold so long as the cost function is Lipschitz.}

The constant $c_\gamma$ can be computed explicitly, see  \Cref{remark_dip_lipschitz} for further details and an empirical evaluation.  We observe that dual index policies remain Lipschitz empirically even without this ``slow-moving'' assumption, see~\Cref{appendix_case_study_dual_index_empirical_evidence}.  However, we leave a thorough treatment of the Lipschitz property for future work.

Combining the results above yields the desired regret bound from \Cref{thm:regret_bound}:
\begin{restatable}{corollary}{RegretDualIndexPolicy}
\label{cor:dual_index_regret}
For any dual sourcing system with fixed $(L_r, L_e)$, if $g_\theta$ is Lipschitz continuous with respect to $\theta$, our algorithm \ALG yields that $\Regret(T) =  \tilde{O}(\gamma^{-\frac{3L_r}{2}}(1 - \gamma)^{-\frac{3}{2}}(\max\{h,p\} + c_r + c_e)U \sqrt{T})$.
\end{restatable}
\begin{rproof}
First, note the bound on the bias is given in \cref{lem:dual_index_bias}.  The final regret bound follows from \Cref{lem:dual_index_partial_orders} by taking $d = 2, w = 1, \alpha = \frac{1 - \gamma}{2} \gamma^{L_r}$, and $\Cupperbound = (\max\{h,p\} + c_r + c_e)U$.  
\end{rproof}

Although the preceding discussion suggests that the \(\gamma^{-L_r}\)-type constant may be avoidable, one may still be concerned about the burn-in period and the practical applicability of~\Cref{alg_policy_elimination} to the dual-index model. We therefore provide a carefully designed empirical study to illustrate that~\Cref{alg_policy_elimination} can indeed be practical; see~\Cref{appendix_case_study_dual_index_empirical_evidence}.
The same subtlety as in \Cref{remark_induced_state_space} arises here and is handled identically. In particular, \Cref{lem:dual_index_bias} is applied on the policy-induced support set \(\S_\theta\), so states with large inventory outside \(\S_\theta\) do not affect the bound on $H$. We omit the repeated argument.
Note that the \(\sqrt{T}\)-regret bound of Corollary~\ref{cor:dual_index_regret} holds for \emph{lost‐sales} dual index policies, which are strictly more challenging than the \emph{backlog} dual index policies studied in \citet{tang2024online}.  In the backlog case, demand is fully observed at every time step, enabling counterfactual estimation under any policy, which is the basis of the algorithm introduced in \citet{tang2024online}.

We also remark that when \(L_r - L_e = 1\), one can obtain a two‐dimensional sample‐path information order over the dual index policy class $\Theta$ with width \(w\) scaling with \(1/r\) (thus \Cref{thm:regret_bound} fails as $w$ is no longer a constant), leading to a \(T^{2/3}\) scaling regret—worse than the $\sqrt{T}$ dependence in Corollary~\ref{cor:dual_index_regret}.
The details are provided in~\Cref{appendix_case_study_dual_index_sample_path_order}.
This shows that even when both sample‐path and distributional orders are available, the distributional order can yield superior \(T\)‐dependence (\(\sqrt{T}\) vs.\ \(T^{2/3}\)), despite having constants exponential in \(L_r\).

\subsection{M/M/1/L Queuing Model with Service Rate Control}\label{case_study:queuing}
In our third case study, we consider an M/M/1/L queueing system with impatient customers and controllable service rates~\citep{walton2021learning}.  Despite the simplicity of the model (featuring only two unknown parameters: arrival and deadline rates), existing learning algorithms suffer from sample complexity that grows with the size of the action space. However, while the queue length is always inherently bounded, the action space may be large when we allow the service rate to take many possible values. We show that, by constructing a suitable distributional information order, our framework yields a regret bound independent of the size of the action space in this setting. Additionally, this problem has no {\em partial feedback}, as required in \citet{dann2020reinforcement}.  % We will see that by leveraging estimates of the arrival and service rates directly we can achieve sublinear regret rates. % highlighting the importance of taking into account potentially biased estimates to achieve sublinear regret rates. \srscomment{Confused what you mean by biased? I guess just importance of using the distributional order? No sample-path feedback structure?}\zzcomment{Sorry for the confusion. Here what I intended to say is that, the estimate $\hat{\lambda}$ and $\hat{\mu}$ may not be unbiased estimate of $\lambda$ and $\mu$. But they are consistent estimators thus we still have the regret bound. This slightly differs from the sample-average estimators in Definition 3 as these sample-average estimators are always unbiased.}

\subsubsection{Model}

We consider an M/M/1/L queuing system in which the decision-maker dynamically selects a service rate from a finite set of options.  Note that these models have also been proposed to represent a dynamic voltage and frequency scaling processor control~\citep{anselmi2021optimal}.  Jobs arrive to a finite buffer of size $L$ according to a Poisson process of (unknown) rate $\lambda$.  Upon each arrival of a job, they draw an unobserved deadline from an exponential distribution with (unknown) rate $\mu$.  Each job requires exactly one unit of processing work.

When a job arrives to the system and the queue is full (i.e. the total number of jobs in the system is $L$), the job is lost.  Otherwise, it enters the queue and waits for service.  However, if a job's deadline elapses before the service completes, it departs immediately and the algorithm incurs a fixed penalty $C$.  We assume that the queue starts at time $t = 1$ and is initially empty.

To control the queue, the algorithm may choose at each state (total number of customers in the system), a processing speed \(a\in \A \subset [0,A_{\max}]\) where $|\A| < \infty$.  Operating at speed \(a\) incurs power cost \(w(a)\), where \(w(\cdot)\) is an arbitrary convex cost function. While the assumption that $w(\cdot)$ is convex aligns with \citet{anselmi2022reinforcement}, our results apply to arbitrary bounded cost functions.  The goal of the controller is to trade off running faster (higher $w(a)$) against letting jobs miss their deadlines (each incurring a cost $C$).

\myparagraph{MDP Formulation.} In order to analyze this problem using a discrete time formulation, we apply the uniformization trick with a constant $U = \lambda_{\max} + L \mu_{\max} + A_{\max}$, where $\lambda_{\max}$ and $\mu_{\max}$ are known upper bounds on the arrival and deadline rates~\citep{anselmi2022reinforcement}.  We construct an MDP $\M = (\S,\A,P,C,s_1)$ as follows.

The state space for the MDP is $\S = \{0, \ldots, L\}$ corresponding to the total number of jobs in the system. The action space is $\A$.  Taking action $a$ in state $s$, the system transitions to a new state via:
\begin{align*}
    P(s' \mid s,a) = \begin{cases}
        \frac{\lambda_s}{U} & s' = s + 1, s < L, \\
        \frac{s \mu + a}{U} & s' = s - 1, s > 0, \\
        1 - \frac{\lambda_s \Ind{s < L} + (s \mu + a)\Ind{s > 0}}{U} & s' = s,
    \end{cases}
\end{align*}

where \(\lambda_s = \lambda\bigl(1 - \tfrac{s}{L}\bigr)\) is the state‐dependent decaying arrival rate~\citep{anselmi2022reinforcement}.  We adopt this decaying rate, in which \(\lambda_s\) represents the true arrival rate at state \(s \in \S = \{0, \ldots, L\}\) from \citet{anselmi2022reinforcement} to enable a fair comparison, but our results do not rely on this strong assumption (see~\cref{remark_queuing}).
Further discussion regarding such decaying rate is in Remark~\ref{remark_queuing}.

The cost $C(s,a)$ in state $s$ having taken action $a$ is a random variable defined as:
\begin{align*}
C(s,a) =
\begin{cases}
    \frac{w(a)}{U} + C, &\text{with probability } \frac{s\mu}{U},\\[8pt]
    \frac{w(a)}{U}, & \text{otherwise.}
\end{cases}
\end{align*}
This corresponds to paying $w(a)$ for servicing jobs at a rate $a$, with a potential cost of $C$ if the current job's deadline is before the service is complete. 

\subsubsection{Policies and Information Order}

Since the state and action spaces are finite, we consider the class of all {\em deterministic} policies, i.e. $\Theta = \A^{L+1}$.~\footnote{Note that \(\Theta\) is discrete in the queuing case study, so the discretization step in~\Cref{alg_policy_elimination} is unnecessary. The remainder of~\Cref{alg_policy_elimination} applies directly.}  Each policy $\theta \in \Theta$ is of the form
\begin{equation}
\label{eq:policy_mm1l}
\pi_\theta(s) = \theta_s,
\end{equation}
corresponding to the service speed $\theta_s \in \A$ employed when the queue has $s$ jobs in the system.  See \citet{anselmi2022reinforcement} for more discussion on properties of the optimal policy.

Note that under mild stability conditions, the controlled Markov chain admits a unique stationary distribution $m_\theta \in \Delta(\S)$ for any policy $\theta \in \Theta$.  Moreover, the long-run average cost of $\pi_\theta$ can be written as:
\begin{equation}
\label{eq:opt_cost_queue}
g_\theta
=\sum_{s=0}^L m_\theta(s)\Exp{C(s,\pi_\theta(s))}.
\end{equation}

\myparagraph{Construction of the Information Order.}
Note that the MDP has only two unknown parameters: $\lambda$ and $\mu$.  Hence, no matter the implemented policy $\theta$, if we can estimate $\lambda$ and $\mu$ accurately, we can counterfactually estimate the performance of any other policy $\theta' \in \Theta$.

Indeed, we start by showing how to use $\H_{\theta}^T$ to estimate $\lambda$ and $\mu$ directly, regardless of the policy $\theta$. With this, we construct the stationary distribution $\hat{m}_{\theta'}$ for an arbitrary policy $\theta'$ as its stationary distribution when the arrival and deadline rates are the estimates, $\hat{\lambda}$ and $\hat{\mu}$, accordingly. We then estimate its average cost via:
\begin{align}
\CGbarprime(\mathcal{H}_{\theta}^T) = \sum_{s=0}^L \hat{m}_{\theta'}(s) \Exp{\hat{C}(s, \pi_{\theta'}(s))},
\label{eq_queuing_estimator}
\end{align}
where $\hat{C}$ differs from $C$ in that $\mu$ is replaced by $\hat{\mu}$.
Hence, in order to formally define $\CGbarprime(\H_{\theta}^T)$ we just need to define how we estimate $\lambda$ and $\mu$, and in turn calculate $\hat{m}_{\theta'}$. Note that the stationary measures of the CTMC and the uniformized discrete chain are the same, thus we don't distinguish them below.

\myparagraph{Estimating $\lambda$ and $\mu$.} Given a trajectory $\H_{\theta}^T$ collected under policy $\pi_\theta$, we denote $\tau_{\theta, i}$ as the time of the $i$-th ``jump'' of the process. Hence, the amount of time spent in state $s_{\tau_{\theta, i}}$ is $\tau_{\theta, i+1} - \tau_{\theta, i}$. Note that although the chain evolves in discrete time, \(\tau_{\theta, i}\) and \(i\) may not coincide, as the chain can remain in the same state for multiple consecutive time steps.  We also denote $\tau_{\theta, 0} = 1$ and $N_\theta(T) = \max\{ n \mid \tau_{\theta, n} \leq T\}$ to refer to the total number of jumps by time $T$.

We first consider estimating $\lambda$.  Note that whenever the system is in state $s = 0$, it leaves only by an arrival of a new job (which occurs with rate $\lambda$).  Hence we can estimate $\lambda$ via
\begin{equation}
\label{eq:lambda_hat}
\hat{\lambda} = \frac{\sum_{i=0}^{N_\theta(T) -1 }\Ind{s_{\tau_{\theta, i}}=0}}{\sum_{i=0}^{N_\theta(T) -1 }(\tau_{\theta, i + 1} - \tau_{\theta, i})\Ind{s_{\tau_{\theta, i}}=0}}\cdot U
\end{equation}
as the inverse of the average time it takes to depart state $s = 0$.
Similarly, when the system is in state $s = L$, it leaves only by a job departure at rate $L\mu + \theta_L$. Thus we estimate $\mu$ via
\[
\hat{\mu} = \frac{1}{L}\left(\frac{\sum_{i=0}^{N_\theta(T) -1 }\Ind{s_{\tau_{\theta, i}}=L}}{\sum_{i=0}^{N_\theta(T) -1 }(\tau_{\theta, i + 1} - \tau_{\theta, i})\Ind{s_{\tau_{\theta, i}}=L}}\cdot U - \theta_L\right).
\]
There are various other estimators for $\lambda$ and $\mu$.  For example, treating arrivals and departures as Bernoulli events, or using data from states beyond just $s = L$ in order to estimate $\mu$, may be helpful. It can be shown that this approach (at least) yields an improved $\tilde{O}(L^2 \sqrt{T})$ regret. However, since our primary aim is to demonstrate that our algorithm achieves regret that does not scale polynomially with \(|\A|\), we employ the simple estimators above for expository purposes.

\myparagraph{Calculating $\hat{m}_{\theta'}(s)$.} Since the state space is finite, the Markov chain is positive recurrent.  Therefore, under mild regularity conditions, for any values of $\hat{\lambda}$ and $\hat{\mu}$, one can calculate the stationary distribution $\hat{m}_{\theta'}(s)$.  This can be done, for instance, by calculating the generator matrix $Q$ for the underlying process and solving the stationary equations (see \citet{puterman2014markov}).

\begin{remark}
With uniformization, the system becomes a standard discrete time Markov chain, on which our algorithm operates. The key estimates, \(\hat{\lambda}\) and \(\hat{\mu}\), are computed directly from running the algorithm over the discrete chain. References to the continuous time process in this section and the appendix are included only to simplify certain derivations and proofs.
\end{remark}

Based on the previous construction, we are ready to formally state the distributional information order.
\begin{restatable}{lemma}{QueuingEstimator}
\label{queuing_dtv}
For any $T$ sufficiently large and any two policies $\theta$ and $\theta'$, there exists $\CGbarprime$ such that with probability at least $1 - 4\delta$,
\[
|\CGbarprime(\mathcal{H}_{\theta}^T) - g_{\theta'}| \le \tilde{O}\left( L^2\sqrt{\frac{1}{m_{\theta}(L)}} \sqrt{\frac{\log(1 / \delta)}{T}}\right),
\]
where \(\tilde{O}(\cdot)\) hides logarithmic factors in \(L\), as well as constants depending on several parameters, including $A_{\max}, w(A_{\max}), C, \lambda_{\max}, \mu_{\max}, \lambda, \mu$, and a lower bound of the spectral gap (see~\Cref{lemma_spectral_gap_1}).
As a result, $\pi_{\theta'} \preceq \pi_{\theta}$.
\end{restatable}

The explicit form of the constant hidden in \(\tilde{O}(\cdot)\) is given in \Cref{queuing_policy_order_definition}; see \cref{eq_queuing_proof_7}.
The information order \(\pi_{\theta'} \preceq \pi_{\theta}\) follows \Cref{definition_general_order} but is slightly modified since we estimate \(g_{\theta'}\) directly rather than via empirical average cost. Details are also deferred to \Cref{queuing_policy_order_definition}, as this change relies on the proof of \Cref{thm:regret_bound} in the Appendix.

% \srscomment{Distinction between $T$ vs $\sqrt{T}$?}
%Note the the difference between \Cref{queuing_dtv} and \Cref{definition_general_order} is that, in this case study we are allowed to estimate $g_{\theta'}$ directly without estimating the empirical estimator of $\theta'$, i.e., $\Gbarprime(\H_{\theta'}^{\alpha T})$ for some $\alpha < 1$. This is also where the high probability term $1 - \delta$ term comes from.

\subsubsection{Performance Guarantee}

Combining \cref{queuing_dtv} with the fact that the resulting problem satisfies \Cref{assumption_continuity,assumption_MRP,assumption_restart} establishes:
\begin{restatable}{corollary}{RegretMML}
\label{cor:queuing_regret}
Assume $m_{\theta}(L) = \Omega(1/L)$ for all $\theta \in \Theta$.
For the M/M/1/L model, \ALG yields that $\Regret(T) = \tilde{O}\bigl(L^3\sqrt{\log(|\A|)T}\bigr)$.
\end{restatable}

\begin{rproof}
First, \Cref{queuing_dtv}, together with a union bound, can replace~\Cref{main_lem1}, yielding an effective parameter \(H=\tilde{O}(L^{5/2})\) after accounting for the factor \(\sqrt{1/m_{\theta}(L)}\).
Since the state and action spaces are finite, \Cref{assumption_continuity} and \Cref{assumption_restart} hold trivially.
Lastly, since $|\Theta| = |\A|^{L+1}$ we have that $\log(|\Theta|) = (L+1)\log(|\A|)$.
It follows directly from \Cref{queuing_dtv} that \(w=1\).
Combining this with \Cref{thm:regret_bound} gives the desired result.
\end{rproof}

We note that the assumption \(m_{\theta}(L)=\Omega(1/L)\) for all \(\theta\in\Theta\) need not hold in general; in extreme cases, \(m_{\theta}(L)\) may be exponentially small. This assumption enters only through the counting process concentration in \cref{eq_counting_2}, and arises from our particular choice of estimator for \(\mu\), which uses only holding intervals at state \(s=L\). We make this choice for expositional simplicity: conditional on visits to \(s=L\), the holding intervals are geometric, and the analysis in \Cref{queuing_policy_order_definition} is standard. As noted above, one can instead estimate \(\mu\) using downward transitions from multiple other states, which removes this assumption while leaving the rest of the algorithm and analysis unchanged. We omit this extension because it is outside the scope of the paper.

Our regret bound only scales logarithmically with \(|\A|\) (actually, it is independent of $|\A|$, see \Cref{remark_queuing_independence}), achieved by taking into account the information order over policies. In contrast, \citet{anselmi2022reinforcement} derive a regret bound with $\sqrt{T}$ scaling that also contains terms linear in $|\A|$, meaning our result improves upon theirs by a factor of \(\sqrt{|\A|}\), at the expense of an additional factor of \(L^3\). 
This highlights that, for the queuing problem considered here, the comparison is more favorable to our approach when the state space is small but the action space is relatively large, including settings that use a fine discretization to approximate an underlying continuous action space~\citep{baik2023qos}.
Moreover, our results extend seamlessly to more complex settings (for example, with state-dependent arrival rates).

Such \(\sqrt{\log(|\A|)}\) dependence on the action space is reminiscent of expert-advice-style model-free algorithms for infinite-horizon average-cost MDPs, such as~\citet{abbasi2019politex}. However, the mechanism is different: in~\citet{abbasi2019politex}, obtaining such logarithmic action-space dependence also requires controlling the \(Q\)-function estimation error, for which the authors assume linear $Q$-function approximability. Such global linear structure, however, has not been established in the literature for the queuing model considered here. Thus, the information-order argument used here is not merely an instance of an expert-advice-style framework, such as that of~\citet{abbasi2019politex}.

\begin{remark}\label{remark_queuing_independence}
By modifying the proof of \cref{thm:regret_bound} one can avoid a union bound {over $\Theta = \A^{L+1}$}, since we are estimating $\hat{\mu}$ and $\hat{\lambda}$ first before constructing counterfactual estimators $\CGbarprime$.  This modification would yield a regret guarantee of $\tilde{O}(L^\frac{5}{2} \sqrt{T})$, independent of $\log(|\A|)$.  We keep the current version here for expository purposes, since it aligns with the regret proofs for the two other case studies.
\end{remark}

\begin{remark}\label{remark_queuing}
Although we use the decaying arrival rate \(\lambda_s\), this assumption is not necessary for our \(\tilde{O}( L^\frac{5}{2}\sqrt{T})\) regret bound (Remark~\ref{remark_queuing_independence}); our analysis holds for any fixed arrival rate.  In contrast, \citet{anselmi2022reinforcement} rely on \(\lambda_s\) to remove state‐space dependence (Lemma C.2 of \citet{anselmi2022reinforcement}).  Without the decaying‐rate assumption, our algorithm still attains \(\tilde{O}( L^\frac{5}{2}\sqrt{T})\) regret, whereas the regret bound in~\citep{anselmi2022reinforcement} would scale with the sizes of both the state and action space.
\end{remark}

% \srsedit{This $\tilde{O}(L^{3/2}\sqrt{\log(|A|)T})$ rate matches the intuition that the sole uncertainties here are the two Poisson and exponential rates, making learning ``easy.''}  Our bound does incur a factor in \(|S|\), since our approach extends seamlessly to richer settings (e.g., state‐dependent rates).
%
% Moreover, this M/M/1/L case starkly contrasts ``rich‐feedback'' methods that rely on side observations between state–action pairs. Here, seeing data from one $(s,a)$ pair reveals no additional information, so feedback‐graph approaches similar to~\citep{dann2020reinforcement} fail. Our policy‐order view instead exploits distributional comparisons across entire policies, yielding tight regret guarantees without per‐pair side information. 

 \ifdefined\operationsresearch

\begin{table}[!t]
  \centering
  \scriptsize  
  \caption{Average cost of the learned policy at the timestep $g_{\theta_T}$ for $T = 10^5$ (small-scale) and $T = 3\times10^5$ (large-scale) achieved by each algorithm, under three demand distributions. The queuing case study is estimated at $T = 3\times10^5$.
  $\star$ indicates a significant improvement and $\circ$ a significant decrease over {\textsf Problem-Specific} by Welch’s $t$-test with $p<0.05$.  
  In parentheses we report relative performance $(g_{\theta_T}-g_{\theta^*})/g_{\theta^*}$.  
  {\textsf Feedback Graph} is omitted in large-scale due to compute constraints. The performance of \textsf{SGD} is omitted in the queuing models as it is essentially equivalent to \textsf{PPO}. See \Cref{tab:configs-by-block} for hyperparameters.}
  \label{tab_simulation_all}
  \begin{tabular}{@{}c l cc cc c@{}}
    \toprule
    \textbf{Distribution} 
      & \textbf{Algorithm} 
      & \multicolumn{2}{c}{Inventory Control} 
      & \multicolumn{2}{c}{Dual Sourcing} 
      & Queuing \\
    \cmidrule(lr){3-4}\cmidrule(lr){5-6}\cmidrule(lr){7-7}
      & 
      & Small & Large 
      & Small & Large 
      & Large \\
    \midrule

    %—— Exponential block ——
    \multirow{7}{*}{\textbf{Exponential}} 
      & \textsf{Optimal} $(g_{\theta^*})$
        & $2.5$       & $39.1$
        & $1.9$       & $20.3$
        & $9.5$           \\ 
      & \textsf{ERM} \citep{sinclair2023hindsight}
        & $2.6(2\%)$  & $39.2(0\%)$ 
        & $1.9(1\%)$  & $20.3(0\%)$
        & $9.6(1\%)$      \\ 
        
      & \ALG
        & $^{\star}2.6(3\%)$  & $40.1(2\%)$ 
        & $^{\star}2.0(5\%)$   & $^{\star}26.0(28\%)$
        & $^{\star}10.5(11\%)$\\
      & \textsf{PPO}~\citep{schulman2017proximal}
        & $3.8(49\%)$ & $^{\circ}132.4(238\%)$ 
        & $2.9(53\%)$  & $^{\circ}112.0(451\%)$
        & $^{\star}9.8(3\%)$ \\
        & \textsf{SGD}
        & $^{\star}2.6(3\%)$ & $^{\circ}43.0(10\%)$ 
        & $^{\star}2.1(10\%)$  & $^{\circ}55.2(172\%)$
        & -- \\
      & \textsf{Feedback Graph}~\citep{dann2020reinforcement}
        & $^{\circ}5.2(103\%)$ & -- 
        & $^{\star}2.4(26\%)$   & --
        & $11.5(21\%)$       \\
      & \textsf{Problem-Specific}%~\citep{agrawal2019learning,chen2024tailored,anselmi2022reinforcement}
        & $4.1(60\%)$ & $40.3(3\%)$ 
        & $3.4(78\%)$  & $30.1(48\%)$  
        & $11.3(19\%)$      \\  
      & \textsf{Random}
        & $^{\star}3.6(43\%)$   & $^{\circ}182.4(366\%)$ 
        & $3.7(95\%)$           & $^{\circ}121.5(498\%)$
        & $11.1(17\%)$      \\
    \midrule

    %—— Normal block ——
    \multirow{7}{*}{\textbf{Normal}} 
      & \textsf{Optimal} $(g_{\theta^*})$
        & $2.2$       & $39.9$
        & $1.3$       & $17.2$
        & --           \\ 
      & \textsf{ERM} \citep{sinclair2023hindsight}
        & $2.2(1\%)$  & $40.3(1\%)$ 
        & $1.3(3\%)$  & $17.5(2\%)$
        & --      \\ 
      & \ALG
        & $^{\star}2.3(1\%)$  & $^{\star}40.2(1\%)$ 
        & $^{\star}1.3(3\%)$   & $^{\star}17.7(3\%)$
        & --\\
      & \textsf{PPO}~\citep{schulman2017proximal}
        & $^{\circ}3.4(52\%)$ & $^{\circ}133.6(235\%)$ 
        & $^{\circ}2.3(76\%)$  & $^{\circ}148.8(765\%)$
        & -- \\
     & \textsf{SGD}
        & $^{\star}2.3(4\%)$ & $^{\star}41.9(5\%)$ 
        & $1.8(31\%)$  & $^{\circ}45.1(162\%)$
        & -- \\
      & \textsf{Feedback Graph}~\citep{dann2020reinforcement}
        & $^{\circ}5.8(158\%)$ & -- 
        & $1.7(30\%)$   & --
        & --      \\
      & \textsf{Problem-Specific}%~\citep{agrawal2019learning,chen2024tailored,anselmi2022reinforcement}
        & $2.9(29\%)$ & $44.9(13\%)$ 
        & $1.6(22\%)$  & $18.8(9\%)$
        & --      \\  
      & \textsf{Random}
        & $^{\circ}3.4(54\%)$   & $^{\circ}96.5(142\%)$
        & $^{\circ}3.0(127\%)$           & $^{\circ}107.4(524\%)$
        & --      \\
    \midrule

    %—— Uniform block ——
    \multirow{7}{*}{\textbf{Uniform}} 
      & \textsf{Optimal} $(g_{\theta^*})$
        & $2.9$       & $49.0$
        & $1.7$       & $22.9$
        & --           \\ 
      & \textsf{ERM} \citep{sinclair2023hindsight}
        & $3.0(1\%)$  & $49.5(1\%)$ 
        & $1.7(1\%)$  & $23.2(1\%)$
        & --      \\ 
      & \ALG
        & $^{\star}3.0(2\%)$  & $49.5(1\%)$ 
        & $^{\star}1.7(1\%)$   & $^{\star}23.5(3\%)$
        & --\\
      & \textsf{PPO}~\citep{schulman2017proximal}
        & $^{\circ}4.1(41\%)$ & $^{\circ}157.3(221\%)$ 
        & $^{\circ}3.7(116\%)$  & $^{\circ}147.7(545\%)$
        & -- \\
    & \textsf{SGD}
        & $3.1(7\%)$ & $^{\circ}51.5(5\%)$ 
        & $^{\circ}2.7(59\%)$  & $^{\circ}37.5(64\%)$
        & -- \\
      & \textsf{Feedback Graph}~\citep{dann2020reinforcement}
        & $^{\circ}5.8(99\%)$ & -- 
        & $2.6(53\%)$   & --
        & --       \\
      & \textsf{Problem-Specific}%~\citep{agrawal2019learning,chen2024tailored,anselmi2022reinforcement}
        & $3.2(11\%)$ & $49.6(1\%)$ 
        & $2.3(37\%)$  & $30.0(30\%)$  
        & --      \\  
      & \textsf{Random}
        & $^{\circ}3.9(34\%)$   & $^{\circ}136.3(178\%)$ 
        & $^{\circ}3.1(81\%)$           & $^{\circ}108.8(375\%)$
        & --      \\
    \bottomrule
  \end{tabular}
\end{table}

\else 

\fi
\section{Numerical Simulations}
\label{sec:simulations}

% \srscomment{Mention comparing to in-class optimal}
% \srscomment{ERM for empirical hindsight (cite \citep{sinclair2023hindsight})}

Finally, we evaluate \ALG on all three case studies above, showing that it achieves improved performance over state-of-the-art algorithms tailored to each of the case studies, despite being a more general solver. 

\subsection{Baselines and Simulation Results}

We compare the performance of \ALG against several baselines:
% \srscomment{Add a one-sentence discussion to each policy class}
\begin{itemize}
  \item \textsf{Model-Based RL with Feedback Graphs}~\citep{dann2020reinforcement} (hereafter referred to as {\textsf Feedback Graph}): This is a generic model‐based algorithm for finite state‐action MDPs. Unlike our information ordering framework which relies on counterfactual estimation between policies, the {\textsf Feedback Graph} method of \citet{dann2020reinforcement} uses observations from the implemented state–action pair \((s,a)\) to infer counterfactual outcomes for other state-action pairs at each time step. This algorithm is restricted to tabular MDPs. Therefore, when applied to problems with continuous state and action spaces, as in \cref{case_study:inventory,case_study:dual_index_policy}, we discretize the state and action spaces.
  \item \textsf{Proximal Policy Optimization (PPO)}~\citep{schulman2017proximal}: A popular on‐policy gradient method that updates the policy via the clipped surrogate objective.
  \item \textsf{Stochastic Gradient Descent (SGD)}:  An SGD method based on finite‐difference gradient estimates. While prior works in operations research have proposed problem‐specific gradient or subgradient constructions for certain models~\citep{huh2009nonparametric}, we employ finite‐difference estimators for broader applicability.
  \item \textsf{Random Policy}: A baseline that selects actions uniformly at random.
  \item \textsf{Problem-Specific}: Algorithms tailored to each case study.
    \begin{itemize}
      \item Inventory Control: The algorithm of \citet{agrawal2019learning} proposes a convexity‐based policy‐elimination method for the inventory problem (\Cref{case_study:inventory}), maintaining a one‐dimensional confidence interval that shrinks at each epoch (whereas the confidence sets of \ALG need not be intervals). 
      \item Dual Sourcing: The {\textsf BASA} method of \citet{chen2024tailored}, which solves the lost-sales dual‐sourcing problem (\Cref{case_study:dual_index_policy}) over tailored base‐surge policies using stochastic gradient descent and convexity. We note that tailored base-surge policies and dual index policies are two different policy classes.
      \item Queuing Model: {\textsf UCRL2}~\citep{anselmi2022reinforcement}, which addresses the service‐rate control problem (\Cref{case_study:queuing}) by building confidence sets over MDPs and using extended value iteration to select the most optimistic model in each epoch.
    \end{itemize}
  \item \textsf{Optimal In-Class Policy}: The best policy within the given policy class, computed with full knowledge of the system.
  \item \textsf{ERM-Based}: An algorithm adapted from \citet{sinclair2023hindsight}, which leverages full-feedback information. The {\textsf ERM‐Based} algorithm assumes access to unobserved randomness thus is not implementable in practice.
\end{itemize}

Although it lacks theoretical guarantees, we include {\textsf PPO}~\citep{schulman2017proximal} as a representative empirical RL method. See the \href{https://github.com/Zhongjun24/IOPEA_PolicyOrder}{code base} for implementation details.

% See the code \href{https://github.com/Zhongjun24/IOPEA_PolicyOrder}{(https://github.com/Zhongjun24/IOPEA\_PolicyOrder)} for further implementation details. 

\ifdefined\operationsresearch

\else
\begin{table}[!t]
  \centering
  \scriptsize  
  \caption{Average cost of the learned policy at the timestep $g_{\theta_T}$ for $T = 10^5$ (small-scale) and $T = 3\times10^5$ (large-scale) achieved by each algorithm, under three demand distributions. The queuing case study is estimated at $T = 3\times10^5$.
  $\star$ indicates a significant improvement and $\circ$ a significant decrease over {\textsf Problem-Specific} by Welch’s $t$-test with $p<0.05$.  
  In parentheses we report relative performance $(g_{\theta_T}-g_{\theta^*})/g_{\theta^*}$.  
  {\textsf Feedback Graph} is omitted in large-scale due to compute constraints. \zzedit{The performance of \textsf{SGD} is omitted in the queuing models as it is essentially equivalent to \textsf{PPO}.} See \Cref{tab:configs-by-block} for scenario hyperparameters.}
  \label{tab_simulation_all}
  \begin{tabular}{@{}c l cc cc c@{}}
    \toprule
    \textbf{Distribution} 
      & \textbf{Algorithm} 
      & \multicolumn{2}{c}{Inventory Control} 
      & \multicolumn{2}{c}{Dual Sourcing} 
      & Queuing \\
    \cmidrule(lr){3-4}\cmidrule(lr){5-6}\cmidrule(lr){7-7}
      & 
      & Small & Large 
      & Small & Large 
      & Large \\
    \midrule

    %—— Exponential block ——
    \multirow{7}{*}{\textbf{Exponential}} 
      & \textsf{Optimal} $(g_{\theta^*})$
        & $2.5$       & $39.1$
        & $1.9$       & $20.3$
        & $9.5$           \\ 
      & \textsf{ERM} \citep{sinclair2023hindsight}
        & $2.6(2\%)$  & $39.2(0\%)$ 
        & $1.9(1\%)$  & $20.3(0\%)$
        & $9.6(1\%)$      \\ 
        
      & \ALG
        & $^{\star}2.6(3\%)$  & $40.1(2\%)$ 
        & $^{\star}2.0(5\%)$   & $^{\star}26.0(28\%)$
        & $^{\star}10.5(11\%)$\\
      & \textsf{PPO}~\citep{schulman2017proximal}
        & $3.8(49\%)$ & $^{\circ}132.4(238\%)$ 
        & $2.9(53\%)$  & $^{\circ}112.0(451\%)$
        & $^{\star}9.8(3\%)$ \\
        & \textsf{SGD}
        & $^{\star}2.6(3\%)$ & $^{\circ}43.0(10\%)$ 
        & $^{\star}2.1(10\%)$  & $^{\circ}55.2(172\%)$
        & -- \\
      & \textsf{Feedback Graph}~\citep{dann2020reinforcement}
        & $^{\circ}5.2(103\%)$ & -- 
        & $^{\star}2.4(26\%)$   & --
        & $11.5(21\%)$       \\
      & \textsf{Problem-Specific}%~\citep{agrawal2019learning,chen2024tailored,anselmi2022reinforcement}
        & $4.1(60\%)$ & $40.3(3\%)$ 
        & $3.4(78\%)$  & $30.1(48\%)$  
        & $11.3(19\%)$      \\  
      & \textsf{Random}
        & $^{\star}3.6(43\%)$   & $^{\circ}182.4(366\%)$ 
        & $3.7(95\%)$           & $^{\circ}121.5(498\%)$
        & $11.1(17\%)$      \\
    \midrule

    %—— Normal block ——
    \multirow{7}{*}{\textbf{Normal}} 
      & \textsf{Optimal} $(g_{\theta^*})$
        & $2.2$       & $39.9$
        & $1.3$       & $17.2$
        & --           \\ 
      & \textsf{ERM} \citep{sinclair2023hindsight}
        & $2.2(1\%)$  & $40.3(1\%)$ 
        & $1.3(3\%)$  & $17.5(2\%)$
        & --      \\ 
      & \ALG
        & $^{\star}2.3(1\%)$  & $^{\star}40.2(1\%)$ 
        & $^{\star}1.3(3\%)$   & $^{\star}17.7(3\%)$
        & --\\
      & \textsf{PPO}~\citep{schulman2017proximal}
        & $^{\circ}3.4(52\%)$ & $^{\circ}133.6(235\%)$ 
        & $^{\circ}2.3(76\%)$  & $^{\circ}148.8(765\%)$
        & -- \\
     & \textsf{SGD}
        & $^{\star}2.3(4\%)$ & $^{\star}41.9(5\%)$ 
        & $1.8(31\%)$  & $^{\circ}45.1(162\%)$
        & -- \\
      & \textsf{Feedback Graph}~\citep{dann2020reinforcement}
        & $^{\circ}5.8(158\%)$ & -- 
        & $1.7(30\%)$   & --
        & --      \\
      & \textsf{Problem-Specific}%~\citep{agrawal2019learning,chen2024tailored,anselmi2022reinforcement}
        & $2.9(29\%)$ & $44.9(13\%)$ 
        & $1.6(22\%)$  & $18.8(9\%)$
        & --      \\  
      & \textsf{Random}
        & $^{\circ}3.4(54\%)$   & $^{\circ}96.5(142\%)$
        & $^{\circ}3.0(127\%)$           & $^{\circ}107.4(524\%)$
        & --      \\
    \midrule

    %—— Uniform block ——
    \multirow{7}{*}{\textbf{Uniform}} 
      & \textsf{Optimal} $(g_{\theta^*})$
        & $2.9$       & $49.0$
        & $1.7$       & $22.9$
        & --           \\ 
      & \textsf{ERM} \citep{sinclair2023hindsight}
        & $3.0(1\%)$  & $49.5(1\%)$ 
        & $1.7(1\%)$  & $23.2(1\%)$
        & --      \\ 
      & \ALG
        & $^{\star}3.0(2\%)$  & $49.5(1\%)$ 
        & $^{\star}1.7(1\%)$   & $^{\star}23.5(3\%)$
        & --\\
      & \textsf{PPO}~\citep{schulman2017proximal}
        & $^{\circ}4.1(41\%)$ & $^{\circ}157.3(221\%)$ 
        & $^{\circ}3.7(116\%)$  & $^{\circ}147.7(545\%)$
        & -- \\
    & \textsf{SGD}
        & $3.1(7\%)$ & $^{\circ}51.5(5\%)$ 
        & $^{\circ}2.7(59\%)$  & $^{\circ}37.5(64\%)$
        & -- \\
      & \textsf{Feedback Graph}~\citep{dann2020reinforcement}
        & $^{\circ}5.8(99\%)$ & -- 
        & $2.6(53\%)$   & --
        & --       \\
      & \textsf{Problem-Specific}%~\citep{agrawal2019learning,chen2024tailored,anselmi2022reinforcement}
        & $3.2(11\%)$ & $49.6(1\%)$ 
        & $2.3(37\%)$  & $30.0(30\%)$  
        & --      \\  
      & \textsf{Random}
        & $^{\circ}3.9(34\%)$   & $^{\circ}136.3(178\%)$ 
        & $^{\circ}3.1(81\%)$           & $^{\circ}108.8(375\%)$
        & --      \\
    \bottomrule
  \end{tabular}
\end{table}
\fi

\ifdefined\operationsresearch

\begin{figure}[!h]
% \begin{wrapfigure}{r}{0.45\textwidth}  % 'r' = right side, width of the wrapped box
  %\vspace{-1em}                        % tweak vertical placement
  \centering
  \includegraphics[width=0.6\textwidth]{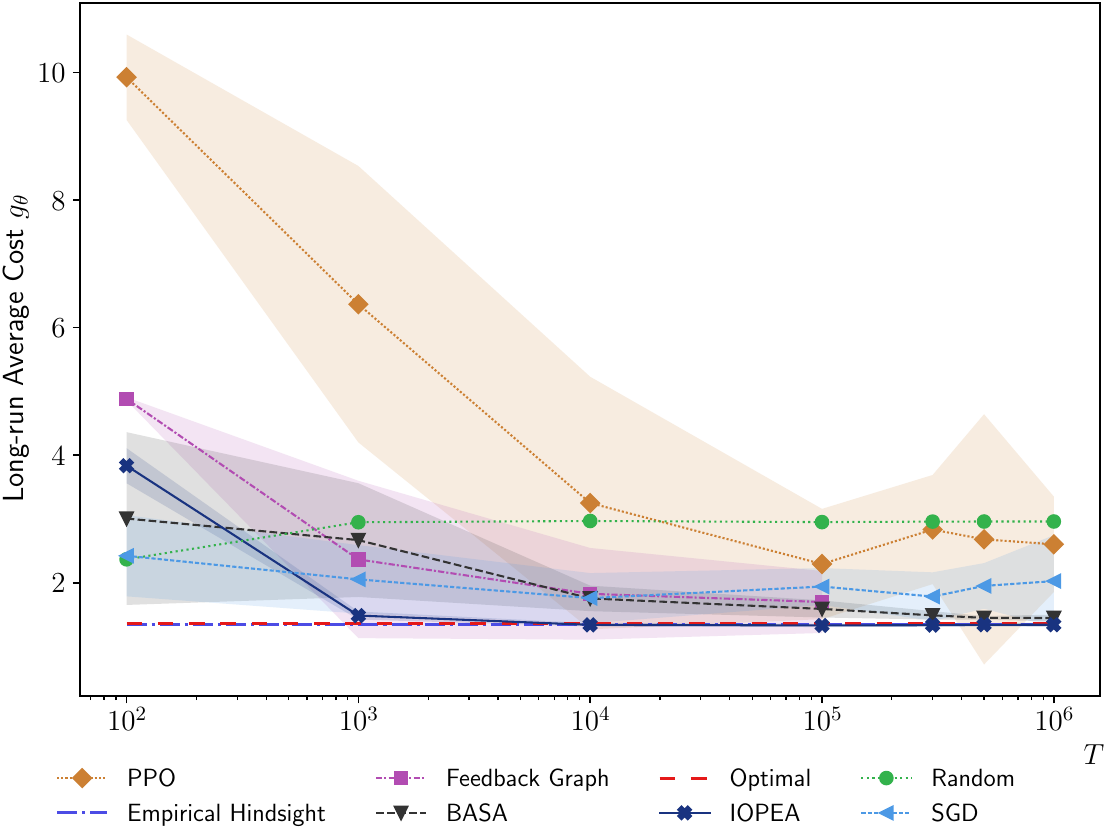}
  \caption{Comparison of the long‐run average cost \(g_{\theta}\) over time \(T\) for different algorithms in the small‐scale dual‐sourcing problem up to \(T=10^6\). We assume the demand follows the normal distribution as in~\Cref{tab:configs-by-block}.  For {\textsf Feedback Graph}~\citep{dann2020reinforcement}, data are shown only to \(T=10^5\). \ALG attains near‐optimal gain by \(T=10^4\); {\textsf BASA}~\citep{chen2024tailored} remains suboptimal and continues to decline even at \(T=10^6\); {\textsf Feedback} Graph matches {\textsf BASA}’s long‐run cost at \(T=10^5\) but with much larger variance; {\textsf PPO} converges markedly slower than all other methods. \textsf{SGD} fails to converge effectively, as the problem is non-convex. Note that in some trials {\textsf PPO} and {\textsf Feedback Graph} may exceed the “Optimal” curve, because “Optimal” denotes the best dual index policy rather than the true problem‐wide optimum.}
  \label{fig_convergence_curve}
  % \vspace{-1em}
% \end{wrapfigure}
\end{figure}

\else 
\fi

\myparagraph{Simulation Results.} In \Cref{tab_simulation_all} we include a comparison of the algorithm performance at $T = 10^5$ (small scale) and $T = 3 \times 10^5$ (large-scale). We use Welch’s $t$‑test, at a significance level of $0.05$, to assess whether each algorithm differs significantly from {\textsf Problem‑Specific} in each setting.  In parentheses we report relative performance $(g_{\theta_T}-g_{\theta^*})/g_{\theta^*}$. Details on the hyperparameters and problem specifications are in \Cref{experiment_settings}. We also provide the numerical experiment results for the queuing case study under the fixed arrival rate in \Cref{appendix_c2}.

First, we observe that \ALG nearly matches the performance of {\textsf ERM-Based} algorithm~\citep{sinclair2023hindsight} despite only having access to partial feedback. For example, in the small-scale inventory control problem, \ALG achieves within 3\% of the optimal cost, while {\textsf PPO} and the {\textsf Problem-Specific} algorithm~\citep{agrawal2019learning} lag behind at 49\% and 60\%, respectively. Also, \ALG delivers stable and competitive performance, consistently matching or outperforming \textsf{Problem-Specific} algorithms. In contrast, the performances of certain baselines are unstable: The convergence behavior of {\textsf Feedback Graph}~\citep{dann2020reinforcement} depends strongly on the size of the state space.  It excels in dual‐sourcing but struggles in inventory control, where the state space is roughly $10$ times larger, and performs even better in the queuing problem with the smallest state space.

Second, \ALG converges substantially faster, indicating superior sample efficiency: by $T = 3\times10^5$ in the large-scale regime, its average cost has stabilized in nearly all simulation settings, whereas all other algorithms require $T > 10^6$ to achieve similar performance. Further evidence of \ALG's rapid convergence is presented in \Cref{fig_convergence_curve}. In the small-scale dual sourcing problem, \ALG achieves near-optimal performance by $T = 10^4$ and remains stable thereafter, converging more quickly than all other algorithms we evaluated. 

An intuitive explanation for the fast convergence of \ALG\ lies in the information order: the rich feedback structure it leverages improves sample efficiency, as observations collected under one policy can be reused to estimate the performance of others, thereby accelerating convergence.  A promising direction for future work is to compare the sample efficiency of low‐width information orders with stochastic gradient descent methods~\citep{ huh2009nonparametric, agrawal2019learning} in convex models; interestingly, in our single-channel inventory experiments (which is convex), \ALG\ even outperforms gradient‐based methods in such settings, despite not leveraging convexity.

Moreover, \ALG is highly scalable. The training of \ALG completes in under three minutes in all simulation experiments, whereas {\textsf PPO} and {\textsf Feedback Graph} are significantly slower or infeasible. This scalability stems from the use of the information order, which enables focused exploration. While computing the information order and associated counterfactual estimates is lightweight during training, it does require domain-specific insight into the structure of the policy class to obtain the information order.

\ifdefined\operationsresearch

\else

\begin{figure}[!t]
% \begin{wrapfigure}{r}{0.45\textwidth}  % 'r' = right side, width of the wrapped box
  %\vspace{-1em}                        % tweak vertical placement
  \centering
  \includegraphics[width=0.6\textwidth]{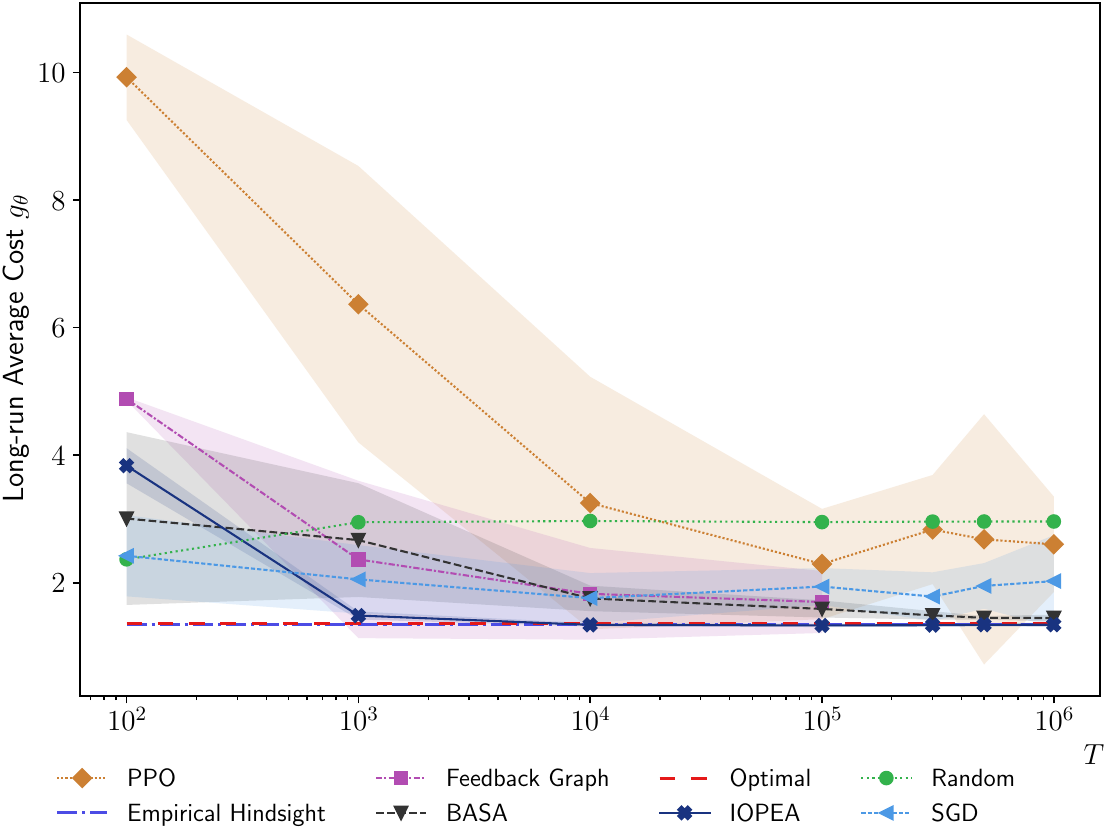}
  \caption{Comparison of the long‐run average cost \(g_{\theta}\) over time \(T\) for different algorithms in the small‐scale dual‐sourcing problem up to \(T=10^6\). \zzedit{We assume the demand follows the normal distribution as in~\Cref{tab:configs-by-block}.}  For {\textsf Feedback Graph}~\citep{dann2020reinforcement}, data are shown only to \(T=10^5\). \ALG attains near‐optimal gain by \(T=10^4\); {\textsf BASA}~\citep{chen2024tailored} remains suboptimal and continues to decline even at \(T=10^6\); {\textsf Feedback} Graph matches {\textsf BASA}’s long‐run cost at \(T=10^5\) but with much larger variance; {\textsf PPO} converges markedly slower than all other methods. \zzedit{\textsf{SGD} fails to converge effectively, as the problem is non-convex.} Note that in some trials {\textsf PPO} and {\textsf Feedback Graph} may exceed the “Optimal” curve, because “Optimal” denotes the best dual index policy rather than the true problem‐wide optimum.}
  \label{fig_convergence_curve}
  % \vspace{-1em}
% \end{wrapfigure}
\end{figure}

\fi

\myparagraph{Suboptimality of {\textsf BASA}~\citep{chen2024tailored}.} While {\textsf BASA} \citep{chen2024tailored} also converges quickly in the dual-sourcing problem (see~\cref{fig_convergence_curve}), it learns over the class of tailored base-surge policies.  These are known to underperform dual index policies in most non-asymptotic settings (i.e. when $L_r < \infty$~\citep{xin2023dual}).  This limitation is reflected in the observed performance gap between \ALG and {\textsf BASA} in our simulations, since {\textsf BASA} converges to a significantly suboptimal policy.  This reinforces the practical advantage of learning within the stronger-performing policy classes, such as dual index policies.

\myparagraph{Suboptimality of \textsf{SGD}.}
We observe three main limitations of \textsf{SGD}.  
First, it requires substantially more samples to converge. In the small‐scale inventory control problem, \textsf{SGD} performs comparably to \textsf{IOPEA}, but its performance deteriorates significantly in the large‐scale setting. Since this model is convex and \textsf{SGD} is theoretically guaranteed to converge to the optimum, this gap highlights its slower convergence rate.  
Second, the performance of \textsf{SGD} is highly sensitive to hyperparameter choices, such as the initial point and learning rate (see \cref{sec:simulation_information} for details).  
Finally, in the dual sourcing problem, \textsf{SGD} performs markedly worse, as expected, since the model is no longer convex. In contrast, our method \textsf{IOPEA} consistently maintains strong performance in the dual sourcing setting.

\myparagraph{Computational Limitations of {\textsf Feedback Graph}~\citep{dann2020reinforcement}.} Note that for the large-scale inventory control and dual sourcing problem, we do not include the performance of {\textsf Feedback Graph}, since this model-based approach requires storing an $O(|S|^2|A|)$ length vector for updating transition dynamics during training.  In our large-scale inventory control problem with $L = 6$ and demand mean $40/3$, a reasonable discretization of the state space with radius $0.1$ leads to a state space of size $\Theta(10^{14})$ after discretization, which is not tractable.  Hence, we omit these comparisons.

\section{Conclusion}\label{sec_conclusion}

In this paper, we introduced the novel concept of an \emph{information order} over a class of policies in infinite‐horizon average‐cost MDPs, and showed how to leverage this information order to improve sample efficiency in learning.  Our main theoretical contribution is a regret bound that scales as \(\tilde O\bigl(H\sqrt{\frac{wd}{\alpha}T}\bigr)\), where \(w\) is the width of the information‐order, which interpolates smoothly between the no‐feedback (\(w=|\Theta_1|\)) and full‐feedback (\(w=1\)) regimes.

We instantiated our Information‐Ordered Epoch‐Based Policy Elimination Algorithm (\ALG) on three canonical operations research problems: inventory control with positive lead time, dual sourcing, and an M/M/1/L queue with controllable service rates, obtaining $\sqrt{T}$ scaling regret guarantees in each setting with a low-width information order. Notably, as far as we know, this is the first result obtaining $\sqrt{T}$-regret for dual index policies in the lost-sales dual sourcing problem.  Numerical simulations confirm that \ALG attains performance close to the theoretical optimum and significantly outperforms baseline methods.

Beyond these case studies, our policy‐order framework offers a unified approach to leveraging partial feedback in a wide range of decision‐making problems and opens several avenues for future research.  One could investigate information orders and counterfactual estimation in nonstationary environments, such as inventory systems with time‐varying demand distributions. Another future direction is to address settings where observations from a single policy~$\theta$ only partially inform about another policy~$\theta'$. In such cases, producing an accurate counterfactual estimate for~$\pi_{\theta'}$ requires implementing multiple policies and collecting data under each one. This challenge appears in many operations research problems, such as online assortment optimization, where different policies have overlapping but nonidentical informational content.

% \medskip

\noindent\textbf{Acknowledgments.} Part of this work was done while the authors were visiting the Simons Institute for the Theory of Computing for the semester on Data-Driven Decision Processes.

\clearpage
\newpage

\bibliographystyle{informs2014} % outcomment this and next line in Case 1
\bibliography{references} % if more than one, comma separated

%\THEEndNotes
% \begingroup \parindent 0pt \parskip 0.0ex \def\enotesize{\normalsize} \theendnotes \endgroup

% Appendix here
% Options are (1) APPENDIX (with or without general title) or
%             (2) APPENDICES (if it has more than one unrelated sections)
% Outcomment the appropriate case if necessary
%
% \begin{APPENDIX}{<Title of the Appendix>}
% \end{APPENDIX}
%
%   or
%

\crefalias{section}{appendix}
\crefalias{subsection}{appendix}
\crefalias{subsubsection}{appendix}
\begin{APPENDICES}

\makeatletter
\renewcommand{\theHsection}{appendix.\Alph{section}}
\renewcommand{\theHsubsection}{appendix.\Alph{section}.\arabic{subsection}}
\renewcommand{\theHsubsubsection}{appendix.\Alph{section}.\arabic{subsection}.\arabic{subsubsection}}
\makeatother
\OneAndAHalfSpacedXI % Current default line spacing

\ifdefined\acm  
\else 
\ifdefined\informs
\renewcommand{\arraystretch}{1.2}
\fi
\fi
\begin{table*}[t!]
\caption{Common notation}
\begin{tabular}{>{\color{edits}}l | >{\color{edits}}l}
\textbf{Symbol} & \textbf{Definition} \\ \hline
\multicolumn{2}{c}{Problem setting specifications}\\
\hline
$T$ & Total time horizon \\
$X,Y, X \in Y$ & Two random variables, and shorthand for $X$ measurable with respect to $Y$ \\
$\S,\A,P,C,s_1$ & State and action space, transition distribution, cost function, starting state \\
$\Cupperbound$ & Upper bound on one-step cost \\
$\Theta, \theta$ & Policy class, and arbitrary policy in the policy class \\
$\pi_\theta$ & $\pi_\theta : \S \rightarrow \A$, policy parameterized by $\theta \in \Theta$ \\
$S_t, \theta_t, A_t$ & State, selected policy parameter, and action at time $t$ where $A_t = \pi_{\theta_t}(S_t)$ \\
$g_\theta(s), v_\theta(s)$ & Long-run average cost and bias for policy $\pi_\theta$ \\
$\theta^*$ & Optimal policy in $\Theta$ \\
$\Regret(T)$ & $\Exp{\sum_{t=1}^T C(S_t, A_t) - T g_{\theta^*}(s_1)}$ \\
$\H_\theta^T$ & Sample path trajectory of length $T$ collected via $\pi_\theta$ \\
$\Gbar(\H_\theta^T)$ & Empirical average cost of policy $\theta$ \\
$\preceq, w, \alpha$ & Information order relation, width, and ``informativeness'' factor \\
$\CGbarprime(\mathcal{H}_{\theta}^T)$ & Counterfactual estimate of $\theta'$ from $\theta$ \\
$\Th(\delta)$ & A constant (horizon threshold) that depends on the confidence level \(\delta\) \\
$U,H$ & Upper bound on $\Theta$, and bound on the bias \\
$L_\Theta, \theta_R$ & Lipschitz constant of $g_\theta$, and the policy with finite expected return time to $s_1$ \\
\hline
\multicolumn{2}{c}{Algorithm specification}\\
\hline
$k, \Theta_k$ & Epoch and confidence set in epoch $k$ \\
$\Thetamax_k$ & Set of maximal policies according to $\preceq$ to $\Theta_k$ \\
$r$ & Discretization parameter \\
$N_k, \beta_k$ & Number of trajectories sampled in epoch $k$, and confidence term \\
\hline
\multicolumn{2}{c}{Inventory case study}\\
\hline
$\Inventory_t, \Order_t$ & Inventory level and ordering level at time $t$ \\
$D_t$ & Demand at time $t$ \\
$L, L_r, L_e$ & Lead time parameters \\
$h,p$ & Holding and lost-sales cost \\
$c_r, c_e$ & The long-lead purchasing cost and short-lead purchasing cost coefficient in dual-sourcing \\
$\Order_t^e, \Order_t^r$ & Order of expedited and regular channels \\
$\gamma$ & $\Pr(D_t = 0)$ \\
\hline
\multicolumn{2}{c}{Queuing case study}\\
\hline
$L, \lambda, \mu$ & Buffer size, arrival rate, deadline rate \\
$C, w(a), A_{\max}$ & Cost for missed deadline, power cost, and maximum service rate \\
$U, \lambda_{\max}, \mu_{\max}$ & Uniformization parameter, with known upper bounds on $\lambda$ and $\mu$ \\
$m_\theta(s)$ & Stationary distribution of policy $\theta$ \\
$V$ & The confidence interval in the information order \\
\hline

\end{tabular}
\label{table:notation}
\end{table*}

\newpage

\clearpage

\section{Omitted Discussions}

\begin{table}[t]
\centering
\caption{Comparison of regret, time, and space complexity under different structural assumptions. For Feedback Graph~\citep{dann2020reinforcement}, $p$ denotes the growth rate of the mas-number $M$, i.e., $M=O((1/r)^p)$; more details in~\citet{dann2020reinforcement}.}
\label{tab_complexity_comparison}
\small
\begin{tabular}{l l c c c}
\toprule
Algorithm & Condition & Regret & Time & Space \\
\midrule

IOPEA & Bounded width $w$ & $\tilde{O}(H\sqrt{wdT/\alpha})$ & $O(w\log(T) + T^{1 + \frac{d}{2}})$ & $O(T^{\frac{d}{2}})$ \\
IOPEA & $w=|\Theta_1|$ & $\tilde{O}(H\sqrt{d}T^{\frac{d+1}{d+2}})$ & $O(T)$ & $O(T^{\frac{d}{d+2}})$ \\
\citet{agrawal2019learning} & Convexity & $\sqrt{T}$-scaling & $O(T)$ & $O(T)$ \\
Feedback graph~\citep{dann2020reinforcement} & & $\tilde{O}\left( K^{\frac{d_S+p}{d_S+p+1}}\right)$ & $O\left( K^{\frac{3d_S+d_A+p+1}{d_S+p+1}} \right)$ & $O\left( K^{\frac{2d_S+d_A}{d_S+p+1}} \right)$ \\
\bottomrule
\end{tabular}
\end{table}

\subsection{Time and Space Complexity}
\label{discussion_complexity}

In this section, we provide the time and space complexity of~\ALG and compare it with related approaches. Concretely, we discuss three algorithms:  (i) \ALG when the information has small width $w$, (ii) \ALG regardless of the information order, i.e. $w = |\Theta_1|$, and (iii) algorithms that additionally assume $g_\theta$ is convex as in \citet{agrawal2019learning}.
We also compare with the feedback graph algorithm of~\citet{dann2020reinforcement}, although this comparison is not exactly one-to-one since (i)--(iii) concern infinite-horizon average-cost MDPs, whereas the feedback graph framework is formulated for finite-horizon MDPs with episode length $H_f$~\citep{dann2020reinforcement} ($H$ therein). In the feedback graph comparison below, we also discuss how this difference in problem settings affects the interpretation of the comparison.

\paragraph{Setting and Notation.}
Recall that $\Theta \subset \RR^{d}$ denotes a continuous compact policy class. The methods in (i)--(iii) rely on policy-class Lipschitzness, and we continue to denote the corresponding Lipschitz constant by $L_\Theta$ following~\Cref{assumption_MRP}. To discuss time and space complexity, we impose the following computational assumption.

\begin{assumption}
\label{assumption_complexity_oracle}
For any $\Theta' \subset \Theta$, the time and storage complexity of finding the maximal policies is $O(w)$. For any policies $\theta',\theta \in \Theta$ and any horizon $T$, given $\mathcal{H}_{\theta}^T$, the time and storage complexity of deriving the counterfactual estimate $\CGbarprime(\mathcal{H}_{\theta}^T)$ under the information order $\pi_{\theta'} \preceq \pi_{\theta}$ is $O(T)$.
\end{assumption}
Note that given a finite policy class $\Theta_1$ equipped with a width-$w$ information order as defined in~\Cref{section_policy_ordering}, finding all maximal policies may require $O(|\Theta_1|^2)$ time complexity in the worst case, analogous to algorithms for sets with partial orders~\citep{daskalakis2011sorting}. Moreover, $O(T)$ time and space complexity for constructing the counterfactual estimate $\CGbarprime(\mathcal{H}_{\theta}^T)$ is not guaranteed in general, and these complexities can in principle become arbitrarily large.
However, when the information order arises from underlying problem structure, as in the three case studies in~\Cref{sec_case_studies}, the maximal policies are often directly characterized by that structure rather than identified through pairwise policy comparisons, leading to $O(w)$ complexity. Furthermore, since the observation history itself has size $O(T)$, assuming $O(T)$ time and space complexity for constructing $\CGbarprime(\mathcal{H}_{\theta}^T)$ is also natural.

\subsubsection{Time and Space Complexity of~\ALG}
\label{discussion_complexity_1}

We start by analyzing the time and space complexity of \ALG under \Cref{assumption_complexity_oracle}, assuming an information order of width $w$.

\paragraph{Time Complexity.}
Since \ALG is an epoch based algorithm, we analyze the time complexity separately within each epoch.  Each epoch of \ALG consists of several parts: finding the set of maximal policies $\Thetamax_k$ according to the information order, implementing those policies to obtain a trajectory of length $N_k$, the reset phases for each of those policies where we play $\pi_{\theta_R}$ to return to $s_1$, performing the counterfactual estimation, and finally the policy elimination step.
Finding the set of maximal policies takes $O(w)$ time via \Cref{assumption_complexity_oracle}.  The implementation cost is straightforward, since each time step is $O(1)$.  Moreover, the counterfactual estimation over $\Theta_k$ to obtain $\CGbarprimez$ for all $\theta' \in \Theta_k$ takes time $|\Theta_k| O(N_k) = O(|\Theta_1| N_k)$ via \Cref{assumption_complexity_oracle}.
The policy elimination step again has time complexity $O(|\Theta_k|) = O(|\Theta_1|)$.  Adding these pieces together the total time complexity for the epoch is $O(w) + O(|\Theta_1| N_k)$. 

To analyze the time complexity for the algorithm across all epochs, let $K$ denote the number of epochs.  By~\Cref{main_lem0}, $K = O\left(\log_4 (T)\right)$ almost surely.
Hence we have that the total time complexity for the algorithm is:
\begin{align}
    \sum_{k=1}^K \left(O(w) + O(|\Theta_1| N_k)\right) & = O(wK) + O(|\Theta_1|) \sum_{k=1}^K N_k \notag \\
    & = O\bigl(w \log(T) + |\Theta_1| \frac{T}{w} \bigr), \label{eq_time_complexity_alg}
\end{align}
where the final equality uses $\sum_{k=1}^K N_k = O(T/w)$ almost surely.

\paragraph{Space Complexity.}
Apart from the implementation outcomes, the algorithm only needs to store the active policies for each $k \in [K]$, whose cardinality is at most $|\Theta_1|$. Note that all estimators, including the counterfactual estimators $\CGbarprimez$, can be updated online, so the algorithm only stores the current scalar estimate for each active policy rather than the full trajectories. Therefore, the total space complexity is $O(|\Theta_1|)$.

As a consequence, we obtain the following statement.
\ComplexityBoundforALG*

We note that while the computational complexity of \ALG scales with $O(|\Theta_1|)$ (which in turn scales with $T$ after discretization), this arises since we do not impose any global structure on the objective $g_\theta$ such as convexity or linearity. As a result, the algorithm must, in general, estimate the performance of each candidate policy in $\Theta_1$.  This is also what allows \ALG to apply beyond settings with global structure, including nonconvex examples such as the dual-index case study in \Cref{case_study:dual_index_policy}.

\subsubsection{Time and Space Complexity of \ALG when $w = |\Theta_1|$}
\label{discussion_complexity_2}

Note that the previous analysis did not impose any restriction on the width $w$.  Hence, in the worst case we have that $w = |\Theta_1|$. Following~\cref{eq_main_proof_3} in the proof of~\Cref{thm:regret_bound}, for any radius $r$, the regret scales as
$$
    O(w\log(T) + \sqrt{wd\log(1/r)T\log(T)}) + Tr,
$$
where we omit parameters that are irrelevant to the dependence on $T$, $r$, and $d$, including $H$, $\alpha$, $\Cupperbound$, $\Th(\cdot)$, $D_\Theta$, and $L_\Theta$. This term is minimized by taking $r = T^{-\frac{1}{d+2}}$.  This leads to the following proposition:

\begin{proposition}
    \label{propo_complexity_2}
    If $w=|\Theta_1|$, the time complexity is $O(T+ |\Theta_1| \log(T))$ and the space complexity is $O(|\Theta_1|)$.
    Moreover, if $\Theta \subset \RR^d$ and $\Theta_1=\mathcal{N}_r(\Theta)$ with $r=T^{-\frac{1}{d+2}}$, then $|\Theta_1|=O(T^{\frac{d}{d+2}})$. In this case, the regret scales as $T^{\frac{d+1}{d+2}}$, the time complexity scales as $O(T)$, and the space complexity scales as $O(T^{\frac{d}{d+2}})$.
\end{proposition}

The comparison between \cref{propo_complexity_1,propo_complexity_2} highlights a trade-off. When the information order has a small width, \ALG has an {\em exponentially} better regret guarantee, scaling as $\sqrt{T}$ rather than $T^{\frac{d+1}{d+2}}$.  This statistical improvement however comes with a mild computational cost, which, as noted above, arises from the finer discretization radius $r = T^{-1/2}$ used in the small-width case.

% In the low-width case, the $\sqrt{T}$ scaling regret guarantee uses the finer discretization radius $r=T^{-1/2}$, so $|\Theta_1|=O(T^{d/2})$. Since~\ALG does not impose global structure such as convexity or linearity, it must in general evaluate each candidate policy in $\Theta_1$, albeit counterfactually through the information order. Thus, the information order reduces the number of policies that must be implemented in the environment, hence improving regret, but it does not remove the dependence on the size of the discretized policy class in the computational and space complexity.

\subsubsection{Time and Space Complexity under Convexity~\citep{agrawal2019learning}}
We present the time and space complexity of the main algorithm in~\citet{agrawal2019learning} (Algorithm~1 therein). Other than the policy-class Lipschitzness introduced at the beginning of~\Cref{discussion_complexity}, following~\citet{agrawal2019learning}, we additionally assume policy-class convexity and restrict attention to the one-dimensional case $d=1$. The policy-class convexity condition is stated as follows: For any $\eta \in [0,1]$ and any $\theta_1,\theta_2,\theta_3 \in \Theta$ satisfying $\theta_2 = \eta\theta_1 + (1-\eta)\theta_3$, we have
\(
    g_{\theta_2} \le \eta g_{\theta_1} + (1 - \eta)g_{\theta_3}.
\)

In their Algorithm~1, at each time step $t$, the algorithm executes two types of operations. First, it runs the selected policy, observes the realized cost, and updates the relevant estimates; see Eq.~(9) of~\citet{agrawal2019learning}. Each of these operations requires only $O(1)$ time. Second, the algorithm checks whether it moves to the next round or the next epoch, and whether the working interval should be updated. These operations also require only $O(1)$ time. Therefore, the time complexity at each time step $t$ is $O(1)$, implying that the total time complexity of Algorithm~1 in~\citet{agrawal2019learning} is $O(T)$. The storage complexity is also $O(T)$ following the same argument.

\begin{proposition}
    \label{propo_complexity_3}
    Assume $d=1$. The time and space complexity of Algorithm~1 in~\citet{agrawal2019learning} are both $O(T)$ under policy-class convexity.
\end{proposition}
This is consistent with the role of convexity in the analysis.
Under convexity, the algorithm can eliminate subintervals of the current working interval $\Theta_k$ by evaluating a small number of representative policies, such as the left, middle, and right points of that interval.
As a result, it avoids estimating every candidate policy in a fine discretization of $\Theta$, while still achieving a $\sqrt{T}$ scaling regret guarantee.
In contrast, in the absence of convexity or other global structures (e.g., linearity), learning over a continuous policy class $\Theta$ generally requires estimating all candidate policies in $\Theta_1$ as in~\ALG. Consequently, $|\Theta_1| = |\mathcal{N}_{r}(\Theta)|$, the number of candidate policies after discretization, enters the time and space complexity of~\ALG, as shown in~\Cref{propo_complexity_1}.

\subsubsection{Time and Space Complexity of Feedback Graph~\citet{dann2020reinforcement}}
\label{discussion_feedback_graph}
We also discuss the time and space complexity of the feedback graph algorithm, namely Algorithm~1 of~\citet{dann2020reinforcement}, when applied to continuous-space MDPs through discretization.
We emphasize that the feedback graph algorithm is formulated for finite-horizon MDPs with episode length $H_f$, whereas our work~\ALG studies infinite-horizon average-cost MDPs. 
We first discuss the time and space complexities of~\citet{dann2020reinforcement} in the finite-horizon setting, and then comment on the gap between this setting and ours at the end.

Unlike settings (i)--(iii), the feedback graph framework of~\citet{dann2020reinforcement} is formulated at the state-action-pair level rather than at the policy level. Let $\S \subset \RR^{d_S}$ and $\A \subset \RR^{d_A}$ denote compact continuous state and action spaces, respectively.
When applying feedback graph algorithm to continuous state and action spaces, one also needs a Lipschitzness condition at the state-action level.
The value-function Lipschitzness condition below is one natural way to formalize this requirement~\citep{sinclair2023adaptive}.
Concretely, let $Q^*_{H_f,h}(s,a)$ and $V^*_{H_f,h}(s)$ denote, respectively, the optimal Q-function at state-action pair $(s,a)$ and step $h \in [H_f]$, and the optimal value function at state $s$ and step $h \in [H_f]$. We assume that there exists $L_f \in \RR^+$ such that, for any $a,a' \in \A$, $s,s' \in \S$, and $h \in [H_f]$,
\begin{align}
    |Q^*_{H_f, h}(s,a) - Q^*_{H_f, h}(s',a')| &\le L_f\|(s,a) - (s',a')\|_\infty. \\
    |V^*_{H_f, h}(s) - V^*_{H_f, h}(s')| &\le L_f\|s - s'\|_\infty.
    \label{finite_lipschitz_condition}
\end{align}
The norm $\|(s,a) - (s',a')\|_\infty$ is well-defined because $\S\subset \RR^{d_S}$ and $\A\subset \RR^{d_A}$. We omit the formal definitions of the $Q$-functions and value functions, which are standard; we refer the reader to Assumptions~1 and~2 of~\citet{sinclair2023adaptive} for further details.

Concretely, we discretize the state and action spaces with radius $r$. Under the $\ell_\infty$ norm, the resulting discrete MDP has $|\mathcal{N}_{r}(\S)| = (1/r)^{d_S}$ and $|\mathcal{N}_{r}(\A)| = (1/r)^{d_A}$. We further assume that the mas-number $M$ of the feedback graph, namely the cardinality of the largest acyclic subgraph as defined in~\citet{dann2020reinforcement}, scales as $O((1/r)^p)$, where $0 \le p \le d_S+d_A$. The full-feedback case corresponds to $p=0$, whereas the case with no nontrivial feedback graph corresponds to $p=d_S+d_A$. Intuitively, a smaller $p$ represents stronger side-observation capability. We discuss how the regret, time complexity, and space complexity scale with $p$.

Under this setting, Theorem~1 of~\citet{dann2020reinforcement} gives the following regret bound for the discrete problem over $K$ episodes:
$$
    \tilde{O}\left(\sqrt{M K} + M |\mathcal{N}_{r}(\S)|\right).
$$
Combining this bound with the discretization error, and omitting factors that are irrelevant to the dependence on $K$, $r$, $d_S$, $d_A$, and $p$, the regret scales as
\begin{align}
    \tilde{O}\left(\sqrt{M K} + M |\mathcal{N}_{r}(\S)|\right) + O(Kr)
    =
    \tilde{O}\left(\sqrt{(1/r)^p K} + (1/r)^p (1/r)^{d_S} + Kr\right). 
    \label{eq_feedback_graph_regret}
\end{align}
Furthermore, the time complexity per episode on the discretized MDP is $O(|\mathcal{N}_{r}(\S)|^2|\mathcal{N}_{r}(\A)|)$; see Appendix~B.2 of~\citet{dann2020reinforcement}. This term is independent of the value of $M$, since the algorithm still updates the transition matrix for all $(s,a)$ pairs; see Lines~5--10 of Algorithm~3 in~\citet{dann2020reinforcement}. Thus, the total time complexity over $K$ episodes is
\(
    O(|\mathcal{N}_{r}(\S)|^2|\mathcal{N}_{r}(\A)|K)=O((1/r)^{2d_S+d_A}K).
\)
Since a transition vector must be stored for each state-action pair, the total space complexity is
\(
    O((1/r)^{2d_S+d_A}).
\)
Optimizing the regret bound in~\cref{eq_feedback_graph_regret} gives, up to logarithmic and other omitted constants, the optimal choice of $r$ as
$
    r = K^{-\frac{1}{d_S+p+1}}.
$
Plugging this choice of $r$ into the above expressions yields the following statement.

\begin{proposition}
    \label{propo_complexity_4}
    Let $M = O((1/r)^p)$ for some $p \in [0,d_S+d_A]$. With the regret-optimizing choice $r=K^{-\frac{1}{d_S+p+1}}$, the regret scales as $\tilde{O}\left(K^{\frac{d_S+p}{d_S+p+1}}\right)$, the time complexity scales as $O\left(K^{\frac{3d_S+d_A+p+1}{d_S+p+1}}\right)$, and the space complexity scales as $O\left(K^{\frac{2d_S+d_A}{d_S+p+1}}\right)$.
\end{proposition}
Note that when there is no feedback graph, i.e., when $p = d_S + d_A$, \Cref{propo_complexity_4} nearly recovers the regret, time, and space complexity of the AdaMB algorithm for Lipschitz MDPs reported in Table~1 of~\citet{sinclair2023adaptive}. 
The bounds in~\Cref{propo_complexity_4} are slightly looser than AdaMB because under the chosen discretization radius $r = K^{-\frac{1}{d_S+p+1}}$, the term $M |\mathcal{N}_{r}(\S)|$ may dominate the statistical term $\sqrt{MK}$ in~\cref{eq_feedback_graph_regret}.

Several observations follow.
First, except for the special case \(p=0\) and \(d_S=1\), the regret bound obtained from the feedback graph is worse than the $\sqrt{T}$ scaling regret bound of~\ALG, treating \(K\) and \(T\) as being of the same order. This remains true even in the {\em full-feedback} case \(p=0\): whenever \(d_S>1\),~\ALG yields a sharper regret bound.

Second, a smaller \(p\) improves the regret bound, but under the regret-optimizing discretization radius it increases both the time and space complexity. This is consistent with the tradeoff discussed above; compared with the trivial information order with \(w = |\mathcal{N}_{r}(\Theta)|\) in~\Cref{discussion_complexity_2}, a small-width information order, as discussed in~\Cref{discussion_complexity_1}, leads to a sharper regret bound, but also incurs heavier time and space complexity.

Third, although the time and space complexity of~\ALG scale with the policy dimension $d$, namely $O(T^{1+\frac{d}{2}})$ and $O(T^{\frac{d}{2}})$, which may appear heavier than those of the feedback graph framework, we emphasize that this is only because~\ALG achieves a sharper regret bound, as discussed in~\Cref{discussion_complexity_2}. In fact, one can always choose a coarser discretization radius when applying~\ALG. Under bounded width $w$, by letting
\(
    r = (1/T)^{\frac{2d_S+d_A}{d(d_S+p+1)}},
\)
from~\cref{eq_time_complexity_alg} we obtain that the time and space complexity become
\(
    O((1/r)^dT)=O\left(T^{\frac{3d_S+d_A+p+1}{d_S+p+1}}\right)
\)
and
\(
    O\left(T^{\frac{2d_S+d_A}{d_S+p+1}}\right),
\)
respectively, matching the bounds in~\Cref{propo_complexity_4}. The corresponding regret scales as
\(
    \sqrt{T} + T^{1-\frac{2d_S+d_A}{d(d_S+p+1)}}.
\)
Whenever $d < 2d_S+d_A$, the regret from~\ALG is strictly smaller than that from the feedback graph framework, since
\(
    T^{1-\frac{2d_S+d_A}{d(d_S+p+1)}} < T^{\frac{d_S+p}{d_S+p+1}}.
\)
Note that $d$ depends heavily on the policy class $\Theta$. Recall the two inventory planning case studies in~\Cref{sec_case_studies}; In~\Cref{case_study:inventory}, we have $d=1$ and $d_S=L+1$; in~\Cref{case_study:dual_index_policy}, we have $d=2$ and $d_S=L_r+L_e+1$. Thus, in both case studies, applying the feedback graph framework would lead to a worse regret bound, even after matching the same time and space complexity scaling, since the condition $d < 2d_S+d_A$ holds trivially.

We lastly comment on the role of $p$, namely the order of the mas-number $M$. Requiring a small $p$ can sometimes be substantially more restrictive than requiring a low-width information order in~\ALG. 
The inventory case study in~\Cref{case_study:inventory} provides such an example. There, we have $w=1$. We now show that, under the state-action feedback relation, we have $p = 1$ instead of $p = 0$. 
Indeed, for any time step $t$, consider two state-action pairs $s_t = (I_t, Q_{t-L}, \ldots, Q_t)$ and $s'_t = (I'_t, Q'_{t-L}, \ldots, Q'_t)$ (see~\cref{eq_ic_dynamics}). 
Regardless of the actions $a_t$ and $a'_t$, the two state-action pairs cannot counterfactually reveal each other if and only if
\(
    I_t+Q_{t-L}\neq I'_t+Q'_{t-L},
\)
since the observability of the censored demand $D_t$ depends solely on $I_t+Q_{t-L}$.
Hence, $M$ scales at least as $\Omega(1/r)$, since after discretization there are at least $O(1/r)$ distinct realizations of $Q_t + I_{t-L}$.
Also, since there are at most $\Theta(1/r)$ distinct realizations of $Q_t + I_{t-L}$, we have $p \le 1$. Therefore, $p = 1$.
As a consequence, by~\Cref{propo_complexity_4}, the feedback graph framework cannot achieve a regret bound better than $\tilde{O}(T^{2/3})$ in this case study.  

To understand why obtaining a smaller $p$ can be substantially harder than obtaining a low width $w$, the key intuition is that the counterfactual estimation structure in~\Cref{case_study:inventory} is essentially asymmetric; higher inventory levels can reveal lower inventory levels, but not conversely. Our information order does not require the feedback relation to be symmetric, since~\ALG {\em~can choose more informative policies to play}. This is not supported by the feedback graph, where the feedback structure does not guide which state-action pairs are visited. This explains why the feedback graph fundamentally relies on loop structures in the graph, and why~\ALG can behave significantly better in the settings discussed above.

One caveat is that the $T^{2/3}$ scaling regret bound of the feedback graph is against the globally optimal policy, whereas our $\sqrt{T}$ scaling guarantee is an in-class regret bound over the predefined policy class \(\Theta\); see~\Cref{section_modelling_assumptions}. Thus, the two regret guarantees are not directly comparable, since a precise comparison would also need to account for the richness of $\Theta$. Nevertheless, the discussion above captures the main distinction between~\ALG and the feedback graph framework.

We finally briefly discuss the relation between the infinite-horizon MDP studied in this paper and the finite-horizon MDP setting where the feedback graph is applied.
Some existing works apply finite-horizon MDP algorithms to infinite-horizon average-cost MDPs by letting the episode length $H_f$ scale with $T$; see, e.g.,~\citet{chen2022learning}, where $H_f$ scales with $T^{1/3}$. Since the regret bound of the feedback graph scales polynomially with $H_f$ (see Theorem~1 in~\citet{dann2020reinforcement}), such finite-horizon approximations would only worsen the regret bound compared with the fixed-$H_f$ setting discussed above in~\Cref{propo_complexity_4}. Thus, applying the feedback graph to infinite-horizon average-cost MDPs through such approximations would not change the conclusions above regarding the superior performance of~\ALG in the corresponding settings.
We leave how to leverage the information order and~\ALG in finite-horizon MDPs for future work.

\subsection{Unknown Parameters in the Implementation of~\ALG}
\label{discussion_unknown_parameters}

Implementing \ALG under the distributional information order requires specifying the constants $\alpha$ and $\Th(\cdot)$ appearing in \cref{alg_policy_elimination}.  These constants may depend on unknown problem primitives. For instance, in the dual-index case study in \cref{case_study:dual_index_policy}, these constants depend on $\gamma = \Pr(D_t = 0)$, which may be unknown to the learner. However, it suffices to have a known lower bound on $\alpha$ and upper bound on $\Th(\cdot)$.
Intuitively, a smaller value of $\alpha$ means that the algorithm assumes less reusable information than is actually available, while a larger value of $\Th(\cdot)$ means that the algorithm waits longer before applying the distributional-order guarantee.

More precisely, suppose that the true distributional-order constants are $\alpha^*$ and $\Th^*(\delta)$.
If the learner uses constants $\underline{\alpha} < \alpha^*$ and $\overline{\Th}(\delta) \geq \Th^*(\delta)$, then the same distributional-order guarantee remains valid. Consequently, \ALG and the regret guarantee in \cref{thm:regret_bound} continue to apply after replacing $\alpha^*$ and $\Th^*(\cdot)$ by $\underline{\alpha}$ and $\overline{\Th}(\cdot)$.  In particular, if these conservative bounds differ from the true ones by at most absolute constant factors, then the regret bound changes only by absolute constant factors.

Even when no constant lower bound $\alpha^*$ is known, one may use a horizon-dependent conservative choice.  Suppose that the distributional-order argument remains valid for every $\underline{\alpha} \leq \alpha^*$ together with a corresponding threshold $\overline{\Th}(\delta)$.  Then one may set $\underline{\alpha} = 1 / \log(T)$.  For sufficiently large $T$, this choice satisfies $\underline{\alpha} \leq \alpha^*$.  The leading term in \cref{thm:regret_bound} then becomes $\tilde O(H\sqrt{wdT / \underline{\alpha}}) = \tilde O(H\sqrt{wdT\log(T)})$, which differs from the original rate only by an additional logarithmic factor.

We now verify that this argument applies to the dual-index case study.  This follows directly from~\Cref{dip_lem6}. \Cref{dip_lem6} states that, with high probability, whenever $T \ge \Th(\delta)= C\frac{\log(1/\delta)}{(1-\gamma)^4\gamma^{4L_r}}$
for some absolute constant $C$, the number of periods in which the system has $z_r^\theta$ units of on-hand inventory after order arrivals is lower-bounded by $\frac{1-\gamma}{2}\gamma^{L_r}T = \alpha^*T$.
Thus, whenever
\[
\frac{1}{\log(T)}\le \alpha^*, 
\qquad 
\log(T) \ge \frac{1}{(1-\gamma)^4\gamma^{4L_r}},
\]
the same lower bound remains valid for $\underline{\alpha}=1/\log(T)$ with $\overline{\Th}(\delta)= C\log(1/\delta)\log(T)$.
Therefore, the distributional order is valid for \(\underline{\alpha}\) and \(\overline{\Th}(\delta)\), and we obtain the same regret bound for the dual-index case study up to additional logarithmic factors, without knowing the value of $\gamma=\Pr(D_t=0)$.

More generally, in practical settings where the distributional order is established by analyzing a high-probability event, as in the dual-index case study (\Cref{dip_lem6}), one may work with conservative bounds of the unknown $\alpha$ and $\Th(\cdot)$, as shown above. However, we also note that this argument for handling unknown $\alpha$ and $\Th(\cdot)$ is still case-specific. In specially designed settings where a distributional order is valid only for one specific value $\alpha^*$, not knowing the true $\alpha^*$ may make \ALG not implementable for that distributional order. Nevertheless, the trivial order with $w=|\Theta_1|$ is always valid. 
We finally note that implementing \ALG also requires knowledge of $H$, which can be handled in the same way by using a conservative upper bound.

\counterwithin{theorem}{section}
\counterwithin{lemma}{section}
\counterwithin{definition}{section}

% \srscomment{TODO List for the Appendix:
% \begin{itemize}
%     \item Details on numerical simulations (parameter specifications, hyperparameter tuning, time for training, etc). Look at \citet{wan2024exploiting}.\zzcomment{updated}
%     \item Further results on numerical simulations (for the other problem specifications) and any additional discussion you think is required\zzcomment{updated}
%     \item Consistent notation for $\Pr(\cdot)$ in the appendix
%     \item Ensure consistent notation with the main paper (namely $h$ and $p$ or $h$ and $b$ selection for the cost function)
%     \item Finalize proof for regret upper bound (including putting in the concentration argument from Shipra's paper)\zzcomment{updated}
%     \item Finalize proof of regret lower bound, alongside the discussion on the $1 / \alpha$ factor for the \zzcomment{updated} 
%     \item Finalize proof of regret bound and the performance guarantees in the dual index and queueing example\zzcomment{updated}
%     \item Any additional discussion for the case studies?
%     \item Add a comment that best-inclass TBS vs ours, in the dual index setting\zzcomment{updated}
%     \item Discuss we do not need the specific form of cost function for the case studies. Can modify it to break convexity, and our results still hold
%     \item Check over table of notation
% \end{itemize}}

% \newpage

\section{Proof of Theorems in Section \ref{sec_main_result}}

\subsection{Regret Upper Bound (\cref{thm:regret_bound})}
\label{app:regret_bound_proof}

We start off this section by providing a complete proof of the regret upper bound of our algorithm \ALG (see \cref{alg_policy_elimination}).
\RegretBound*
Note that the regret bound for \ALG explicitly relies on the choice of the information order. While all policy classes have a trivial information order, the resulting regret guarantee will be exponential with respect to $d$. 
 Hence, our results require a policy class that is sufficiently ``informative'' (i.e. of low width).  See \Cref{sec_main_result} (Choice of Information Order) for more discussion on the choice of policy order and its implications on regret. 

We denote by $K$ the number of implemented epochs; note that $K$ is stochastic, as it depends on the ``restart'' phases. We slightly abuse notation and set 
\(
    \deltalocal = \frac{\delta}{(|\mathcal{N}_{r}(\Theta)| \log(T))},
\)
$\beta_k = \frac{ H}{\alpha N_k} + (H + 2\Cupperbound)\sqrt{\frac{2 \log(4/\deltalocal)}{\alpha N_k}}$, and $N_k = \frac{2^{2k} \Th(\deltalocal)}{\alpha}$ throughout the proof, which are consistent with the definitions in~\Cref{thm:regret_bound} when $\delta = 1/T$.
Our result requires a series of lemmas.  
We first upper bound the number of epochs $K$ as follows.
\begin{lemma}\label{main_lem0}
    For any $\delta > 0$, under $N_k = \frac{2^{2k} \Th(\deltalocal)}{\alpha}$, we have that $K = O(\log(T))$ almost surely.
\end{lemma}
\begin{rproof}
For any $\delta > 0$,
\begin{align}
    T & \ge \sum_{k=1}^K wN_k = \sum_{k=1}^K w\left(\frac{2^{2k} \Th(\deltalocal)}{\alpha}\right) \nonumber\\
    & = \frac{4w\Th(\deltalocal)}{3\alpha}(4^K - 1), \label{eq_main_proof_2}
\end{align}
where the inequality follows from ignoring the ``restart'' phases.
Solving this equation for $K$ gives that $K = O\left(\log_4 \left(\frac{\alpha T}{w \Th(\deltalocal)}\right)\right) = O(\log(T))$ almost surely, as $\alpha \le 1$, $w \ge 1$, and $\Th(\deltalocal) \ge 1$. 
\end{rproof}

Note that~\Cref{main_lem0} also requires $T \ge 4\frac{w\Th(\deltalocal)}{\alpha}$, since $K$ must be greater than 1. The next lemma establishes concentration on the estimates $\CGbarprimez$ versus $g_{\theta'}$ in terms of the number of ``useful'' samples in the epoch $\alpha N_k$.

\begin{lemma}\label{main_lem1}
Recall that with fixed $\alpha \in (0, 1]$, $\CGbarprimez$ is the counterfactual average cost estimate of policy $\theta'$ from observed sample trajectories under the maximal policy $\tilde{\theta}_k(\theta') \in \Thetamax_k$. For any fixed $\delta > 0$, define the following event:
\[
\E_0 = \left\{ \forall k, \theta' \in \Theta_k \mid \abs{\CGbarprimez - g_{\theta'}} \leq \frac{ H}{\alpha N_k} + (H + 2\Cupperbound) \sqrt{\frac{2 \log(4 |\mathcal{N}_{r}(\Theta)| \log(T) / \delta)}{\alpha N_k}}\right\}.
\]
Then we have that $\E_0$ occurs with probability at least $1 - 2\delta$.
\end{lemma}

% \srscomment{Add a comment here re. the dependence on $\alpha$. Emphasize that while $N_k$ is expanded to include $\alpha$, only a fraction of the data is actually used to counterfactual estimate the policy.}\zzcomment{updated}

% \srscomment{To introduce the proof you should instead be: We show the concentration of the estimates for $\theta$ using the counterfactual estimates combined with Lemma 3 of ~\citet{agrawal2019learning} (which we copy here for completion). Then comment on the union bound on $K$ and $N_r(\Theta)$ at the end of the proof.}\zzcomment{updated}

We establish \Cref{main_lem1} by showing that the counterfactual estimate $\CGbarprimez$ concentrates around the true cost $g_{\theta'}$ uniformly over all $\theta' \in \mathcal{N}_r(\Theta)$ and all epochs $k \in [K]$ with high probability. The main idea is to apply the concentration bound of a Markov process (Lemma 3 of~\citet{agrawal2019learning}), restated in \Cref{lem_concentration_lem_shipra} for completeness, together with the fact that $\CGbarprimez$ approximates the empirical estimate of $\Gbarprime(\H_{\theta'}^{\alpha T})$ (as guaranteed by the information order in \Cref{definition_general_order}). The final result follows via a union bound over $\log(T)$, which is an upper bound of $K$ almost surely, and the covering size $|\mathcal{N}_r(\Theta)|$.

\begin{lemma}[Lemma 3 of \citet{agrawal2019learning}]
\label{lem_concentration_lem_shipra}
Suppose that for any given policy $\theta'$, the gain function $g_{\theta'}(s)$ is constant over all $s \in \S_{\theta'}$ and the span of the bias is at most $H$. Then, given a sample trajectory $\H_{\theta'}^{T}$ of length $T$, we have that, for any $\delta > 0$, with probability at least $1 - \delta$,
\[
|\Gbarprime(\H_{\theta'}^{T}) - g_{\theta'}(s_1) | \leq \frac{H}{T} + (H + 2\Cupperbound) \sqrt{\frac{2 \log(4 / \delta)}{T}}.
\]
\end{lemma}

\begin{rproof}
We recall the Theorem 8.2.6 from \citet{puterman2014markov} first:
\begin{lemma}[Theorem 8.2.6 of \citet{puterman2014markov}]\label{lem_826}
For any fixed policy $\theta'$, for any $t$, and for any state $S_t \in \mathcal{S}$, the gain and bias satisfy
\[
g_{\theta'}(S_t) = \Ex_{\theta'}[C(S_t, \pi_{\theta'}(S_t))] + \Ex_{S'\sim P_{\theta'}(S_t)}[v_{\theta'}(S')] - v_{\theta'}(S_t).
\]
\end{lemma}
We bound $\left|\frac{1}{T}\sum_{t = 1}^T \Ex_{\theta'}[C(S_t, \pi_{\theta'}(S_t))] - g_{\theta'}\right|$ first. Denote $S_1, \ldots, S_T$ as the observed states within the sample trajectory $\H_{\theta'}^{T}$,
\begin{align}
\Bigl|\frac{1}{T}\sum_{t=1}^T \Ex_{\theta'}\bigl[C(S_t,\pi_{\theta'}(S_t))\bigr]
   -g_{\theta'}\Bigr|
&= \frac{1}{T}\Bigl|\sum_{t=1}^T\bigl(\Ex_{\theta'}[C(S_t,\pi_{\theta'}(S_t))]-g_{\theta'}\bigr)\Bigr| \nonumber\\
&= \frac{1}{T}\Bigl|\sum_{t=1}^T
    \bigl(\Ex_{\theta'}[C(S_t,\pi_{\theta'}(S_t))] \nonumber\\
&\quad-\bigl[\Ex_{\theta'}[C(S_t,\pi_{\theta'}(S_t))]
      +\Ex_{S'\sim P_{\theta'}(S_t)}[v_{\theta'}(S')]
      -v_{\theta'}(S_t)\bigr]\bigr)\Bigr| \label{eq_g_uniform}\\
&= \frac{1}{T}\Bigl|\sum_{t=1}^T\bigl(v_{\theta'}(S_t)
    -\Ex_{S'\sim P_{\theta'}(S_t)}[v_{\theta'}(S')]\bigr)\Bigr| \nonumber\\
&\le \frac{1}{T}\Bigl|v_{\theta'}(S_1)
    -\Ex_{S'\sim P_{\theta'}(S_T)}[v_{\theta'}(S')]\Bigr| \nonumber\\
&\quad+\frac{1}{T}\Bigl|\sum_{t=1}^{T-1}
    \bigl(v_{\theta'}(S_{t+1})
    -\Ex_{S'\sim P_{\theta'}(S_t)}[v_{\theta'}(S')]\bigr)\Bigr| \nonumber\\
&\le \frac{H}{T}
   +\frac{1}{T}\Bigl|\sum_{t=1}^{T-1}\Delta_{t+1}\Bigr|,\label{eq_g_uniform_2}
\end{align}
where the second line is due to \Cref{lem_826} and \Cref{assumption_MRP}, and we set
\[
\Delta_{t+1} := v_{\theta'}(S_{t+1}) -\Ex_{S'\sim \Pr_{\theta'}(S_t)}[v_{\theta'}(S')].
\]
Noting \(\Ex[\Delta_{t+1}| S_t]=0\) and \(|\Delta_{t+1}|\le H\), Azuma–Hoeffding~\citep{wainwright2019high} yields for any $\epsilon > 0$,
\[
\Pr\Bigl(\Bigl|\sum_{t=2}^T \Delta_t\Bigr|\ge \epsilon \Bigr)
\le 2\exp\Bigl(-\frac{\epsilon^2}{2(T-1)H^2}\Bigr).
\]
By setting $\epsilon = H \sqrt{2(T-1)\log\bigl(2/\delta\bigr)}$, we have that with probability at least $1-\delta$,
\begin{align}
\Biggl|\frac{1}{T}\sum_{t=1}^T \Ex_{\theta'}\bigl[C(S_t,\pi_{\theta'}(S_t))\bigr]
   -g_{\theta'}\Biggr|
\le
\frac{H}{T} + H\sqrt{\frac{2\log(2/\delta)}{T}} \label{eq_lemmaa31}.
\end{align}
Now we bound $|\Gbarprime(\H_{\theta'}^{T}) - g_{\theta'}(s_1) |$ correspondingly. Define $X_t = C(S_t,  \pi_{\theta'}(S_t)) - \Ex_{\theta'}[C(S_t, \pi_{\theta'}(S_t))]$, then
\[
|X_t| \le |C(S_t,  \pi_{\theta'}(S_t))|+| \Ex_{\theta'}[C(S_t, \pi_{\theta'}(S_t))]| \le 2\Cupperbound.
\]
Hence we have that $\{X_t\}_{t=1}^T$ is also a bounded martingale difference sequence.  With Azuma–Hoeffding~\citep{wainwright2019high}, with probability at least $1-\frac\delta2$,
\begin{align}
\frac{1}{T}\Bigl|\sum_{t=1}^T C(S_t,  \pi_{\theta'}(S_t)) - \Ex_{\theta'}[C(S_t, \pi_{\theta'}(S_t))]\Bigr| \le
2\Cupperbound\sqrt{\frac{2\log\bigl(4/\delta\bigr)}{T}}\label{eq_lemmaa32}.
\end{align}
The proof of \Cref{lem_concentration_lem_shipra} is complete by combining \cref{eq_lemmaa31} and (\ref{eq_lemmaa32}) above.
\end{rproof}

With the previous lemma in hand, we are now ready to show \Cref{main_lem1}.
\begin{rproofof}{\Cref{main_lem1}}
Observe that each epoch $k$ begins from the same fixed initial state $s_1$, regardless of the policy being implemented. As a counterfactual, consider executing policy $\theta'$ for $\alpha N_k = 4^k \Th(\deltalocal)$ steps during epoch $k$, in place of the maximal policies in $\Thetamax_k$. This would yield an empirical estimate $\Gbarprime(\H_{\theta'}^{\alpha N_k})$ of $g_{\theta'}$, which concentrates around $g_{\theta'}$ by \Cref{lem_concentration_lem_shipra}. 
Concretely, with probability at least $1 - \deltalocal$,
\[
|\Gbarprime(\H_{\theta'}^{\alpha N_k}) - g_{\theta'}| \leq \frac{ H}{\alpha N_k} + (H + 2\Cupperbound) \sqrt{\frac{2 \log(4 / \deltalocal)}{\alpha N_k}}.
\]

However, the policy $\theta'$ may not be actually implemented in epoch $k$, as $\theta'$ may not belong to $\Thetamax_k$. Due to the information order, we know that there exists $\tilde{\theta}_k(\theta') \in \Thetamax_k$ such that $\pi_{\theta'} \preceq \pi_{\tilde{\theta}_k(\theta')}$. Under \Cref{assumption_MRP}, note that the average cost $g_{\theta'}$ has no dependence on the state $s$, so $[g_{\theta'} - (\frac{H}{\alpha N_k} + (H + 2\Cupperbound) \sqrt{\frac{2 \log(4 / \deltalocal)}{\alpha N_k}}), g_{\theta'} + (\frac{H}{\alpha N_k} + (H + 2\Cupperbound) \sqrt{\frac{2 \log(4 / \deltalocal)}{\alpha N_k}})]$ is Borel measurable.
Then due to $N_k \ge \frac{4^k \Th(\deltalocal)}{\alpha} \ge \Th(\deltalocal)$, we have
$$
d_{\mathrm{TV}}(\CGbarprimez, \Gbarprime(\H_{\theta'}^{\alpha N_k})) \leq \deltalocal.
$$ 
By the definition of the total variation distance we get
\begin{align*}
    \Pr \left(|\CGbarprimez - g_{\theta'}| \le \frac{H}{\alpha N_k} + (H + 2\Cupperbound) \sqrt{\frac{2 \log(4 / \deltalocal)}{\alpha N_k}}\right) &\ge 1 - \deltalocal - \deltalocal = 1 - 2\deltalocal.
\end{align*}
The final result follows via a union bound over both $\log(T)$ and $|\mathcal{N}_r(\Theta)|$.
\end{rproofof}
Similarly, we have concentration for the empirical costs corresponding to the actual policies implemented from $\Thetamax_k$ in each epoch $k$.  Note that we analyze the costs for these policies separately since we can directly construct $\Gbar(\H_\theta^{N_k})$ as these policies were employed, instead of using the distributional order.

\begin{lemma}
    Define the following event
\[
\E_1 = \left\{ \forall k, \theta \in \Thetamax_k \mid |\Gbar(\H_{\theta}^{N_k}) - g_\theta| \leq \frac{H}{N_k} + (H + 2\Cupperbound) \sqrt{\frac{2 \log(4 w \log(T) / \delta)}{N_k}}\right\}.
\]
Then we have that $\E_1$ occurs with probability at least $1 - \delta$.
\end{lemma}
This lemma follows immediately from \cref{lem_concentration_lem_shipra} and a union bound.
Note that $w \leq |\mathcal{N}_{r}(\Theta)|$ and $\alpha \leq 1$. Applying a union bound over the events $\E_0$ and $\E_1$ then yields the following:

\begin{lemma}\label{lem_concentration_final}
Define the event
\[
\E_2 = \left\{ \forall k,\, \theta' \in \Theta_k,\, \theta \in \Thetamax_k \;\middle|\; \left| \CGbarprimez - g_{\theta'} \right| \leq \beta_k \text{ and } \left| \Gbar(\H_\theta^{N_k}) - g_\theta \right| \leq \beta_k \right\}.
\]
Then $\E_2$ holds with probability at least $1 - 3\delta$.
\end{lemma}

Next we establish that under the event $\E_2$, \ALG maintains that the optimal parameter $\theta^*_{r} \in \Theta_k$ for each $k$ with high probability, where 
$$
\theta^*_r = \argmin_{\theta \in \mathcal{N}_{r}(\Theta)} g_\theta.
$$
We similarly establish that the average cost of {\em all} policies contained in $\Theta_k$ is bounded by $O(\beta_k)$, as long as $k \ge 2$.
\begin{lemma}\label{lem_a.5}
Under the event $\E_2$ we have that $\theta^*_r \in \Theta_k$ for all $k$. Moreover, for any two $\theta, \theta' \in \Theta_k^2$ and $k \ge 2$, we have that:
\[
|g_{\theta} - g_{\theta'}| \leq 16 \beta_{k}.
\]
\end{lemma}
\begin{rproof}
We first start by showing that $\theta^*_r \in \Theta_k$ for all $k$.  We show this via induction over the epochs $k$.  Clearly for the base case when $k = 1$ we have that $\theta^*_r \in \mathcal{N}_{r}(\Theta) = \Theta_1$.  For the step case $k \rightarrow k+1$ suppose that $\theta^*_r \in \Theta_k$.  Then we have that if $\ermthetap_k = \argmin_{\theta' \in \Theta_k} \CGbarprimez$:
\begin{align*}
     \tilde{G}_{\theta^*_r|\tilde{\theta}_k(\theta^*_r)} - \tilde{G}_{\ermthetap_k|\tilde{\theta}_k(\ermthetap_k)} & = \tilde{G}_{\theta^*_r|\tilde{\theta}_k(\theta^*_r)} - g_{\theta^*_r} + g_{\theta^*_r} - g_{\ermthetap_k} + g_{\ermthetap_k} - \tilde{G}_{\ermthetap_k|\tilde{\theta}_k(\ermthetap_k)} \\
    & \leq \tilde{G}_{\theta^*_r|\tilde{\theta}_k(\theta^*_r)} - g_{\theta^*_r} +g_{\ermthetap_k} - \tilde{G}_{\ermthetap_k|\tilde{\theta}_k(\ermthetap_k)} \\
    & \leq 2 \beta_k.
\end{align*}
Note that in the first inequality we used that $g_{\theta^*_r} \le g_{\ermthetap_k}$ since $\theta^*_r$ is the optimizer over $\mathcal{N}_{r}(\Theta)$, and the second inequality is from the definition of the event $\E_2$.

To show the second property, we note that for any $\theta, \theta' \in \Theta_k$ and $k \ge 2$, $|\tilde{G}_{\theta|\tilde{\theta}_{k-1}(\theta)} - \tilde{G}_{\theta'|\tilde{\theta}_{k-1}(\theta')}| \leq 2 \beta_{k-1}$, from the definition of $\Theta_k$.  Hence we have that
\begin{align*}
 |g_\theta - g_{\theta'}| = |g_\theta - \tilde{G}_{\theta|\tilde{\theta}_{k-1}(\theta)} + \tilde{G}_{\theta|\tilde{\theta}_{k-1}(\theta)} - \tilde{G}_{\theta'|\tilde{\theta}_{k-1}(\theta')} + \tilde{G}_{\theta'|\tilde{\theta}_{k-1}(\theta')}- g_{\theta'}| \le 2 \beta_{k-1} + 2 \beta_{k-1} \le 16\beta_k,
\end{align*}
where we again used the event $\E_2$ and the fact that both $\theta, \theta' \in \Theta_k$. The last inequality is from that $N_k = 4N_{k-1}$, which leads to 
\begin{align*}
    \beta_{k-1} &= \frac{ H}{\alpha N_{k-1}} + (H + 2\Cupperbound)\sqrt{\frac{2 \log(4 |\mathcal{N}_{r}(\Theta)| \log(T)/\delta)}{\alpha N_{k-1}}}  \\
    &= \frac{ 4H}{\alpha N_{k}} + (H + 2\Cupperbound)\sqrt{\frac{8 \log(4 |\mathcal{N}_{r}(\Theta)| \log(T)/\delta)}{\alpha N_{k}}} \\
    &\le 4\beta_k.
\end{align*}
\end{rproof}

With the previous two lemmas in hand we are finally ready to prove \Cref{thm:regret_bound}.
\begin{rproofof}{\Cref{thm:regret_bound}}
% Each epoch $k$ consists of at most $w(N_k + D_{\Theta})$ timesteps in expectation: the first $wN_k$ steps correspond to executing the maximal policy set $\Thetamax_k$ to estimate the performance of all policies in $\Theta_k$, and the remaining $wD_{\Theta}$ steps account for returning to the fixed initial state $s_1$, as ensured by \Cref{assumption_MRP}. 
With $K = O(\log(T))$ almost surely, as established in~\Cref{main_lem0}, we next derive a regret decomposition for the algorithm:
\begin{align}
    \Regret(T) \leq \Ex\left[\sum_{k=1}^{K} \sum_{\theta \in \Thetamax_k} N_k(\Gbar(\H_\theta^{N_k})- g_{\theta^*_r})\right] + O(w\Cupperbound D_\Theta\log(T)) + TL_{\Theta}r. \label{eq_main_proof}
\end{align}
The first term reflects the cost incurred while executing each $\theta \in \Thetamax_k$ within an epoch. The second term captures the $D_{\Theta}$ steps needed to return to $s_1$ in expectation after each policy execution, since $K = O(\log (T))$ almost surely and the per-step cost lies in $[-\Cupperbound,\Cupperbound]$. The third term captures the discretization error $|g_{\theta^*_r} - g_{\theta^*}|$, which is bounded by the Lipschitz continuity of the cost function.

For the first term in~\cref{eq_main_proof}, we condition on the event $\E_2$. Specifically, under $\E_2$,
\begin{align*}
    \sum_{k=1}^{K} \sum_{\theta \in \Thetamax_k} (N_k(\Gbar(\H_\theta^{N_k})- g_{\theta^*_r})) & \leq \sum_{k=2}^{K}\sum_{\theta \in \Thetamax_k} (N_k \beta_k +N_k(g_{\theta}(s_1) - g_{\theta^*_r}(s_1))) + 2w\Cupperbound N_1 \\
    & \leq \sum_{k=2}^{K} 17 w N_k \beta_k + 2w\Cupperbound N_1 \quad \text{almost surely.}
\end{align*}
The first inequality follows from the event $\E_2$, the almost-sure bound $\Gbar(\H_\theta^{N_1}) \le \Cupperbound$ for all $\theta$, and the bound $|g_{\theta^*_r}| \le \Cupperbound$. The second follows from \Cref{lem_a.5}, since $\theta^*_r$ belongs to $\Theta_k$ and $\Thetamax_k \subset \Theta_k$.
With the definition of $N_k$ and $\beta_k$ we have:
\begin{align*}
    17 \sum_{k=2}^K w N_k \beta_k & = 17 w \sum_{k=2}^K N_k \left( \frac{H}{\alpha N_k} + (H+2\Cupperbound) \sqrt{\frac{2 \log(4 |\mathcal{N}_{r}(\Theta)|\log(T) / \delta)}{\alpha N_k}} \right) \\
    & \le \frac{17 w K H}{\alpha} + 17 w (H+2\Cupperbound)  \sqrt{\frac{2}{\alpha} \log (4 |\mathcal{N}_{r}(\Theta)| \log(T) / \delta)} \sum_{k=2}^K \sqrt{N_k} \\
    & = O\left(\frac{w H}{\alpha} \log_4 \left(\frac{\alpha T}{w \Th(\deltalocal)}\right)\right) + 17w (H+2\Cupperbound) \sqrt{\frac{2}{\alpha} \log(4 |\mathcal{N}_{r}(\Theta)| \log(T) / \delta)} \sum_{k=2}^K \sqrt{N_k}.
\end{align*}
Note that $\sum_{k=2}^K \sqrt{N_k} = O\left(\sqrt{\frac{\Th(\deltalocal)}{\alpha}}(\frac{\alpha T}{ w \Th(\deltalocal)})^{1/2}\right) = O\left(\sqrt{\frac{T}{w}}\right)$. Combining all the terms together gives that, with probability at least $1 - 3\delta$,
\begin{align}
\sum_{k=1}^{K} \sum_{\theta \in \Thetamax_k} N_k(\Gbar(\H_\theta^{N_k})- g_{\theta^*_r}) &\le  O\left(\frac{wH}{\alpha}\log_4 \left(\frac{\alpha T}{w \Th(\deltalocal)}\right)\right) + O\left((H+2\Cupperbound) \sqrt{\frac{w}{\alpha} \log(4 |\mathcal{N}_{r}(\Theta)| \log(T) / \delta)T}\right) \nonumber\\
&\qquad + 8\Cupperbound\frac{w\Th(\deltalocal)}{\alpha}.
\label{eq_main_proof_3}
\end{align}
By taking $r =(1/T)^{1/2}$, we have $|\mathcal{N}_{r}(\Theta)| = O((U\sqrt{T})^d)$.
Recall that
\(
\log_4 \left(\frac{\alpha T}{w \Th(\deltalocal)}\right) \le \log_4 (T) = O(\log (T)).
\)
Thus by taking $\delta = \frac{1}{T}$ and noticing that 
\(
\sum_{t = 1}^T C(S_t, A_t) - T g_{\theta^*}(s_1)
\)
is bounded by $2\Cupperbound T$ almost surely under any policy,
\begin{align}
    \Regret(T) &\le O\left(\frac{wH}{\alpha}\log(T) + (H+2\Cupperbound) \sqrt{\frac{w}{\alpha} \log(4 |\mathcal{N}_{r}(\Theta)| T\log(T) )T}\right) +  8\Cupperbound\frac{w\Th(\deltalocal)}{\alpha} + \frac{3}{T}\cdot 2\Cupperbound T \nonumber\\
    &\qquad  + O(w\Cupperbound D_{\Theta}\log(T)) + TL_{\Theta}r \nonumber\\
    &= O\left(\left(\frac{wH}{\alpha} + w\Cupperbound D_\Theta\right)\log(T) + \left((H+2\Cupperbound) \sqrt{\frac{wd}{\alpha} \log(UT^{\frac{3}{2}}\log(T))} + L_{\Theta}\right)\sqrt{T}\right) \nonumber \\
    &\qquad + O\left(\frac{w\Cupperbound}{\alpha}\Th\left(\frac{1}{|\mathcal{N}_{r}(\Theta)| T\log(T)}\right)\right), \nonumber
\end{align}
where the term $\frac{3}{T}$ accounts for the failure of the event $\E_2$.
\end{rproofof}

\subsection{Regret Lower Bounds}
\label{app_proof_lower_bound_new}

We begin this section by proving the lower bound stated in~\Cref{theorem_lower_bound_new}. We then extend this result to the setting in which the cost is \emph{not fully observed} by the algorithm, and show that the lower bound increases by a factor of $\sqrt{1/\alpha}$. Consequently, the regret bound in~\Cref{thm:regret_bound} is also tight in $\alpha$.

\RegretLowerBoundNew*

\begin{rproof}
We first show in parts (i)--(iii) below that, for any $T \ge \max\{128\log(2)d, \frac{2560\log(40)}{3}\}$, for any algorithm, there exists a rich-feedback Lipschitz bandit instance with action space $\A = [0,2]^d$ and width $w$ for which the regret is lower bounded by
\[
\Regret(T)=
\Omega\left(\min\left\{\sqrt{wdT}, T^{\frac{d+1}{d+2}}\right\}\right).
\]
Thus, viewing this bandit instance as a simplified MDP, we have $\Theta=\A=[0,2]^d$.
We then extend this bandit instance to an infinite-horizon average-cost MDP satisfying \Cref{assumption_continuity,assumption_restart,assumption_MRP} in parts (iv)--(vi). For consistency with the literature, we present the bandit instance in terms of reward maximization. The same construction applies to cost minimization under mild modifications.

\paragraph{(i) The Hard Bandit Instance Family.}
We begin by presenting the Lipschitz bandit instance family, inspired by~\citet{slivkins2011contextual}. Consider the arm space $\A=[0,2]^d$, together with a discretization radius $r < 1$ to be specified later. For simplicity, we assume $\frac{1}{r}$ is an integer. Denote $\V = \{r\cdot (z_1, \ldots, z_d)\mid z_i \in \{0, 1, 2, \ldots, \frac{1}{r} - 1\}\}$ as a $r$-packing grid of $[0,1]^d$ under the $\ell_\infty$ norm. Let $v_j$ for $j \in [|\V|]$ denote the $j$-th element in $\V$. For each $v_j \in \V$, define the function
\begin{equation}
\label{eq_lowerbound_cost_design_new}
    \phi_{v_j}(a) = \left[1 - \frac{2\|v_j - a\|_{\infty}}{r}\right]^+.
\end{equation}
Note that, for any $a\in [0,1]^d \subset \A$, if there exists $v\in\V$ such that $\phi_v(a)>0$, then $\phi_{v'}(a)=0$ for all $v'\in\V\setminus\{v\}$.

We now consider $|\V|+1$ instances, which we index by $j\in[|\V|]\cup\{0\}$. For instance $j=0$ (hereafter referred to as environment $P_0$), let $B_{t,v}$ be a Bernoulli random variable for each $t\in[T]$ and $v\in\V$, satisfying
\(
  B_{t,v}\sim \Bern\bigl(\frac12\bigr).
\)
For each instance $j \in [|\V|]$ (hereafter referred to as environment $P_j$), we define $B_{t,v}$ analogously, except that
\[
  B_{t,v}\sim
  \begin{cases}
    \Bern\bigl(\frac12+\Delta\bigr), & v = v_j,\\
    \Bern\bigl(\frac12\bigr),         & v\neq v_j.
  \end{cases}
\]
We assume that the collection $\{B_{t,v}\}_{t,v}$ is independent across $t$ and $v$.  We specify $\Delta$ later.  We define the reward function separately for $a \in [0,1]^d$ and $a \in [0,2]^d \setminus [0,1]^d$:
\begin{itemize}
    \item For any action $a \in [0, 1]^d$, under environment $P_j$,  define the observed reward function at time $t$ by
\[
    Y_{t}(a) = \frac{1}{2} + \frac{1}{4} \sum_{v \in \V}\phi_v(a) (2B_{t,v} - 1).
\]
    \item For any action $a \in [0,2]^d \setminus [0,1]^d$, for every environment $P_j$ we let 
\(
Y_t(a) = \frac{1}{2} + \frac{1}{4} (2B_{t, a} - 1),
\)
where $B_{t,a} \sim \Bern\bigl(\frac12\bigr)$. The collection $\{B_{t,a}\}$ is mutually independent across $t$ and $a$. 
\end{itemize}

\noindent Note that under any environment $P_j$ and for any $a \in \A = [0, 2]^d$, we have $Y_t(a)\in[1/4,3/4]$ for all $t\in[T]$. 
Also, for this instance family, every arm $a\in[0,2]^d$ is optimal under $P_0$, whereas $v_j$ is the unique optimal arm under $P_j$ for each $j\in[|\V|]$.

Whenever $\Delta/r$ is uniformly bounded, the expected reward is Lipschitz continuous in the arms under each environment $P_j$. In particular, since $\phi_{v_j}(a)=0$ for all $a\in[0,2]^d\setminus[0,1]^d$, we have for every $a\in\A$,
\begin{equation}
\label{eq_lowerbound_expected_return_new}
    \Ex[Y_t(a)]=\frac{1}{2}+\frac{1}{2}\Delta\phi_{v_j}(a).
\end{equation}
The desired Lipschitzness therefore follows from the Lipschitz continuity of $\phi_{v_j}(\cdot)$. The Lipschitzness under $P_0$ is straightforward.

We next establish regret lower bounds on this bandit instance class by designing $\Delta$ and $r$ carefully to satisfy the Lipschitzness and information order with width $w$. We discuss two cases, $w \le |\V|$ and $w > |\V|$, respectively.

\paragraph{(ii) Regret Lower Bound When $w \le |\V|$.}
We first consider the case $w \le |\V|$. Note that $w$ need not be a bounded constant; $|\V|$ depends on $r$, and $r$ itself may depend on $T$. We defer the explicit condition ensuring $w \le |\V|$ to the end of this subsection. For simplicity, we further assume that $w$ divides $|\V|$.
Since $w \le |\V|$, partition $\V$ arbitrarily into $w$ disjoint groups $\V_1,\ldots,\V_w$ of {\bf equal size}. For any $i\in[w]$, each element $v\in\V_i$ can be equivalently indexed by a pair $(i,j)$, where $j\in[|\V_i|]$ denotes the position of $v$ within $\V_i$. We also use $(i,j)$ to denote this corresponding element of $\V$.

For each $v\in\V$, define
\begin{equation}
\label{relation_partition_and_A}
    \A_v=\{a\in\A:\phi_v(a) > 0\}.
\end{equation}
The sets $\{\A_v\}_{v \in \V}$ are mutually disjoint by~\cref{eq_lowerbound_cost_design_new}. Additionally, we have $\bigcup_{ i = 1}^w \bigcup_{j = 1}^{|\V_i|}\A_{(i, j)} \subset [0,1]^d$, since \cref{eq_lowerbound_cost_design_new} implies that $\phi_v(a)=0$ for every $a \in [0,2]^d \setminus [0,1]^d$ and $v \in \V$. Also, note that $\bigcup_{ i = 1}^w \bigcup_{j = 1}^{|\V_i|}\A_{(i, j)}$ does not cover all of $[0,1]^d$, e.g., $(1, 1, \ldots, 1)$ does not belong to this union.

We construct the information order by partitioning the informative arms into $w$ feedback groups. For each $i \in [w]$, define
\[
\A_{\V_i} := \bigcup_{v \in \V_i} \A_v.
\]
Whenever the algorithm selects an arm $a \in \A_{\V_i}$, it observes the full collection $\{B_{t,v} : v \in \V_i\}$. Hence, from a single observation in $\A_{\V_i}$, the rewards of all arms in the same group can be counterfactually evaluated, while no information about the variables $\{B_{t,v} : v \in \V_{i'}\}$ for $i' \neq i$ is revealed.  For any arm $a \in [0,1]^d \setminus \bigcup_{i=1}^w \A_{\V_i},$ we have $\phi_v(a)=0$ for every $v \in \V$, and therefore $Y_t(a)=1/2$ deterministically.  Finally, for arms $a \in [0,2]^d \setminus [0,1]^d$, we assume that the corresponding variables $B_{t,a}$ are observable regardless of the selected arm. Thus, these arms also require no additional exploration. Consequently, one arm from each of the $w$ sets $\A_{\V_i}$ is sufficient and necessary to counterfactually evaluate all arms, so the resulting information order over $\A=[0,2]^d$ has width $w$.

% The information order is constructed as follows. For any $t\in[T]$, if the algorithm selects an arm in $\A_{\V_i} := \bigcup_{j = 1}^{|\V_i|}\A_{(i, j)}$, it observes $\{B_{t,v}\}_{v\in\V_i}$. Thus, rewards within the same piece $\A_{\V_i}$ are counterfactually estimable, whereas rewards across different pieces of $\{\A_{\V_i}\}_{i \in [w]}$ are not. $\{B_{t,a}\}$ for every $a \in [0,2]^d \setminus [0,1]^d$ are always observable. \srscomment{Huh? $B_{t,a}$ here? Must be typo and should be $B_{t,v}$ for $v \not\in \mathcal{V}_i$}
% The one-step reward for all arms belonging to $[0,1]^d \setminus \bigcup_{ i = 1}^w \bigcup_{j = 1}^{|\V_i|}\A_{(i, j)}$ is always fixed as $1/2$ thus observable due to~\cref{eq_lowerbound_expected_return_new}. Thus, the resulting information order over $[0, 2]^d$ has width $w$. \srscomment{Sorry I didn't get this. One-step reward is still random?}

For the instance class $\{P_j\}_{j \in [|\V|]\cup \{0\}}$ equipped with the constructed information order above, we establish the following result in~\Cref{lemma_lower_bound_new}. \Cref{lemma_lower_bound_new} largely builds on the argument of~\citet{chen2023interpolating}, who study a finite-armed bandit problem partitioned into multiple groups, with full feedback available within each group. We extend their argument to our continuous setting with $\A = [0, 2]^d.$

\begin{restatable}{lemma}{LemmaBAI} 
\label{lemma_lower_bound_new}
    For any partition $\{\V_i\}_{i \in [w]}$ of $\V$, assume $\bigcup_{v\in\V}\A_v \subset [0,1]^d$, $\{\A_v\}_{v\in\V}$ are mutually disjoint, and for any $t \in [T]$, playing any $a \in \A_{\V_i}, \forall i \in [w]$, does not reveal $\{B_{t,v}\}_{v \notin \V_i}$.
    The feedback structure in $[0, 2]^d \setminus \bigcup_{ i = 1}^w \bigcup_{j = 1}^{|\V_i|}\A_{(i, j)}$ can be arbitrary, but reveals no $\{B_{t,v}\}_{v\in\V}$.
    Let $\Delta = \sqrt{\frac{\sum_{i = 1}^w \log(1 + |\V_i|)}{T}}$.
    Assume $|\V_i| \ge 2$ for all $i \in [w]$, and $\Delta \le 1/8$. Assume $T \ge \frac{2560\log(40)}{3}$.
    Then for any algorithm, there exists $j \in [|\V|] \cup \{0\}$ such that under environment $P_j$,
    \[
        \Regret_j(T)= \Omega\left(\sqrt{T\sum_{i = 1}^w \log(1 + |\V_i|)}\right),
    \]
where $\Omega(\cdot)$ hides only absolute constants, and we write $\Regret_j$ to emphasize that the regret is evaluated under environment $P_j$.
\end{restatable}
\noindent We defer the proof of~\Cref{lemma_lower_bound_new} into~\Cref{proof_of_lemma_lower_bound_new}. 

We now study the range of $w$ such that there exists some $\Delta$ and $r$ satisfying the requirements in~\Cref{lemma_lower_bound_new}, the Lipschitzness, as well as $w \le |\V|$. Set $r = 4\Delta$ so that Lipschitzness holds.  Since $|\V| = r^{-d}$ and each $\V_i$ for $i \in [w]$ has equal size, we have
\(
|\V_i| = \frac{|\V|}{w} = \frac{1}{w(4\Delta)^d}.
\)
Therefore, the choice of $\Delta$ required by \Cref{lemma_lower_bound_new} is equivalent to
% \srscomment{Q: What is required for Lipschitzness? $\Delta r>0$? A: No, we need $\Delta/r$ to be bounded. Under $P_0$, every arm has expected reward $1/2$, so Lipschitzness holds trivially. Under $P_j$, every arm outside $\A_{v_j}$ still has expected reward $1/2$. The function $\phi_{v_j}$ creates a triangular bump centered at $v_j$: its peak has expected reward $1/2+\Delta/2$, and its width is of order $r$. Thus, the slope of the bump is of order $\Delta/r$ (actually the slope is exactly $1/2\Delta/r$) which must be bounded.}
% \srscomment{I guess my point is you only need $\Delta > 0$ and $r > 0$ right? Or something else?}
\begin{equation}
\label{eq_delta_fixed_point}
    \Delta^2 = \frac{w}{T}\log\left(1+\frac{1}{w(4\Delta)^d}\right).
\end{equation}
The left-hand side of \cref{eq_delta_fixed_point} is strictly increasing in $\Delta$, while the right-hand side is strictly decreasing in $\Delta$. Also, both sides are unbounded for $\Delta \in (0, \infty)$. Hence, \cref{eq_delta_fixed_point} admits a unique positive solution. We next upper bound this unique solution.
Define 
\(
    \bar\Delta = \frac{1}{8w^{1/d}} \le \frac{1}{8}.
\)
Thus,
\(
    \frac{1}{w(4\bar\Delta)^d} = 2^d.
\)
Hence, by the monotonicity discussed above, the solution of~\cref{eq_delta_fixed_point} satisfies $\Delta\le\bar\Delta$ whenever
\(
    \bar\Delta^2 \ge \frac{w}{T}\log(1+2^d),
\)
or equivalently,
\begin{equation}
\label{eq_small_width_condition_intermediate}
    T \ge 64\cdot w^{1+2/d}\log(1+2^d).
\end{equation}
Since
\(
    \log(1+2^d) \le (d+1)\log (2) \le 2d\log (2)
\)
for any $d\ge1$, a sufficient condition for
\cref{eq_small_width_condition_intermediate} is that
\(
    T \ge 128\log (2)\,d\,w^{1+2/d}.
\)
Thus, there exists an absolute constant $c_w>0$ such that
\begin{equation}
\label{eq_width_range}
    w \le c_w \left(\frac{T}{d}\right)^{\frac{d}{d+2}}
\end{equation}
implies $\Delta\le\bar\Delta \le \frac{1}{8}$. Also, for any $i \in [w]$,
\(
    |\V_i| = \frac{1}{w(4\Delta)^d} \ge \frac{1}{w(4\bar\Delta)^d} = 2^d \ge 2.
\)
Consequently, $|\V| \ge 2^dw \ge w$, so $\V$ can indeed be partitioned into $w$ nonempty pieces, as required.

Thus it is valid to apply~\Cref{lemma_lower_bound_new} in our constructed instance family $\{P_j\}_{j \in [|\V|]\cup \{0\}}$ with the selected value of $\Delta$ and $r$ as above, whenever 
\(
    w \le c_w \left(\frac{T}{d}\right)^{\frac{d}{d+2}}.
\)
Since $|\V_i| \ge 2^d$ for every $i\in[w]$,
\(
    \sum_{i=1}^w\log(1+|\V_i|) \ge w\log(1+2^d) \ge wd\log (2).
\)
Therefore, from~\Cref{lemma_lower_bound_new}, there is an instance $P_j$, under which 
\[
    \Regret_j(T)= \Omega\left(\sqrt{T\sum_{i = 1}^w \log(1 + |\V_i|)}\right) = \Omega(\sqrt{wdT}).
\]

\paragraph{(iii) Regret Lower Bound When $w > |\V|$.}
We now consider the complementary case $w > |\V|$. In this regime, the construction in part~(ii) cannot be applied directly, since $\V$ cannot be partitioned into $w$ nonempty groups. Instead, we construct the hard instance using a smaller ``effective width''
\[
w' = c_w \left(\frac{T}{d}\right)^{\frac{d}{d+2}} < w,
\]
as in~\cref{eq_width_range}. By construction, $\V$ can be partitioned into $w'$ nonempty groups while preserving the Lipschitz condition, so \Cref{lemma_lower_bound_new} applies with $w'$ in place of $w$. 
We also note that, since $T \ge 128\log(2)d$ and the derivation of~\cref{eq_width_range} shows that $(128\log(2))^{-\frac{d}{d + 2}}$ is a valid choice of $c_w$, we have $w' \ge 1$. Thus, the definition of the ``effective width'' $w'$ is valid.

It remains to enlarge the resulting information order from width $w'$ to width $w$ without changing the statistical difficulty of the hard instance. The partition above induces an information order of width $w'$ on
\[
\bigcup_{v\in\V}\A_v.
\]
We use the region $[0,2]^d \setminus [0,1]^d$ to contribute the remaining width $w-w'$. Since this region contains infinitely many arms, we may partition it into $w-w'$ mutually non-estimable groups, with feedback chosen so that playing any such arm reveals no information about $\{B_{t,v}\}_{v\in\V}$. Thus, these additional groups increase the width without helping distinguish among the environments $P_j$.

Finally, arms in $[0,1]^d \setminus \bigcup_{v\in\V}\A_v$
have reward identically equal to $1/2$ and therefore do not affect the width. Hence, by \Cref{def_width}, the resulting information order on $\A=[0,2]^d$ has width $w$.
Thus, applying~\Cref{lemma_lower_bound_new} with respect to $w'$ implies that there exists an instance $P_j, j \in [|\V|] \cup \{0\}$, under which
\[
\Regret_j(T)= \Omega\left(\sqrt{T\sum_{i = 1}^{w'} \log(1 + |\V_i|)}\right) = \Omega(\sqrt{w'dT}) = \Omega(d^{\frac{1}{d+2}}T^{\frac{d+1}{d+2}}) = \Omega(T^{\frac{d+1}{d+2}}),
\]
where the third equality follows by plugging in $w' = c_w \left(\frac{T}{d}\right)^{\frac{d}{d+2}}$, and the last equality follows from the fact that $1 \le d^{\frac{1}{d+2}} \le 2$ for all $d > 1$.

Combining (ii) and (iii) implies that there exists an instance $P_j$, for some $j \in [|\V|] \cup \{0\}$, under which
\[
\Regret_j(T)= \Omega\left(\min\left\{\sqrt{wdT}, T^{\frac{d+1}{d+2}}\right\}\right).
\]

\paragraph{(iv) Extension to General \(H\).}
Finally, we construct our MDP hard instance with an arbitrary bound $H$ on the bias, which reduces to learning the bandit instance above.  For simplicity we assume $T/H$ is an integer.

We consider an MDP with finite episode length $H$, i.e. after $H$ timesteps the state transitions back to a fixed starting state $s_1$~\citep{puterman2014markov}. 
Note that, since the rewards in our constructed bandit instance in part~(i) are bounded in $[0,1]$, the resulting MDP has bias span at most $H$.
Following the bandit instance described in parts (i)--(iii), let $\Theta=[0,2]^d$ and set the action space of the MDP to $\A=\Theta$.
The state space for the MDP is:
\[
\S = \{s_1\} \cup \bigl\{(\theta,b,h): \theta\in\Theta, b\in \left[\frac14, \frac34\right], h\in[H] \bigr\}.
\]
The first index corresponds to a fixed initial state. The second component corresponds to an action, observed reward, as well as the timestep in the episode.  The dynamics for the MDP are as follows.

The initial state is $s_1$.  At $s_1$, selecting action $a \in \Theta$ yields an observed cost $C$, following the dynamics in (i)--(iii). 
The corresponding latent variables $\{B_{t,v}\mid v \in \V\}$ and $\{B_{t,a}\mid a \in [0,2]^d\setminus [0,1]^d\}$ are revealed simultaneously, as described in parts~(ii) and~(iii).
The state then transitions to $(a, C, 2)$.  From this point forward, the only valid action is $a$, yielding cost of $C$ and transitioning to state $(a, C, h+1)$.  This repeats $H$ steps until the system returns to state $s_1$.

\paragraph{(v) Verifying Assumptions.}
As mentioned in (iv), the bias span of the MDP is upper bounded by $H$.  Moreover, \Cref{assumption_continuity,assumption_MRP} hold trivially, with Lipschitzness following directly from the argument in parts~(i)--(iii).
Returning to $s_1$ every $H$ periods ensures Assumption \ref{assumption_restart}. The width $w$ of the instance class is also established already in part (ii)--(iii).

\paragraph{(vi) Establishing Lower Bound.} With the MDP defined above, we are ready to put together the pieces to establish the lower bound.
Note that the optimal policy $\theta^*$ satisfies $g_{\theta^*} = \mu^*$, the mean of the optimal action for the underlying bandit instance.  Thus we can rewrite regret via:
\begin{align*}
    \Regret(T) & = \sum_{t} \Exp{C_t(S_t, A_t)} - T\mu^* \\
    & \geq H \Omega\left(\min\left\{\sqrt{wdT/H}, (T/H)^{\frac{d+1}{d+2}}\right\}\right) = \Omega\left(\min\left\{\sqrt{HwdT}, H^{\frac{1}{d+2}}T^{\frac{d+1}{d+2}}\right\}\right),
\end{align*}
where the inequality follows by applying the corresponding bandit regret lower bound with horizon $T/H$.
\end{rproof}

\begin{remark}
\label{remark_sqrtH_discrepancy}
Our upper bound in~\Cref{thm:regret_bound} is not minimax optimal up to $\sqrt{H}$ (or $H^{\frac{d+1}{d+2}}$). This gap is well studied in the literature; see, for example, \citet{auer2008near,boone2024achieving}. It can be closed using variance-aware analyses~\citep{boone2024achieving}. Since the primary purpose of this work is to introduce the information order, we leave this refinement for future work, as it requires a more delicate analysis of lower-order terms, as in~\citet{azar2017minimax}.
\end{remark}

The lower bound above does not reflect the dependence on $\alpha$.
To capture the \(\alpha\) dependence, we therefore analyze a family of MDPs where costs are revealed only on an \(\alpha\)-fraction of time steps.  We show by explicit construction that with access to exactly \(\alpha T\) cost observations, any algorithm incurs \(\Omega(\sqrt{HwdT/\alpha})\) regret, when $w$ and $d$ are bounded constants. Thus our results are tight in $\alpha$ as well. Again, we state~\Cref{theorem_lower_bound_new_alpha_dependence} under the normalization \(C(s,a) \in \Delta[0,1]\).

\begin{theorem}
\label{theorem_lower_bound_new_alpha_dependence}
For any fixed bounded integers $d$, $w$, and $H$, for any $T \ge \max\{128\log(2)Hd/\alpha, \frac{2560\log(40)}{3}H/\alpha\}$, for any algorithm, there exists an MDP with continuous state and action spaces and a compact policy class $\Theta \subset \RR^d$ satisfying an information order of width $w$ and \Cref{assumption_continuity,assumption_restart,assumption_MRP}, such that
\[
    \Regret(T) = \Omega(\sqrt{HdwT/\alpha}),
\]
where $\Omega(\cdot)$ only hides absolute constants.
\end{theorem}

\noindent Recall that when $w$ and $d$ are bounded, the lower bound in~\Cref{theorem_lower_bound_new} is $\Omega(\sqrt{HwdT})$, since this term is smaller than $\Omega\left(H^{\frac{1}{d+2}}T^{\frac{d+1}{d+2}}\right)$ for sufficiently large $T$.

\begin{rproof}
Similar to the proof of \cref{theorem_lower_bound_new}, we organize the proof in several parts.  For simplicity, we assume that $1 / \alpha$ is an integer.

\paragraph{(i) Extension to General $\alpha$ via Reward Censoring.} We begin by modifying the MDP from the proof of \cref{theorem_lower_bound_new} as follows.  First, whenever the chain visits $s_1$ and selects an action $a$, the algorithm observes the cost $C$ which is incurred for the next $H$ timesteps (as in part~(iv) in the proof of~\cref{theorem_lower_bound_new}).  However, for the next $H (\frac{1}{\alpha} - 1)$ timesteps, the algorithm incurs additional costs according to their selected action $a$ that are {\em unobserved}.

\paragraph{MDP Definition.} We again consider an MDP with policy class and action space $\Theta=\A=[0,2]^d$.
The state space for the MDP is:
\[
\S = \{s_1\} \cup \bigl\{(\theta,b,h, m): \theta\in\Theta, b\in \left[\frac14, \frac34\right], h\in[H], m \in [1/\alpha] \bigr\}.
\]
The first index corresponds to a fixed initial state.  The second component corresponds to an action, observed reward, as well as the within-block time index $h$ and the block index $m$.  We use $h$ to repeat the rewards $H$ times, and $m$ to denote blocks.
The cost is observed in the first block ($m=1$), whereas the costs in the remaining $1/\alpha-1$ blocks are unobserved.

The dynamics for the MDP are as follows. The initial state is $s_1$.  
In state $s_1$, selecting action $a \in \Theta$ yields an observed cost $C$, following the dynamics in (i)--(iii) in the proof of~\Cref{theorem_lower_bound_new}. 
The corresponding latent variables $\{B_{t,v}\mid v \in \V\}$ and $\{B_{t,a}\mid a \in [0,2]^d\setminus [0,1]^d\}$ are revealed simultaneously according to the feedback structure. 
The state then transitions to $(a,C,2,1)$, corresponding to playing action $a$, observing cost $C$, step $h = 2$ within the episode, and in block $m = 1$.  From this point forward, the only valid action is $a$.  In state $(a,C,h,m)$, if the current step $h < H$, the MDP transitions to $(a,C,h+1,m)$, incurring the same cost $C$. If $h=H$, then the MDP transitions to $(a,C',1,m+1)$ if $m+1\le 1/\alpha$, and to $s_1$ otherwise, where $C'$ denotes a \emph{new} but \emph{unobserved} sample from the reward distribution associated with the same action $a$.

\paragraph{(ii) Verification of Information Order.} 
To distinguish $\CGbarprime(\mathcal{H}_{\theta}^T)$ and $\Gbarprime(\H_{\theta'}^{\alpha T})$ carefully, we argue that $\mathcal{H}_{\theta}^T$ in $\CGbarprime(\mathcal{H}_{\theta}^T)$ and $\mathcal{H}_{\theta}^T$ in $\Gbarprime$ have different meanings.  In $\CGbarprime(\mathcal{H}_{\theta}^T)$, the trajectory $\mathcal{H}_{\theta}^T$ refers to the measurability over observations, but in $\Gbarprime$, it is used for the empirical estimation of the true gain function $g_\theta$. Thus in our bandit cases above, although we collect samples from a length $T$ trajectory, to empirically estimate the gain function $g_\theta$, the length of ``usable'' trajectory is only of length $\alpha T$. 
Thus, our constructed bandit examples above satisfy the distributional order with factor $\alpha$, due to reward censoring, by letting $\CGbarprime$ be the empirical estimate of $g_\theta$ only over the observable samples.

\paragraph{(iii) Verifying Assumptions.}
As in the proof of~\Cref{theorem_lower_bound_new}, \Cref{assumption_continuity} holds trivially, and the Lipschitz bound follows directly.
Returning to $s_1$ every $H / \alpha$ timesteps ensures \Cref{assumption_restart}. Since the costs are bounded within $[0,1]$, the span of the bias is also bounded by $H$, as each realized cost, observable or not, affects the process for at most $H$ time steps.

\paragraph{(iv) Establishing Lower Bound.} With the MDP defined above, we are ready to put together the pieces to establish the lower bound.  Note again that the optimal policy $\theta^*$ satisfies $g_{\theta^*} = \mu^*$, the mean of the optimal action for the underlying bandit instance.  Thus,
\begin{align*}
    \Regret(T) & = \sum_{t=1}^T \Exp{C_t(S_t, A_t)} - T \mu^* \\
    & \geq \frac{H}{\alpha} \sqrt{wd \frac{\alpha T}{H}} = \sqrt{\frac{H wd T}{\alpha}},
\end{align*}
where the inequality follows from the corresponding bandit regret lower bound with horizon $\alpha T/H$. Moreover, the assumption $T \ge \max\{128\log(2)Hd/\alpha, \frac{2560\log(40)}{3}H/\alpha\}$ in~\Cref{theorem_lower_bound_new_alpha_dependence} guarantees that the bandit horizon is at least $\max\{128\log(2)d, \frac{2560\log(40)}{3}\}$, as required; see the proof of~\Cref{theorem_lower_bound_new}.
\end{rproof}

\begin{remark}
    \Cref{theorem_lower_bound_new_alpha_dependence} establishes that our regret in \cref{thm:regret_bound} is minimax optimal with respect to $\alpha$.  This is because \ALG (and our result in \cref{thm:regret_bound}) allows for reward censoring.  Indeed, our algorithm only relies on counterfactual estimates up to a factor of $\alpha$ for every policy in $\Theta_k$ for epoch $k$, including the policy that is actually implemented.  As a result, the algorithm remains valid even if rewards for all policies are partially censored.
\end{remark}

\subsubsection{Proof of~\Cref{lemma_lower_bound_new}}
\label{proof_of_lemma_lower_bound_new}

\LemmaBAI*

\begin{rproof}
The proof is structured as follows. 
In part (i) below (\Cref{lemma_lower_bound_part_1}), we show that if an algorithm achieves regret
\(
    o\left(\sqrt{T\sum_{i = 1}^w \log(1 + |\V_i|)}\right)
\)
uniformly over all instances $P_j$ for $j\in [|\V|] \cup \{0\}$, then it induces a $(\frac{1}{2}\Delta, 0.05)$-PAC algorithm for best-arm-identification (BAI) over the same class of instances that stops by time $T_1^*$. That is, at time $T_1^*$, with probability at least $0.95$, the induced BAI algorithm returns an arm whose expected reward is {\bf strictly} less than $\frac12 \Delta$ below that of an optimal arm.
In part (ii), we show that under mild conditions, any such $(\frac{1}{2}\Delta,0.05)$-PAC BAI algorithm must run at least $T_2^*$ rounds in expectation; see~\Cref{lemma_lower_bound_part_2}.
We then complete the proof by establishing the contradiction that $T_2^* > T_1^*$.

\paragraph{(i) Characterization of $T_1^*$.}
Intuitively, the characterization of $T_1^*$ follows from that if an algorithm incurs sufficiently small regret, then it must quickly identify an arm whose expected reward is close to that of an optimal arm.

Suppose that an algorithm satisfies, uniformly over all
$P_j$ for $j \in [|\V|] \cup\{0\}$,
\(
    \Regret_j(T) = o\left(\sqrt{T\sum_{i = 1}^w \log(1 + |\V_i|)}\right).
\)
Concretely, assume that there exists some $T$ such that for all instances $j \in [|\V|] \cup\{0\}$,
\begin{equation}
\label{eq_uniform_small_regret}
    \Regret_j(T) \le \frac{3}{409600 \log (40)} \sqrt{T\sum_{i=1}^w\log(1+|\V_i|)}.
\end{equation}
Recall that
\(
    \Delta = \sqrt{\frac{\sum_{i=1}^w\log(1+|\V_i|)}{T}}.
\)
Define
\begin{equation}
\label{eq_T1_star}
    T_1^* := \left\lfloor \frac{3T}{2560\log(40)} \right\rfloor \ge 1,
\end{equation}
where the inequality is from the assumption that $T \ge \frac{2560\log(40)}{3}$.

From~\cref{eq_lowerbound_expected_return_new}, for each $j \in [|\V|]$, let
\(
    \mu_j(a) :=\frac{1}{2}+\frac{1}{2}\Delta\phi_{v_j}(a)
\)
denote the expected reward of arm $a \in \A$ under environment $P_j$. Since $\phi_{v_j}(v_j)=1$, the optimal arm under $P_j$ is $v_j$, and the optimal reward under $P_j$ is
\(
    \mu_j^* = \mu_j(v_j) = \frac12+\frac{\Delta}{2}.
\)
Recalling the definition of $\A_{v_j}$ in~\cref{relation_partition_and_A}, we further define
\(
    \C_{v_j} := \left\{a\in\A_{v_j}:\phi_{v_j}(a)> \frac12 \right\}.
\)
Note that $\{\C_{v_j}\}_{j \in [|\V|]}$ are mutually disjoint.
For any $a \notin \C_{v_j}$, we have $\phi_{v_j}(a) \le 1/2$. Thus, under $P_j$, 
\begin{align}
\label{eq_reward_gap}
    \mu_j^*-\mu_j(a) = \frac{\Delta}{2}\bigl(1-\phi_{v_j}(a)\bigr) \ge \frac{\Delta}{4}.
\end{align}

Let $\textsf{ALG}_{\text{REG}}$ denote the original regret minimization algorithm. Based on $\textsf{ALG}_{\text{REG}}$, we construct a $(\frac{1}{2}\Delta,0.05)$-PAC BAI algorithm $\textsf{ALG}_{\text{BAI}}$ that uses $T_1^*$ rounds.
Concretely, $\textsf{ALG}_{\text{BAI}}$ follows $\textsf{ALG}_{\text{REG}}$ exactly for all $t \in [T_1^*]$. At time $T_1^*$, it draws
\(
    U \sim \Unif([T_1^*])
\)
independently. If $a_U \in \C_{v_j}$ for some $j \in [|\V|]$, then $\textsf{ALG}_{\text{BAI}}$ outputs $v_j$. Since the sets $\{\C_{v_j}\}_{j \in [|\V|]}$ are mutually disjoint, such $j$ is unique. If $a_U$ does not belong to any $\C_{v_j}$, then $\textsf{ALG}_{\text{BAI}}$ outputs an arbitrary fixed arm in $\V$. We denote the output of $\textsf{ALG}_{\text{BAI}}$ by $\hat a_{\textsf{BAI}}$.

Fix an arbitrary environment $P_j$ for $j \in [|\V|]$. Let $\H_{T_1^*}$ denote the stochastic trajectory under $\textsf{ALG}_{\text{REG}}$ until $T_1^*$. Conditional on $\H_{T_1^*}$, we have that
\[
    \sum_{t = 1}^{T_1^*} \mu_j^*-\mu_j(a_t) \ge \sum_{t = 1}^{T_1^*} \Ind{a_t \notin \C_{v_j}} \cdot \frac{\Delta}{4} + \Ind{a_t \in \C_{v_j}} \cdot 0 \ge \frac{1}{4}\Delta \Pr_{P_j}(\hat a_{\textsf{BAI}} \neq v_j \mid \H_{T_1^*})T_1^*,
\]
where $\Pr_{P_j}(\hat a_{\textsf{BAI}} \neq v_j \mid \H_{T_1^*})$ denotes the probability that the BAI algorithm fails to output the optimal arm given $\H_{T_1^*}$. The first inequality follows from~\cref{eq_reward_gap}, and the second follows from the definition of $U$.

Taking the expectations over the trajectories, one obtains
\begin{align*}
    \Pr_{P_j}(\hat a_{\textsf{BAI}} \neq v_j) &\le \frac{4}{\Delta T_1^*} \Regret_j(T_1^*) 
    \le \frac{4}{\Delta T_1^*} \Regret_j(T) 
    \le \frac{4}{\Delta T_1^*}\frac{3}{409600 \log(40)}\sqrt{T\sum_{i = 1}^{w}\log(1 + |\V_i|)}  \\
    &\le \frac{4}{\Delta} \frac{2560\log(40) \cdot 2}{3T} \frac{3}{409600 \log(40)}\sqrt{T\sum_{i = 1}^{w}\log(1 + |\V_i|)} 
    = 0.05,
\end{align*}
where the first inequality follows from the definition of regret, the second follows from the fact that $T_1^* \le T$ (see~\cref{eq_T1_star}) and the per-step regret is nonnegative, the third follows from~\cref{eq_uniform_small_regret}, and the fourth follows by substituting the definition of $T_1^*$ from~\cref{eq_T1_star} and using
\(
T_1^* = \left\lfloor \frac{3T}{2560\log(40)} \right\rfloor \ge \frac{1}{2}\frac{3T}{2560\log(40)},
\)
where the inequality holds because $T_1^* \ge 1$.

In summary, we have the following lemma:
\begin{lemma}
\label{lemma_lower_bound_part_1}
    Define $T_1^*$ as in~\cref{eq_T1_star}. Fix some $T \ge \frac{2560\log(40)}{3}$. If there exists an algorithm $\textsf{ALG}_{\text{REG}}$ that achieves
    \(
        \Regret_j(T) \le \frac{3}{409600 \log (40)} \left(\sqrt{T\sum_{i = 1}^w \log(1 + |\V_i|)}\right)
    \)
    uniformly over all instances $j$, then there exists a $(\frac{1}{2}\Delta,0.05)$-PAC BAI algorithm $\textsf{ALG}_{\text{BAI}}$ that stops by time $T_1^*$.
\end{lemma}

\paragraph{(ii) Characterization of $T^*_2$.}
Intuitively, the characterization of $T_2^*$ follows from a standard minimax change-of-measure argument. To distinguish $P_0$ from every alternative environment $P_j$ for $j \in [|\V|]$, the rich-feedback structure requires each partition $\mathcal{V}_i$, $i \in [w]$, to be sampled sufficiently often under $P_0$.
We first establish the required change-of-measure result for one fixed piece $i \in [w]$.

\begin{lemma}
\label{lemma_distinguishing_lower_bound}
Fix $i \in [w]$. Consider the environments $P_0$ and $\{P_j \mid v_j\in\V_i\}$. Let $T_3^*$ denote the stopping time of any algorithm that identifies the underlying environment with probability at least $0.925$. Define
\(
    t^* :=\frac{\log(|\V_i|+1)}{4\Delta^2}.
\)
Then under environment $P_0$,
\(
    \Pr_{0}(T_3^* \ge t^*) \ge 0.1.
\)
\end{lemma}
\noindent Note that identifying the underlying environment is harder than best-arm-identification. Under $P_0$, environment identification requires the output $\widehat J=0$, whereas every arm is optimal for BAI.

\begin{rproof}
We consider a new environment $\bar{P}_i$, which is a uniform combination of all environments $P_j$ for  $v_j \in \V_i$. Let $\E_{t^*}$ be the event that the algorithm stops by $t^*$ and outputs $\widehat{J} = 0$. If~\Cref{lemma_distinguishing_lower_bound} does not hold, i.e., the algorithm stops at $T_3^*$ and
\(
    \Pr_{P_0}(T_3^* \ge t^*) < 0.1,
\)
then
\(
    \Pr_{P_0}(\E_{t^*}^c) < (1 - 0.925) + 0.1 = 0.175.
\)
Since $\E_{t^*}$ is measurable at time $t^*$, the Bretagnolle--Huber inequality~\citep{lattimore2020bandit} implies that
\begin{equation}
\label{lower_bound_bh_inequality}
    \Pr_{\bar{P}_i}(\E_{t^*}) \ge \frac12 \exp(-\KL\left(P_{\bar{P}_i} \middle\|P_0\right)) - \Pr_{P_0}(\E_{t^*}^c) \ge \frac12 \exp(-\KL\left(P_{\bar{P}_i} \middle\|P_0\right)) - 0.175.
\end{equation}

We next bound $\KL\left(P_{\bar{P}_i} \middle\|P_0\right)$. 
For the fixed $i \in [w]$, recall that each $v \in \V_i$ has a unique index $(i, j)$. For each $t \in [t^*]$, let $X_t := (B_{t,v_{(i, 1)}},\ldots,B_{t,v_{(i, |\V_i|)}}) \in \{0,1\}^{|\V_i|}$ denote the full-feedback observation from partition $\V_i$.
Let $p_{(i, j)}$ and $p_0$ denote the probability mass functions of $(X_1, \ldots, X_{t^*})$ under $P_{(i,j)}$ and $P_0$, respectively. The probability mass function under the mixture distribution $\bar{P}_i$ is
\(
    p_{\bar{P}_i}(x) = \frac{1}{|\V_i|} \sum_{j=1}^{|\V_i|} p_{(i, j)}(x).
\)
For each $j \in [|\V_i|]$, define the likelihood ratio $L_{(i, j)}(x) := p_{(i, j)}(x)/p_0(x)$. Since $P_{(i, j)}$ and $P_0$ differ only in one coordinate, for any realization $x=(x_1, \ldots, x_{t^*})$,
\begin{equation}
\label{KL_term_first_order}
    L_{(i, j)}(x) 
    = \prod_{t=1}^{t^*} \frac{\left(\frac12+\Delta\right)^{x_{t,(i, j)}}\left(\frac12-\Delta\right)^{1-x_{t,(i, j)}}}{\left(\frac12\right)^{x_{t,(i, j)}}\left(\frac12\right)^{1-x_{t,(i, j)}}}
    = \prod_{t=1}^{t^*}(1+2\Delta)^{x_{t,(i, j)}}(1-2\Delta)^{1-x_{t,(i, j)}}.
\end{equation}
Note that $\Ex_{P_0}[L_{(i, j)}]=1$. Moreover, for distinct $j, j' \in [|\V_i|]$, $L_{(i, j)}$ and $L_{(i, j')}$ are independent under $P_0$, since $B_{t, v_{(i, j)}}$ and $B_{t, v_{(i, j')}}$ are independent for all $t \in [t^*]$. Therefore,
\(
    \Ex_{P_0}[L_{(i, j)}L_{(i, j')}] = \Ex_{P_0}[L_{(i, j)}]\Ex_{P_0}[L_{(i, j')}] = 1.
\)
For the second order terms, independence across $t$, together with~\cref{KL_term_first_order}, gives that for all $j \in [|\V_i|]$,
\begin{equation}
\label{eq_KL_term_second_order}
    \Ex_{P_0}[L_{(i, j)}^2] = \prod_{t=1}^{t^*}    \Ex_{P_0}\left[(1+2\Delta)^{2x_{t,(i, j)}}(1-2\Delta)^{2(1-x_{t,(i, j)})}\right]
    =\prod_{t=1}^{t^*}\left(\frac12(1+2\Delta)^2 + \frac12(1-2\Delta)^2\right)
    =(1+4\Delta^2)^{t^*}.
\end{equation}
Thus,
\begin{align*}
    \KL\left(\bar{P}_i\middle\|P_0\right)
    &= \sum_x p_{\bar{P}_i}(x) \log\left(\frac{p_{\bar{P}_i}(x)}{p_0(x)}\right)
    \le \log\left(\sum_x \frac{p_{\bar{P}_i}(x)^2}{p_0(x)}\right)
    = \log\left(\frac{1}{|\V_i|^2} \sum_{j=1}^{|\V_i|} \sum_{j'=1}^{|\V_i|} \sum_x \frac{p_{(i, j)}(x)p_{(i, j')}(x)}{p_0(x)}\right)\\
    &= \log\left(\frac{1}{|\V_i|^2} \sum_{j=1}^{|\V_i|} \sum_{j'=1}^{|\V_i|} \Ex_{P_0}[L_{(i, j)}L_{(i, j')}]\right)
    = \log\left(\frac{|\V_i|(|\V_i| - 1) + |\V_i|(1 + 4\Delta^2)^{t^*}}{|\V_i|^2}\right) \\
    &= \log\left(\frac{|\V_i| - 1 + (1 + 4\Delta^2)^{t^*}}{|\V_i|}\right),
\end{align*}
where the inequality is from Jensen's inequality, and the fourth equality is from~\cref{eq_KL_term_second_order}.
Recall that
\(
    t^* = \frac{\log(|\V_i|+1)}{4\Delta^2}.
\)
Since $1+x < e^x$ for $x > 0$,
\(
    (1+4\Delta^2)^{t^*} < \exp(4\Delta^2t^*) = |\V_i| + 1.
\)
It follows that for any $|\V_i|$,
\begin{equation}
\label{eq_mixture_KL_final}
    \KL\left(\bar{P}_i\middle\|P_0\right) < \log\left(\frac{|\V_i| - 1 + |\V_i| + 1}{|\V_i|}\right) = \log(2).
\end{equation}

Combining~\cref{eq_mixture_KL_final,lower_bound_bh_inequality} yields
\(
    \Pr_{\bar{P}_i}(\E_{t^*}) > 0.25 - 0.175 = 0.075.
\)
Therefore, there exists some $j''\in[|\V_i|]$ such that
\(
    \Pr_{P_{(i,j'')}}(\E_{t^*})>0.075.
\)
This contradicts the assumption that the algorithm outputs $(i,j'')$ under $P_{(i,j'')}$ with probability at least $0.925$.
\end{rproof}

For the same instance class with same partition $\{\V_i\}_{i \in [w]}$, we next use \Cref{lemma_distinguishing_lower_bound} to lower bound the expected number of implementations in each $\A_{\V_i}$ for $i \in [w]$ for any $(\frac{1}{2}\Delta, 0.05)$-PAC BAI algorithm. For each $i \in [w]$, let
\(
    N_i := \sum_{t=1}^{T_{\textsf{BAI}}} \Ind{A_t\in\A_{\V_i}}
\)
denote the number of times that the BAI algorithm implements an arm in
$\A_{\V_i}$ before stopping, where $T_{\textsf{BAI}}$ denotes its stopping time.

\begin{lemma}
\label{lemma_lower_bound_part_2}
Assume $|\V_i| \ge 2$ for all $i \in [w]$ and $\Delta \le \frac{1}{8}$. Then there exists an absolute constant $c_B = \frac{3}{1280\log(40)}$ such that, for every $i \in [w]$, any $(\frac{1}{2}\Delta, 0.05)$-PAC BAI algorithm must satisfy
\begin{equation}
\label{eq_bai_lower_bound_T_2}
    \Ex_{P_0}[N_i] \ge c_B\frac{\log(1+|\V_i|)}{\Delta^2}.
\end{equation}
\end{lemma}

\begin{rproof}
We consider separately the cases where $|\V_i|$ is sufficiently large and where $|\V_i|$ is small.
\paragraph{Case 1: $\log(|\V_i|+1) \ge 64\log(40)$.}
For this case, we use the given BAI algorithm to construct an algorithm that identifies the underlying environment among $P_j$ for $j\in [|\V|] \cup \{0\}$, and then apply \Cref{lemma_distinguishing_lower_bound} to derive \cref{eq_bai_lower_bound_T_2}. Concretely, given any $(\frac{1}{2}\Delta,0.05)$-PAC BAI algorithm, we construct an environment identification algorithm $\textsf{ALG}_{\text{IDEN}}$ as follows. The algorithm $\textsf{ALG}_{\text{IDEN}}$ first follows the given BAI algorithm until it stops. Let $\hat a_{\textsf{BAI}}$ denote the arm returned by the BAI algorithm. We then implement $\hat a_{\textsf{BAI}}$ for an additional
\(
    r_\Delta=\frac{8\log(40)}{\Delta^2}
\)
rounds and let $\widehat\mu_{\hat a}$ denote its empirical average reward over these rounds. If
\(
    \widehat\mu_{\hat a}\ge \frac12 + \frac{\Delta}{4},
\)
then $\textsf{ALG}_{\text{IDEN}}$ outputs $j$ if $\hat a_{\textsf{BAI}} \in \A_{v_j}$ for some $j \in [|\V|]$, and outputs
$0$ otherwise. If
\(
    \widehat\mu_{\hat a}<\frac12 + \frac{\Delta}{4},
\)
then $\textsf{ALG}_{\text{IDEN}}$ outputs $0$.

We next show that $\textsf{ALG}_{\text{IDEN}}$ identifies the underlying environment with probability at least $0.925$, allowing us to apply~\Cref{lemma_distinguishing_lower_bound}.
First consider environment $P_0$. Under $P_0$, every arm has expected
reward $1/2$. Thus, conditional on any output $\hat a_{\textsf{BAI}}$ of the BAI algorithm, Hoeffding's inequality~\citep{lattimore2020bandit} implies
\[
    \Pr_{P_0}\left(\widehat\mu_{\hat a} \ge \frac12+\frac{\Delta}{4} \right) \le \exp\left(-2r_\Delta\left(\frac{\Delta}{4}\right)^2\right)
    = \exp\left(-\frac{r_\Delta\Delta^2}{8}\right) 
    = \frac1{40}.
\]
Therefore, under $P_0$, $\textsf{ALG}_{\text{IDEN}}$ outputs $0$
with probability at least
\(
    1-\frac1{40} > 0.925.
\)

Next, consider any environment $P_j$ for $j\in[|\V|]$. Since the given BAI algorithm is $(\frac{1}{2}\Delta,0.05)$-PAC, $\hat{a}_{\textsf{BAI}} = v_j$ with probability at least $0.95$. Conditional on the event $\hat a_{\textsf{BAI}}=v_j$, we have
\(
    \Ex_{P_j}[Y_t(v_j)] = \frac12+\frac{\Delta}{2}.
\)
Hence, again, by Hoeffding's inequality,
\(
    \Pr_{P_j}\left(\widehat\mu_{\hat a} < \frac12+\frac{\Delta}{4} \mid \hat a_{\textsf{BAI}}=v_j\right) \le \frac1{40}.
\)
A union bound gives that 
\[
    \Pr_{P_j}\left(\textsf{ALG}_{\text{IDEN}}\text{ correctly identifies}P_j\right) \ge 0.925.
\]
Therefore, $\textsf{ALG}_{\text{IDEN}}$ identifies the underlying
environment with probability at least $0.925$ under every $P_j$, $j \in [|\V|] \cup \{0\}$. Whenever $\log(|\V_i|+1)\ge 64\log(40)$, we have
$r_\Delta\le t^*/2$ (see the definition of $t^*$ in~\Cref{lemma_distinguishing_lower_bound}). Applying \Cref{lemma_distinguishing_lower_bound} to $\textsf{ALG}_{\text{IDEN}}$ gives
\[
    0.1 \le \Pr_{P_0}(N_i+r_\Delta\ge t^*) \le \Pr_{P_0}(N_i\ge t^*/2).
\]
Note that here we implicitly replace $T_3^*$ in~\Cref{lemma_distinguishing_lower_bound} by $N_i$. This is because, when distinguishing $P_0$ from the alternatives $\{P_j:v_j\in\V_i\}$, observations obtained outside $\A_{\V_i}$ have the same distribution under all hypotheses and therefore do not contribute to the KL divergence.
By Markov's inequality,
\(
    \Ex_{P_0}[N_i] \ge \frac{t^*}{20} = \frac{\log(|\V_i|+1)}{80\Delta^2}.
\)

\paragraph{Case 2: $\log(|\V_i|+1) < 64\log(40)$.}
For this case, note that for the $(\frac{1}{2}\Delta, 0.05)$-PAC BAI algorithm, for each $i \in [w]$, there exists some $v_j \in \V_i$ such that
\(
    \Pr_{P_0}(\hat a_{\textsf{BAI}} = v_j) \le \frac{1}{|\V_i|} \le \frac12,
\)
where the last inequality is from the assumption that $|\V_i| \ge 2, \forall i \in [w]$. Also, by the definition, 
\(
    \Pr_{P_j}(\hat a_{\textsf{BAI}} = v_j) \ge 0.95.
\)
Data processing inequality together with Pinsker's inequality imply that
\[
    \KL \left(P_0\middle\|P_j\right) \ge \KL\left(\Bern\left(\Pr_{P_0}(\hat a_{\textsf{BAI}}=v_j)\right)\middle\| \Bern\left(\Pr_{P_j}(\hat a_{\textsf{BAI}}=v_j)\right)\right) \ge 2(0.95-0.5)^2 = 0.405.
\]
On the other side, 
\(
     \KL \left(P_0\middle\|P_j\right) \le \frac{8}{3}\Delta^2\Ex_{P_0} [N_i],
\)
where we use the fact that $\Delta \le 1/8$ and for Bernoulli random variables with means $\mu_1, \mu_2 \in [1/4,3/4]$,
\(
    \KL \left(\Bern(\mu_1)\middle\|\Bern(\mu_2)\right) \le \frac{8}{3}(\mu_1-\mu_2)^2.
\)
We omit the proof of this. Combining the two inequalities above implies that for all $i \in [w]$,
\(
    \Ex_{P_0}[N_i] \ge \frac{3\cdot0.405}{8\Delta^2} > \frac{3}{20\Delta^2}.
\)
Since $\log(|\V_i| + 1) < 64 \log(40)$, it follows that
\(
    \Ex_{P_0}[N_i] \ge \frac{3}{1280\log(40)} \frac{\log(|\V_i|+1)}{\Delta^2}.
\)

Combining the two cases and letting 
\(
c_B = \min\left\{\frac1{80}, \frac{3}{1280\log(40)}\right\} = \frac{3}{1280\log(40)}
\)
complete the proof.
\end{rproof}

We are now ready to prove~\Cref{lemma_lower_bound_new}.
As shown in~\Cref{lemma_lower_bound_part_1}, if there exists an algorithm $\textsf{ALG}_{\text{REG}}$ satisfying
    \(
        \Regret_j(T) \le \frac{3}{409600 \log (40)} \left(\sqrt{T\sum_{i = 1}^w \log(1 + |\V_i|)}\right)
    \)
uniformly over all instances $j$, then there exists a $(\frac{1}{2}\Delta,0.05)$-PAC BAI algorithm $\textsf{ALG}_{\text{BAI}}$ that stops by time $T_1^* = \left\lfloor\frac{3}{2560 \log (40)} T\right\rfloor$. 
On the other hand, \Cref{lemma_lower_bound_part_2} implies that any such $(\frac{1}{2}\Delta,0.05)$-PAC BAI algorithm must play
\[
    T_2^* = \sum_{i \in [w]}\Ex_{P_0}[N_i] \ge c_B\frac{\sum_{i \in [w]} \log(1+|\V_i|)}{\Delta^2} = c_B T = \frac{3}{1280\log(40)}T \ge 2 T_1^*
\]
rounds. The contradiction $T_2^* > T_1^*$ completes the proof.
\end{rproof}

\section{Proofs in Section \ref{sec_case_studies}}\label{appendix_case_studies}

\subsection{Single-Retailer Inventory Control with Positive Lead Time (\cref{case_study:inventory})}
\label{sec:inventory_control_app_proofs}

\BaseStockInformationOrder*
\begin{rproofof}{\cref{lem:base_stock_policy_orders}}
We show the proof by establishing that $\H_{\theta'}^{T} \in \H_{\theta}^T.$  To do so, we start off with the following lemma.
\begin{lemma}\label{ic_lemma1}
Suppose that $\{S_t\}_{t = 1}^T$ is a state trajectory collected under a base stock policy $\pi_{\theta}$ and $\{S_t'\}_{t = 1}^T$ collected under a base stock policy $\pi_{\theta'}$ over a fixed sequence of demands $(D_1, \ldots, D_T)$ with $\theta \geq \theta'$.  Then if
\begin{itemize}
    \item $S_0 \geq S'_0$
    \item $\theta - \norm{S_0}_1 \geq \theta' - \norm{S'_0}_1 \geq 0$,
\end{itemize}
we have that for all $t \leq T$:
\begin{itemize}
    \item $S_t \geq S'_t$
    \item $\theta - \norm{S_t}_1 \geq \theta' - \norm{S'_t}_1 \geq 0$.
\end{itemize}
\end{lemma}
\begin{rproof}
We show the result by induction over the timestep $t$.  For the base case all of the statements are true by assumption, so we only have to focus on the step case for $t$ implying $t+1$.  We abbreviate IH as the induction hypothesis.

Recall that for any $t \le T$, the state $S_t = (I_t, Q_{t - L}, \ldots, Q_{t - 1}) \in \mathbb{R}^{L+1}$, refers to on-hand inventory and $L$ unfinished orders in the pipeline. For clarity, we denote the components as $(S_{t,0}, \ldots, S_{t,L}) \triangleq (I_t, Q_{t - L}, \ldots, Q_{t - 1})$, and similarly, $S'_t = (S'_{t,0}, \ldots, S'_{t,L}) = (I'_t, Q'_{t - L}, \ldots, Q'_{t - 1})$. Then, under the system dynamics given in \cref{eq_ic_dynamics}, and assuming the demands $D_1, \ldots, D_T$ are fixed, we have for all $t < T$:
\begin{align*}
S_{t+1} = ((S_{t,0} + S_{t,1} - D_t)^+, S_{t,2}, \ldots, S_{t, L}, (\theta - \norm{S_t}_1)^+) \\
S'_{t+1} = ((S'_{t,0} + S'_{t,1} - D_t)^+, S'_{t,2}, \ldots, S'_{t, L}, (\theta' - \norm{S'_t}_1)^+).
\end{align*}
First we show that $S_{t+1} \geq S'_{t+1}$.  For any index $j = 1, \ldots, L-1$ we have that
\[
S_{t+1, j} = S_{t, j+1} \geq S'_{t, j+1} = S'_{t+1, j},
\]
where the inequality in the middle uses the IH.  For the index $j = L$ we have that:
\[
S_{t+1,L} = (\theta - \norm{S_t}_1) \geq (\theta' - \norm{S'_t}_1) = S'_{t+1, L},
\]
where again the last line uses the IH.  For the first index $j = 0$ we have:
\[
S_{t+1, 0} = (S_{t,0} + S_{t,1} - D_t)^+ \geq (S'_{t,0} + S'_{t,1} - D_t)^+ = S'_{t+1,0}.
\]

Lastly, we show that 
\[
\theta - \norm{S_{t+1}}_1 \geq \theta' - \norm{S'_{t+1}}_1 \geq 0.
\]
The fact that $\norm{S'_{t+1}}_1 \leq \theta'$ is clear by the induction hypothesis and the definition of the dynamics. We further note that
\begin{align*}
    \theta - \norm{S_{t+1}}_1 = (S_{t,0} +  S_{t,1} - (S_{t,0} + S_{t,1} - D_t)^+), \quad 
    \theta' - \norm{S'_{t+1}}_1 = (S'_{t,0} +  S'_{t,1} - (S'_{t,0} + S'_{t,1} - D_t)^+).
\end{align*}

Now we compare the expressions under different realizations of \( D_t \):
\begin{enumerate}
    \item If \( D_t \le S'_{t,0} + S'_{t,1} \), then since \( S'_{t,0} + S'_{t,1} \le S_{t,0} + S_{t,1} \), we also have \( D_t \le S_{t,0} + S_{t,1} \). Thus:
    \[
        (S_{t,0} +  S_{t,1} - (S_{t,0} + S_{t,1} - D_t)^+) = D_t = (S'_{t,0} +  S'_{t,1} - (S'_{t,0} + S'_{t,1} - D_t)^+).
    \]
    \item If \( D_t \ge S_{t,0} + S_{t,1} \), then since \( S'_{t,0} + S'_{t,1} \le S_{t,0} + S_{t,1} \), we have \( D_t \ge S'_{t,0} + S'_{t,1} \). Thus:
    \[
        (S_{t,0} +  S_{t,1} - (S_{t,0} + S_{t,1} - D_t)^+) = S_{t,0} +  S_{t,1} \ge S'_{t,0} +  S'_{t,1} = (S'_{t,0} +  S'_{t,1} - (S'_{t,0} + S'_{t,1} - D_t)^+).
    \]
    \item If \( S'_{t,0} + S'_{t,1} < D_t < S_{t,0} + S_{t,1} \), then:
    \[
        (S_{t,0} +  S_{t,1} - (S_{t,0} + S_{t,1} - D_t)^+) = D_t > S'_{t,0} +  S'_{t,1} = (S'_{t,0} +  S'_{t,1} - (S'_{t,0} + S'_{t,1} - D_t)^+).
    \]
\end{enumerate}
Combining all of the different cases we see that $\theta - \norm{S_{t+1}}_1 \geq \theta' - \norm{S'_{t+1}}_1$ as needed.
\end{rproof}
Now, in order to show that the base stock policies satisfy the information order, we recall that the starting state $s_1 = (0, \ldots, 0)$.  Hence, by \cref{ic_lemma1} for any two base stock levels $\theta' \leq \theta$ we see that the on-hand inventory under base stock policy $\pi_\theta$ ($S_{t,0}$) is always larger than that under $\pi_{\theta'}$ ($S'_{t,0}$).  Thus, we just need to show that $\tilde{C}(S'_t)$ is measurable with respect to $\H_{\theta}^T$.  We show by induction on $t$ that: $S'_t \in \H_{\theta}^T$ and $\tilde{C}(S'_t) \in \H_\theta^T$.  Indeed, this is true for $t = 0$. For $t > 0$:
\[
S'_{t+1} = ((S'_{t,0} + S'_{t,1} - D_t)^+, S'_{t,2}, \ldots, (\theta' - \norm{S'_t}_1)^+).
\]
However, by the induction step, we know that $S'_t$ is measurable with respect to $\H_{\theta}^T$.  Moreover, since $S'_t \leq S_t$ via \cref{ic_lemma1} we know if $S_{t+1,0} = (S_{t,0} + S_{t,1} - D_t)^+ > 0$, then $S'_{t+1,0}$ is trivially in $\H_{\theta}^T$. Otherwise, if $S_{t+1,0} = 0$ then $D_t \geq S_{t,0} + S_{t,1} \geq S'_{t,0} + S'_{t,1}$ and so $S'_{t+1,0} = 0$ as well (implying $S'_{t+1,0} \in \H_{\theta}^T$).  Overall, we see that $S'_t \in \H_{\theta}^T$ as required.

Similarly for the costs,
\[
\tilde{C}(S'_t) = h(S'_{t,0} + S'_{t,1} - \min\{S'_{t,0} + S'_{t,1}, D_t\}) 
    - p \min\{S'_{t,0} + S'_{t,1}, D_t\}.
\]
However, $\min\{S'_{t,0} + S'_{t,1}, D_t\} \in \H_{\theta}^T$ since $S'_t \leq S_t$ via \cref{ic_lemma1}.  Thus we have that $\tilde{C}(S'_t) \in \H_{\theta}^T$.
All together, this implies that $\pi_{\theta'} \preceq \pi_{\theta}$ as needed.
\end{rproofof}

\subsection{Dual Index Policy for the Dual Sourcing Problem (\cref{case_study:dual_index_policy})}
\label{sec:dual_index_app_proofs}

% Recall that in Section \ref{case_study:dual_index_policy}, we denote $l$ as the lead-time of the long-lead channel, and $\beta = P[D_t = 0] > 0$. Now we intend to show that for the dual index policy, the regret is with order $O(\sqrt{\beta^{-3l}T})$. We will give the proof with the following manner: 
% \begin{itemize}
%     \item For dual index policies in dual sourcing problem, we will show that \Cref{assumption_MRP} is satisfied.
%     \item We will show that for each policy set $\Theta'$, there exits a policy $\theta'$ s.t. for any policy $\theta \in \Theta'$, we have $\theta \preceq_P \theta'$. Specifically, we have $w = 1, \alpha = \frac{2}{\beta^l}, d = 2$, where the definition of these notations follow Definition \ref{definition_general_order}.
% \end{itemize}
% When the two statements above are established, following Theorem \ref{thm:regret_bound}, we have the desired regret upper bound as $O(\sqrt{\beta^{3l}T})$.
% \zzedit{
\subsubsection{Non-Convexity of the Problem}\label{appendix_sec_convexity}
Unlike the convex cost curve of the lost‐sales single‐channel base‐stock policy in \citet{agrawal2019learning}, and the empirically observed non‐convexity of the backlog dual index policy in \citet{veeraraghavan2008now}, the convexity of the long‐run average cost under the lost‐sales dual index policy (our setting in \cref{case_study:dual_index_policy}) has received little attention in the literature.  In \Cref{dual_index_convexity}, we follow \citet{veeraraghavan2008now} to show empirically that this cost is non‐convex in the base‐stock parameters.  Notably, when the expedited base‐stock level \(z_e^\theta=0\), the dual index policy reduces to the single‐channel base-stock policy and recovers the convex cost curve of \citet{agrawal2019learning} (see the \(z_e^\theta=0\) trace in~\Cref{dual_index_convexity}).

\begin{figure}[!t]
  \centering
  \includegraphics[width=0.6\textwidth]{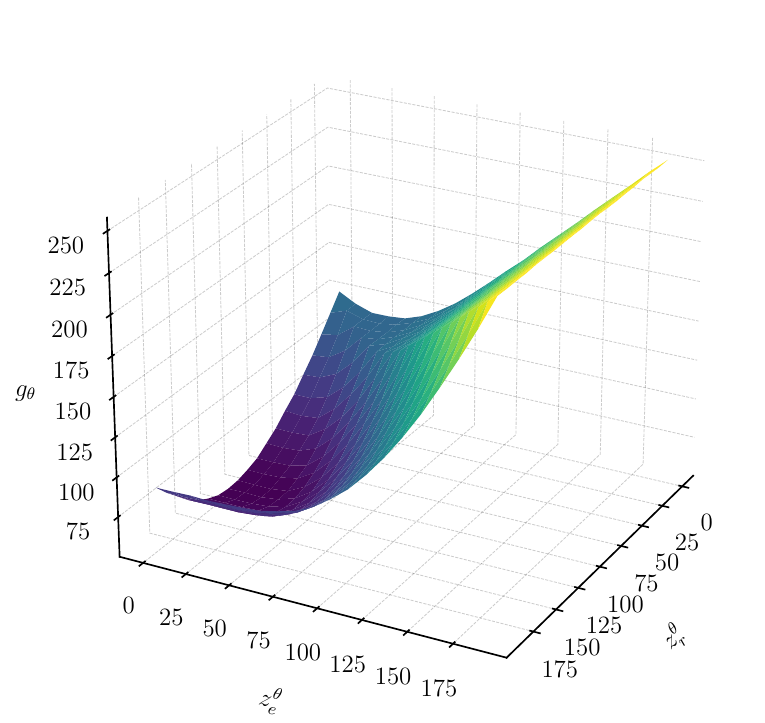}
  \caption{An empirical illustration of non‐convexity in the long‐run average cost with respect to the base‐stock levels \((z_e^{\theta}, z_r^{\theta})\) under the lost‐sales dual index policy.  Parameters are \(L_r=8\), \(L_e=2\), \(c_r=1\), \(c_e=8\), \(h=1\), \(p=10\), and demand follows an exponential distribution with rate $3/40$.  Costs are approximated by sample averages over \(T=10^5\) steps, evaluated on a grid \((z_e^{\theta},z_r^{\theta})\in\{0,5,10,\dots,200\}^2\).}
  \label{dual_index_convexity}
  % \vspace{-1em}
% \end{wrapfigure}
\end{figure}

% }

\subsubsection{Verification of Assumptions} \label{proof_dip_sec1}

Before formally describing the information order over policies, we start by showing two required results to apply \cref{thm:regret_bound}.  First, we provide an upper bound on the bias in terms of $\gamma = \Pr(D_t = 0)$ and the long lead time $L_r$.  Second, we provide conditions for which the cost function $g_\theta$ is Lipschitz.
%Following Proposition 8.1.1 of \citet{puterman2014markov} and Remark 3 of \citet{agrawal2019learning}, we will ignore discussion around the existence of any limits.  Indeed, \citet{agrawal2019learning} establishes that employing an appropriate discretization of state space and action space allows the existence of all limits within our focus.  
Before presenting these results we start with a technical lemma.

\begin{lemma}\label{lem_consecutive_zero}
Define $Y_1$ as the first time having $L_r$ consecutive $0$ demands, i.e.,
\[
    Y_1 = \inf \{t \geq L_r \mid \forall k \in \{0, 1, \ldots, L_r - 1\}, D_{t - k} = 0\},
\]
and $\gamma = \Pr(D_t = 0)$. Then
\[
\Ex[Y_1] = \frac{1 - \gamma^{L_r}}{(1-\gamma)\gamma^{L_r}} \le \frac{1}{(1 - \gamma) \gamma^{L_r}}.
\]
\end{lemma}

One version of proof of \Cref{lem_consecutive_zero} based on induction is given in \citet{ross2014introduction}.

\DipGainSpan*
\begin{rproof}
    Fix any policy $\theta \in \Theta$, and let $s_1, s_2$ be two arbitrary initial states. For any time horizon $T$, let $G_{\theta}(\H_{\theta}^T)(s)$ denote the empirical average cost from initial state $s$ as defined in \Cref{definition_el}. Then:
\[
\left| TG_{\theta}(\H_{\theta}^T)(s_1) - TG_{\theta}(\H_{\theta}^T)(s_2) \right| \le 2\Cupperbound\min\{T, Y_1\} \le 2\Cupperbound Y_1,
\]
where $Y_1$ is the first time $L_r$ consecutive $0$-demand events occur. Since the costs are bounded by $\Cupperbound$, the upper bound of $2\Cupperbound T$ follows immediately.  However, if $Y_1 \leq T$, we note that for all $t' \geq Y_1$ the two systems will be in the same state.  Hence, the difference in costs is similarly upper bounded by $2\Cupperbound Y_1$.  Taking expectations and applying \Cref{lem_consecutive_zero}, we get:
\[
\left| \mathbb{E}\left[TG_{\theta}(\H_{\theta}^T)(s_1) - TG_{\theta}(\H_{\theta}^T)(s_2)\right] \right| \le 2\Cupperbound\mathbb{E}[Y_1] \le \frac{2\Cupperbound}{(1 - \gamma)\gamma^{L_r}}.
\]

It follows that the long-run average cost is uniform:
\[
\left| g_\theta(s_1) - g_\theta(s_2) \right| = \left|\lim_{T \to \infty} \frac{1}{T}  \mathbb{E}\left[TG_{\theta}(\H_{\theta}^T)(s_1) - TG_{\theta}(\H_{\theta}^T)(s_2)\right] \right| \le \lim_{T \to \infty} \frac{2\Cupperbound}{(1 - \gamma)\gamma^{L_r} T} = 0.
\]

Additionally, the span of the bias is bounded:
\begin{align}
    \left| v_\theta(s_1) - v_\theta(s_2) \right| 
    &= \left| \lim_{T \to \infty} \Ex[ TG_{\theta}(\H_{\theta}^T)(s_1) - Tg_\theta - TG_{\theta}(\H_{\theta}^T)(s_2) + Tg_\theta ] \right| \\
    &\le \frac{2\Cupperbound}{(1 - \gamma)\gamma^{L_r}} \label{dip_bias_bound}.
\end{align}

All limits exist, and the interchange of limits and expectations is justified by the uniform integrability. For simplicity, we defer the proof of uniform integrability in \Cref{dip_lem1}.
\end{rproof}

\DipLipschitz*
    \begin{rproof}
    Recall that the policy class is characterized by the tuple $(z_e^{\theta}, z_r^{\theta})$ where $z_e^{\theta}$ and $z_r^{\theta}$ refer to the base-stock level of the short-lead channel and the base-stock level of the long-lead channel, respectively. Hence, the policy class is isomorphic to $\Theta \subset \mathbb{R}^2$.

    Now we show that $g_\theta$ is Lipschitz continuous. Specifically, for any $\theta_1 = (z_e^{\theta_1}, z_r^{\theta_1})$ and $\theta_2 = (z_e^{\theta_2}, z_r^{\theta_2})$ such that $\|\theta_1 - \theta_2\|_{\infty} = \delta$, we intend to show that $|g_{\theta_1} - g_{\theta_2}| \le L_{\Theta}\delta$ for some $L_{\Theta}$.

    Recall that we denote the state at time $t$ as 
    \[
    S_t = [I_t, (Q^r_{t - L_r}, Q^r_{t - L_r + 1}, \ldots, Q^r_{t - 1}), (Q^e_{t - L_e}, Q^e_{t - L_e + 1}, \ldots, Q^e_{t - 1})],
    \]
    where $I_t$ refers to the on-hand inventory at time step $t$, $Q^e_{t}$ refers to the inventory in the short-lead pipeline that was ordered at time $t$, and $Q^r_{t}$ refers to the inventory in the long-lead pipeline that was ordered at time $t$. Since $g_\theta$ is uniform for any $\theta$ (\cref{lem:dual_index_bias}), we  assume that for policy $\theta_1$, the initial state is $[z_r^{\theta_1}, (0, 0, \ldots, 0), (0, 0, \ldots, 0)]$, and the initial state for policy $\theta_2$ is $[z_r^{\theta_2}, (0, 0, \ldots, 0), (0, 0, \ldots, 0)]$. 

    Now denote $C_t^{\theta}$ as the observed cost at time $t$ under policy $\theta$. We introduce a sequence of random variables $(Y_i)_{i \in \mathbb{N}_+}$, where $Y_i$ refers to the interarrival time between $(i-1)$-st occurrence of $L_r$ consecutive periods where demands are all $0$ and the $i$-th occurrence of $L_r$ consecutive periods where demands are all $0$. Formally, for $i \ge 1$, we define $\tau_i = \inf\{t \ge \tau_{i - 1} + L_r \mid D_{t - k} = 0 \text{ for all } k \in \{0, 1, \ldots, L_r - 1\}\}$ and $Y_i = \tau_i - \tau_{i - 1}$, with $\tau_0 = 0$.

    %    
    % if $Y_{i-1}$ occurs at time $t'$ (which implies that $\forall k \in \{0, 1, \ldots, l - 1\}, D_{t' - k} = 0$), then 
    % $$
    %     Y_i = \inf \{t|t \ge t' + l, \forall k \in \{0, 1, \ldots, l - 1\}, D_{t - k} = 0\} - t'
    % $$
    Note that following our definition, $\{Y_i, i \in \mathbb{N}^+\}$ is a sequence of non-negative and i.i.d. random variables. Following convention, if denoting 
    \begin{align}
        N(t) = \sup\left\{n \mid \sum_{i = 1}^n Y_i \le t\right\} \label{eq_renewal_process},
    \end{align}
    then $N(t)$ is a renewal process. By our construction of $Y_i$, we have $\Ex[Y_i] < \infty$, and $\Ex[\sum_{t \in Y_1} \abs*{C_t^{\theta_1} - C_t^{\theta_2}}] \le 2 \Cupperbound \Ex[Y_1] < \infty$. Here we slightly abuse notation by interpreting $\{t\in \mathbb{Z} \mid t \in Y_i,i \le N(T)\}$ as $\{t\in \mathbb{Z} \mid \tau_{i-1} < t \le \tau_i,1 \le i \le N(T)\}$.
        % \srscomment{Also using $t \in Y_i$ for the index which is slightly wrong.}\zzcomment{updated above}

    Furthermore, $\Ex[(\sum_{t = 1}^T (C_t^{\theta_1} - C_t^{\theta_2}))^2] \le 4 \Cupperbound^2 T^2$. 
    Using Markov's inequality we get for all $x > 0$:
    \[
    \Pr\left(\frac{\sum_{t = 1}^T \abs*{C_t^{\theta_1} - C_t^{\theta_2}}}{T} > x\right) \le \frac{4\Cupperbound^2}{x^2}.
    \]
    
    Using this uniform integrability, we are free to switch the limit and the expectation. Thus:
    \begin{align}
        \abs*{g_{\theta_1} - g_{\theta_2}}&= \abs*{\Ex[\lim_{T \to \infty} \frac{1}{T}\sum_{i = 1}^{N(T)} \sum_{t \in Y_i} (C_t^{\theta_1} - C_t^{\theta_2})] +\Ex[\lim_{T \to \infty} \frac{1}{T}\sum_{t = \tau_{N(T)} + 1}^{T} (C_t^{\theta_1} - C_t^{\theta_2})]} \label{eq:b3.0} \\
            &\le \Ex[\lim_{T \to \infty} \frac{1}{T}\sum_{i = 1}^{N(T)} \sum_{t \in Y_i} \abs*{C_t^{\theta_1} - C_t^{\theta_2}}] \\
            &= \frac{\Ex[\sum_{t \in Y_1} \abs*{C_t^{\theta_1} - C_t^{\theta_2}}]}{\Ex[Y_1]}\label{eq:b3.1}.
    \end{align}
    Thus, to provide a bound on $|g_{\theta_1} - g_{\theta_2}|$ it suffices to bound $\Ex[\sum_{t \in Y_i} |C_t^{\theta_1} - C_t^{\theta_2}|]$, which we do by induction. 

    Before providing a bound, we give several iterations below to illustrate our regenerative cycle construction.
    We slightly abuse the notations and let $I_t(\theta_i)$ as the on-hand inventory at $t$ for policy $\pi_i$, (and $Q^e_t(\theta_i)$ and $Q^r_t(\theta_i)$ accordingly), and $S_t(\theta_i)$ as the state at time step $t$ for policy $\theta_i$.
    We further denote $D_t$ as the realized demand at time $t$.  Then we have:
    \begin{align*}
        S_1(\theta_1) = [z_r^{\theta_1}, (0, 0, \ldots, 0), (0, 0, \ldots, 0)] \\
        S_1(\theta_2) = [z_r^{\theta_2}, (0, 0, \ldots, 0), (0, 0, \ldots, 0)],
    \end{align*}
    by our construction described earlier.
    For $t = 2$:
    \begin{align*}
        S_2(\theta_1) = [[z_r^{\theta_1} - D_1]^+, (0, 0, \ldots, 0), (0, 0, \ldots, 0)] \\
        S_2(\theta_2) = [[z_r^{\theta_2} - D_1]^+, (0, 0, \ldots, 0), (0, 0, \ldots, 0)].
    \end{align*}

    Next at $t = 3$ we have:
    \begin{align*}
        S_3(\theta_1) = [I_3(\theta_1), (0, 0, \ldots, Q^e_2(\theta_1)), (0, 0, \ldots, Q^r_2(\theta_1))] \\
        S_3(\theta_2) = [I_3(\theta_2), (0, 0, \ldots, Q^e_2(\theta_2)), (0, 0, \ldots, Q^r_2(\theta_2))].
    \end{align*}

    In order to finish the proof, we establish the following:
    \begin{lemma}
    \label{lem:recursive_inventory_bound}
    For any $t \leq Y_1$, we have that
        \begin{align*}
        |I_t(\theta_1) - I_t(\theta_2)| + \sum_{i = 1}^{L_e} |Q^e_{t - i}(\theta_1) - Q^e_{t - i}(\theta_2)| + \sum_{i = 1}^{L_r} |Q^r_{t - i}(\theta_1) - Q^r_{t - i}(\theta_2)|\le W(t)\delta,
    \end{align*}
    where $W(1) = 1$ and $W(t) = 4 W(t-1) + 3$.
    Correspondingly, 
    \begin{itemize}
        \item $|I_t(\theta_1) - I_t(\theta_2)| \leq W(t) \delta$
        \item $|C_t^{\theta_1} - C_t^{\theta_2}| \leq (h + p + c_r + c_e) W(t+1) \delta.$
    \end{itemize}
    \end{lemma}
    \begin{rproof}
    We show the proof by induction on $t$.  For the base case when $t = 1$ we have that $I_1(\theta_1) = z_r^{\theta_1}$ and $I_1(\theta_2) = z_r^{\theta_2}$.  Thus, due to the fact that $|z_r^{\theta_1} - z_r^{\theta_2}| \le \delta$, $|I_1(\theta_1) - I_1(\theta_2)| \leq \delta$.  
    Next we note that $C_1^{\theta_1} \leq (h+p)I_1(\theta_1)$, so the bound on $|C_1^{\theta_1} - C_1^{\theta_2}|$ follows immediately.

    Next we show that $t \rightarrow t+1$.  By the induction hypothesis we assume that
    \begin{align}
        |I_t(\theta_1) - I_t(\theta_2)| + \sum_{i = 1}^{L_e} |Q^e_{t - i}(\theta_1) - Q^e_{t - i}(\theta_2)| + \sum_{i = 1}^{L_r} |Q^r_{t - i}(\theta_1) - Q^r_{t - i}(\theta_2)|\le W(t)\delta.\label{eq_dip_ih}
    \end{align}
    Denote by
    \begin{align*}
        IP^e_t(\theta_i) & = I_t(\theta_i) + \sum_{i = 1}^{L_e} (Q^e_{t - i}(\theta_i) + Q^{r}_{t - L_r + L_e - i}(\theta_i)) + Q^{r}_{t - L_r + L_e}(\theta_i) \\
        IP^r_t(\theta_i) & = I_t(\theta_i) + \sum_{i = 1}^{L_e} Q^e_{t - i}(\theta_i) + \sum_{i = 1}^{L_r} Q^r_{t - i}(\theta_i) + Q^e_{t}(\theta_i)
    \end{align*}
    as the total inventory position for the short-lead and long-lead channel respectively at timestep $t$. Note that by definition of the dual index base stock policies, $Q_t^e(\theta_i) = (z_e^{\theta_i} - IP^e_t(\theta_i))^+$ (and similarly with $Q_t^r(\theta_i)$). Then we have
\begin{align*}
  \bigl|IP^e_t(\theta_1) - IP^e_t(\theta_2)\bigr|
  &\le \bigl|I_t(\theta_1) - I_t(\theta_2)\bigr|
    + \sum_{i=1}^{L_e} \bigl|Q^e_{t-i}(\theta_1) - Q^e_{t-i}(\theta_2)\bigr|\nonumber\\
  &\quad + \sum_{i=1}^{L_e} \bigl|Q^r_{t - L_r + L_e - i}(\theta_1)
    - Q^r_{t - L_r + L_e - i}(\theta_2)\bigr|\nonumber\\
  &\quad + \bigl|Q^r_{t - L_r + L_e}(\theta_1)
    - Q^r_{t - L_r + L_e}(\theta_2)\bigr|\nonumber\\
  &\le W(t)\,\delta.
\end{align*}

    Thus due to $|z_e^{\theta_1} - z_e^{\theta_2}| \le \delta$, we have that
    \begin{align}
    |Q^e_{t}(\theta_1) - Q^e_{t}(\theta_2)| \le |(z_e^{\theta_1} - IP^e_t(\theta_1)) - (z_e^{\theta_2} - IP^e_t(\theta_2))| \le (W(t) + 1)\delta,\label{eq_dip_part2}
    \end{align}
    and 
    \begin{align}
    |Q^r_{t}(\theta_1) - Q^r_{t}(\theta_2)| &\le |(z_r^{\theta_1} - IP^r_t(\theta_1)) - (z_r^{\theta_2} - IP^r_t(\theta_2))| \\
    &\le W(t)\delta + |Q^e_{t}(\theta_1) - Q^e_{t}(\theta_2)| + |z_r^{\theta_1} - z_r^{\theta_2}| \\
    &\le (2W(t) + 2)\delta . \label{eq_dip_part3}
    \end{align}
    Note that
    \begin{align*}
    |I_{t+1}(\theta_1) - I_{t+1}(\theta_2)| &= |[I_t(\theta_1) + Q_{t - L_e}^e(\theta_1) + Q_{t - L_r}^r(\theta_1) - D_t]^+ - [I_t(\theta_2) + Q_{t - L_e}^e(\theta_2) + Q_{t - L_r}^r(\theta_2) - D_t]^+| \\
        &\le |I_t(\theta_1) - I_t(\theta_2)| + |Q_{t - L_e}^e(\theta_1) - Q_{t - L_e}^e(\theta_2)| + |Q_{t - L_r}^r(\theta_1) - Q_{t - L_r}^r(\theta_2)|.
    \end{align*}
    Correspondingly, 
    \begin{align*}
        |I_{t+1}(\theta_1) - I_{t+1}(\theta_2)| &+ \sum_{i = 1}^{L_e} |Q^e_{t - i + 1}(\theta_1) - Q^e_{t - i + 1}(\theta_2)| + \sum_{i = 1}^{L_r} |Q^r_{t - i + 1}(\theta_1) - Q^r_{t - i + 1}(\theta_2)| \\
        & \le W(t)\delta +  (W(t) + 1)\delta + (2W(t) + 2)\delta \\ 
        & \le (4W(t) + 3)\delta = W(t+1)\delta,
    \end{align*}
    where the first part is from \eqref{eq_dip_ih}, the second part is from \eqref{eq_dip_part2}, and the third part is from \eqref{eq_dip_part3}. This completes the induction part of \Cref{lem:recursive_inventory_bound}. Furthermore, we directly obtain that $|I_{t}(\theta_1) - I_{t}(\theta_2)| \le W(t)\delta$. Also,
    \begin{align}
        |C_{t}^{\theta_1} - C_{t}^{\theta_2}| &\le (h+p + c_r + c_e)[|I_{t}(\theta_1) - I_{t}(\theta_2)| + |Q^e_{t}(\theta_1) - Q^e_{t}(\theta_2)| + |Q^r_{t}(\theta_1) - Q^r_{t}(\theta_2)|] \\
        &\le (h+p + c_r + c_e)W(t+1)\delta,
    \end{align}
    as over-stocking and shortage penalty only relate to $|I_{t+1}(\theta_1) - I_{t+1}(\theta_2)|$, and purchasing cost only relates to $|Q^e_{t}(\theta_1) - Q^e_{t}(\theta_2)|$ and $|Q^r_{t}(\theta_1) - Q^r_{t}(\theta_2)|$. Here $h$, $p$, $c_r$, $c_e$ refer to the holding cost, the shortage penalty, the long-lead purchasing cost, and the short-lead purchasing cost coefficient, respectively.
    \end{rproof}
    Finally we combine \cref{eq:b3.1} with \cref{lem:recursive_inventory_bound} to establish the Lipschitz continuity.  Note that
    \begin{align*}
        |g_{\theta_1} - g_{\theta_2}| &\le \frac{\Exp{\sum_{t = 1}^{Y_1} |C_t^{\theta_1} - C_t^{\theta_2}|}}{\Exp{Y_1}} \\
        &\le (h + p + c_r + c_e)\delta \frac{\Exp{\sum_{t = 1}^{Y_1} W(t + 1)}}{\Exp{Y_1}} \\
        &= (h + p + c_r + c_e)\delta \frac{\sum_{t = 1}^{\infty} \Pr(Y_1 = t)\sum_{s = 1}^t W(s + 1)}{\Exp{Y_1}} \\
        &\le 7(h + p + c_r + c_e)\delta \frac{\sum_{t = 1}^{\infty} \Pr(Y_1 = t)W(t)}{\Exp{Y_1}}.
    \end{align*}
    
    Also note that the distribution of $Y_1$ is given in \citet{drekic2021number}, and known to be sub-exponential, and $\Ex[Y_1]$ is known to be finite.  Also, $\sum_{t=1}^\infty \Pr(Y_1 = t) W(t)$ is fully characterized by $L_r$ and $\gamma$ by our definition of $Y_1$. Assuming $\sum_{t=1}^\infty \Pr(Y_1 = t) W(t)$ to be finite, the desired Lipschitz continuity follows with
    $$
        L_{\Theta} = 7(h+p+c_r+c_e) \frac{\sum_{t=1}^\infty \Pr(Y_1 = t) W(t)}{\Exp{Y_1}}.
    $$
    Due to the distribution of $Y_1$ given in \citet{drekic2021number}, $\sum_{t=1}^\infty \Pr(Y_1 = t) W(t)$ is finite when $\gamma \to 1$. However, $\sum_{t=1}^\infty \Pr(Y_1 = t) W(t)$ is not necessarily bounded for any $\gamma > 0$. Correspondingly, with $\gamma$ sufficiently large to have a bounded $L_{\Theta}$, the $L_{\Theta}$ term in \Cref{thm:regret_bound} is dominated by $ H\sqrt{ \frac{w d}{\alpha} \log\left( \frac{ U\sqrt{ T} \log (T) }{ \delta } \right)}$ for sufficiently large $T$. For this reason, we omit the explicit \(L_{\Theta}\) term in our regret bound.
\end{rproof}

\begin{remark}\label{remark_dip_lipschitz}
While the exact distribution of $Y_1$ is given in \citet{drekic2021number}, its closed-form is cumbersome to analyze. Instead, we compute the exact value of \(c_\gamma\) numerically in a specific example to assess how restrictive this requirement is for ensuring Lipschitz continuity: when $L_r = 2$, taking $\gamma > 0.96$ suffices. The exact value of \(c_\gamma\) for other hyperparameter settings can be computed in the same manner. Recall that most SKUs in real-world inventories are slow-moving. Thus, our approach still yields $\sqrt{T}$ regret for the majority of those items.

We believe that one can further relax the constraints on $L_r$ and $\gamma$ by carefully analyzing the growth of the inventory gap between any two policies.  Since each exponential jump requires on the order of \(\Omega(L_r)\) increments, the effective growth rate becomes slower (see~\Cref{appendix_case_study_dual_index_empirical_evidence}), weakening the necessary bounds on \(L_r\) and \(\gamma\).  Nevertheless, because the gap still grows exponentially, a fully general Lipschitz proof remains elusive for arbitrary \(L_r\) and \(\gamma\).  We defer a complete characterization of sufficient conditions for Lipschitz continuity for future work.
\end{remark}

\begin{comment}

%\srscomment{Other random thought - is there any result for Lipschitzness of the stationary distribution relative to mild modifications to the transition kernel?  Then use that the two base stock policies induce slightly perturbed transition distributions, this gives close stationary distributions, and hence close average cost? Not certain if there's a possibility of getting the analysis to work out that way. Meyn + Tweedie might have some information on this.}

%We intend to bound the regret based on the argument about the times of on-hand inventory level being extreme,i.e., for policy $(\pi_s, \pi_l)$, the on-hand inventory being $\pi_s + \pi_l$.  \srscomment{Sorry - what are you referring to here as regret?}
%\begin{lemma}
%    Following our definition of $Y_i$ and $N(T)$ above, we have 
    $$
%        \Exp{\lim_{T \to \infty} \frac{[N(T)]}{T}} = \lim_{T \to \infty} \frac{\Ex{[N(T)]}}{T} = \frac{1}{\Ex{[Y_1]}}
    $$
%    And the distribution of $Y_1$ is given in \citet{drekic2021number}.
%\end{lemma}
%\begin{rproof}
%    Notice that $\{Y_i, i = \{1, 2, \ldots\}\}$ is a sequence of i.i.d. random variables. Thus the first equality is from the uniform integrability similarly as given in Lemma \ref{dip_lem1}, and the second inequality is obtained immediately from renewal theory.
%\end{rproof}
\end{comment}

\subsubsection{Existence of Distributional Ordering}\label{proof_dip_sec2}

\DefDipEstimator*

Recall that $\gamma = \Pr(D_t = 0)$. We now present a sequence of lemmas demonstrating how the estimator $\CGbarprime(\H_{\theta}^T)$ defined in \cref{eq_estimator_dual_index} can be leveraged to establish a distributional order. We now define $I_t^{\theta, \dagger} = I_t^\theta + \Order^{\theta, r}_{t - L_r} + \Order^{\theta, e}_{t - L_e}$ as the on-hand inventory at time $t$ after order arrivals.

\begin{lemma}\label{dip_lem4}
    Recall that the initial state is fixed at $s_1$, and consider the system under the fixed policy $\pi_\theta$. Then for any $\delta > 0$, with probability at least $1 - \delta$,
    $$
        |\frac{1}{T}\sum_{t = 1}^T \mathbb{I}(I_t^{\theta, \dagger} = z_r^\theta) - 
        \Ex{[\lim_{T \to \infty} \frac{1}{T}\sum_{t = 1}^T \mathbb{I}(I_t^{\theta, \dagger} = z_r^\theta)]|} \le \frac{2}{1 - \gamma}\left(\frac{1}{\gamma^{L_r}T} + \frac{1}{\gamma^{L_r}}\sqrt{\frac{2\log(2/\delta)}{T}}\right),
    $$
    where $\mathbb{I}(\cdot)$ refers to the indicator function.
\end{lemma}
\begin{rproof}
Recall from \Cref{lem_concentration_lem_shipra} that the empirical estimator concentrates provided the gain is uniform across states and the span of the bias is bounded. We verify these conditions for the cost function defined by $\mathbb{I}(I_t^{\theta, \dagger} = z_r^\theta)$, which indicates whether the on-hand inventory after order arrivals at time $t$ equals $z_r^\theta$. To apply \Cref{lem_concentration_lem_shipra} and prove \Cref{dip_lem4}, it suffices to show: (i) the gain function is uniform across states, and (ii) the bias span is bounded by $\frac{2}{(1 - \gamma)\gamma^{L_r}}$.

However, \Cref{lem:dual_index_bias} only relies on the costs being bounded in $[-\Cupperbound, \Cupperbound]$, which applies under this cost function with $[0,1]$.  Thus, the two conditions are satisfied directly.  Applying \cref{lem_concentration_lem_shipra} yields the bound here.
\end{rproof}

The previous lemma establishes that for an observed trajectory $\H_\theta^T$ of length $T$ under policy $\theta$, we have a lower bound of the number of observations where the on-hand inventory is $z_r^\theta$ after order arrivals. Formally, we have:
\begin{lemma}\label{dip_lem6}
Let $\theta \in \Theta$ be any policy. For any $\delta > 0$, suppose $T \ge \Th(\delta)$, where $\Th(\delta) = \Omega(\frac{\log(1 / \delta)}{(1 - \gamma)^4 \gamma^{4L_r}})$. Then, with probability at least $1 - \delta$, we have:
\[
\sum_{t = 1}^T \mathbb{I}(I_t^{\theta, \dagger} = z_r^\theta) \ge \frac{1 - \gamma}{2} \gamma^{L_r} T.
\]
\end{lemma}
\begin{rproof}
From \Cref{dip_lem4}, we have that with probability at least $1 - \delta$,
\begin{align*}
\sum_{t = 1}^T \mathbb{I}(I_t^{\theta, \dagger} = z_r^\theta) 
&\ge T \cdot \mathbb{E}\left[\lim_{T \to \infty} \frac{1}{T} \sum_{t = 1}^T \mathbb{I}(I_t^{\theta, \dagger} = z_r^\theta)\right]
    - \frac{2T}{1 - \gamma} \left( \frac{1}{\gamma^{L_r} T} + \frac{1}{\gamma^{L_r}} \sqrt{\frac{2 \log(2/\delta)}{T}} \right) \\
&\ge \frac{T}{\mathbb{E}[Y_1]} 
    - \frac{2}{(1 - \gamma)\gamma^{L_r}} 
    - \frac{2}{(1 - \gamma)\gamma^{L_r}} \sqrt{2 \log(2/\delta) T} \\
&\ge (1 - \gamma)\gamma^{L_r} T 
    - \frac{2}{(1 - \gamma)\gamma^{L_r}} 
    - \frac{2}{(1 - \gamma)\gamma^{L_r}} \sqrt{2 \log(2/\delta) T}.
\end{align*}
% \srscomment{Seems like some $T$ factors are disappearing here? How are you going from $1 / \sqrt{T}$ to $\sqrt{T}$?  Or the entire thing is multiplied by $T$ not just the first term? And where is the $Y_1$ coming from?}
The second inequality uses the fact that $\sum_{t = 1}^T \mathbb{I}(I_t^{\theta, \dagger} = z_r^\theta) \ge N(T)$ almost surely (with $N(T)$ from \cref{eq_renewal_process}), as each occurrence of $L_r$ consecutive zero demands implies at least one $t$ with $I_t^{\theta, \dagger} = z_r^\theta$.
%\srscomment{And the relationship between $N(T)$ and $Y_1$?}\zzcomment{my understanding is that this is directly from renewal theory; we have $E[Y_1] < \infty$ in Lemma B.3, thus $\lim_{T \to \infty} \frac{N(T)}{T} = \frac{1}{E[Y_1]}$, and we are free to change limits and expectation due to uniform integrability.} 
The third inequality applies \(\mathbb{E}[Y_1] \le \frac{1}{(1 - \gamma)\gamma^{L_r}}\) from \cref{lem_consecutive_zero}.

To ensure the right-hand side exceeds \(\frac{1 - \gamma}{2} \gamma^{L_r} T\), it suffices that:
\begin{equation}
\sqrt{T} \geq \frac{4 \sqrt{2 \log \left( \tfrac{2}{\delta} \right)} + 4}{(1 - \gamma)^2 \gamma^{2L_r}}.
\label{dip_th_bound}
\end{equation}

Thus, for all \(T \ge \Th(\delta)\) satisfying \cref{dip_th_bound}, we have that
\(
\sum_{t = 1}^T \mathbb{I}(I_t^{\theta, \dagger} = z_r^\theta) \ge \frac{1 - \gamma}{2} \gamma^{L_r} T.
\)
\end{rproof}

Combining \Cref{dip_lem6} with the definition of our counterfactual estimator in \Cref{def_dip_ERM}, we formally establish the distributional equivalence that underpins the information ordering.

\begin{lemma}\label{dip_lem7}
Let $\alpha = \frac{1 - \gamma}{2} \gamma^{L_r}$. For any policy $\theta \in \Theta$, any $\delta > 0$, and any $T \ge \Th(\delta)$ where the definition of $\Th(\delta)$ follows \Cref{dip_lem6}, define the event
\[
\E_4(T, \delta) = \left\{ \sum_{t = 1}^T \mathbb{I}(I_t^{\theta, \dagger} = z_r^\theta) \ge \frac{1 - \gamma}{2} \gamma^{L_r} T \right\}.
\]
Then, for any $\theta' \in \Theta$ satisfying $z_r^{\theta'} \le z_r^\theta$, there exists a coupling of $\H_\theta^T$ and $\H_{\theta'}^{\alpha T}$ such that $\CGbarprime(\H_\theta^T) = G_{\theta'}(\H_{\theta'}^{\alpha T})$ almost surely on $\E_4(T, \delta)$.
\end{lemma}

\begin{rproof}
This result follows directly from the definition of $\CGbarprime$ in \Cref{def_dip_ERM}. Under the event $\E_4(T, \delta)$, the index set $\mathcal{I}_\theta$ contains at least $\alpha T$ time steps at which the on-hand inventory equals $z_r^\theta$ after order arrivals, thus $\CGbarprime(\H_\theta^T) = G_{\theta'}(\tilde{\H}_{\theta'}^{\alpha T})$. At these time steps, the observed sales are uncensored with respect to any policy $\theta'$ such that $z_r^{\theta'} \le z_r^\theta$, and thus can be treated as valid demand samples for estimating $G_{\theta'}$. As a result, the pseudo-trajectory $\tilde{\H}_{\theta'}^{\alpha T}$ constructed from these samples under policy $\theta'$ yields a counterfactual estimate $\CGbarprime(\H_\theta^T)$ that, under a suitable coupling, coincides almost surely with $G_{\theta'}(\H_{\theta'}^{\alpha T})$ on $\E_4(T, \delta)$.
\end{rproof}

\DualIndexOrder*
\begin{rproof}
    We denote $G_{\theta'}(\H_{\theta'}^{\alpha T})$ as $G_{\theta'}$, and $\E_4(T, \delta)$ as $\E_4$ in this proof for simplicity. For any Borel measurable set $B$, we have that
    $$
        \left|\Pr(G_{\theta'} \in B) - \Pr(\CGbarprime \in B)\right| = \left|\Pr(G_{\theta'} \in B, \E_4^c) - \Pr(\CGbarprime \in B, \E_4^c)\right| \le \Pr(\E_4^c) \le \delta,
    $$
    {thus} $\theta' \preceq \theta$.
\end{rproof}

\subsubsection{Sample-Path Order with Width \(\Theta(U/r)\)}
\label{appendix_case_study_dual_index_sample_path_order}

In this section, we show that the dual-index policy class also admits a
sample-path information order when \(L_r-L_e=1\), with width \(\Theta(U/r)\). The corresponding $T^{2/3}$ scaling regret bound follows by applying the proof of~\Cref{cor:dual_index_regret} with this information order and optimizing the discretization radius \(r\) to balance the statistical and discretization errors, as discussed in~\Cref{discussion_complexity}. This bound has a worse dependence on $T$ than the $\sqrt{T}$-scaling regret bound established in~\Cref{cor:dual_index_regret} using the distributional information order.

Concretely, the sample-path information order is characterized as follows.
\begin{lemma}
Suppose that \(L_r=L_e+1\). For any \(\theta=(z_e,z_r)\) and \(\theta'=(z'_e,z'_r)\), if \(z_r-z_e=z'_r-z'_e\) and \(z'_r\le z_r\), then \(\pi_{\theta'}\preceq\pi_\theta\) in the sample-path information order.
Consequently, for \(\Theta=\{(z_e,z_r)\in[0,U]^2:z_e\le z_r\}\), the width of the information order on an \(r\)-discretization of \(\Theta\) is \(\Theta(U/r)\).
\end{lemma}

\begin{rproof}
The proof follows the same argument as that of~\Cref{ic_lemma1}, i.e., fix any realized demand trajectory and couple the two systems along this sample path. As in~\Cref{ic_lemma1}, it suffices to show that, at every \(t\in [T]\), the on-hand inventory and the inventory at every position of both pipelines under \(\pi_{\theta'}\) are no larger than their counterparts under \(\pi_\theta\).
We prove this claim by induction on \(t\). The claim is immediate at \(t=1\), since both systems start from the same empty state. Suppose that it holds at time \(t\). Then
\[
    I_{\theta,t}+Q^r_{\theta,t-L_r}+Q^e_{\theta,t-L_e} 
    \ge
    I_{\theta',t}+Q^r_{\theta',t-L_r}+Q^e_{\theta',t-L_e},
\]
where, for each \(\bar\theta\in\{\theta,\theta'\}\), \(I_{\bar\theta,t}\), \(Q^r_{\bar\theta,t-L_r}\), and \(Q^e_{\bar\theta,t-L_e}\) denote, respectively, the on-hand inventory, the regular order arriving at time \(t\), and the expedited order arriving at time \(t\) under policy \(\pi_{\bar\theta}\), defined analogously to the corresponding quantities in~\Cref{case_study:dual_index_policy}. Under the same demand realization \(D_t\), it follows immediately that
\begin{equation}
\label{eq_dual_index_sample_path_order_sales_relation}
    N_{\theta',t}\le N_{\theta,t},
\end{equation}
where \(N_{\theta,t}\) and \(N_{\theta',t}\) denote the sales at time \(t\) under policies \(\pi_\theta\) and \(\pi_{\theta'}\), respectively. As in~\Cref{ic_lemma1}, it remains to show that
\begin{equation}
\label{eq_dual_index_sample_path_order}
    Q^r_{\theta,t+1}\ge Q^r_{\theta',t+1},
    \qquad
    Q^e_{\theta,t+1}\ge Q^e_{\theta',t+1}.
\end{equation}

Since \(L_r=L_e+1\), the total inventory position immediately before ordering at time \(t+1\) is \(z_r-N_{\theta,t}\) under \(\pi_\theta\). Recalling that the expedited order-up-to level is \(z_e\), we obtain
\(
    Q^e_{\theta,t+1} = \bigl(z_e-(z_r-N_{\theta,t})\bigr)^+ = \bigl(N_{\theta,t}-(z_r-z_e)\bigr)^+.
\)
Moreover,
\(
    Q^r_{\theta,t+1} = N_{\theta,t}-Q^e_{\theta,t+1} = N_{\theta,t}-\bigl(N_{\theta,t}-(z_r-z_e)\bigr)^+ = \min\{z_r-z_e,N_{\theta,t}\}.
\)
The same identities hold under \(\pi_{\theta'}\). Since both \(\bigl(x-(z_r-z_e)\bigr)^+\) and \(\min\{z_r-z_e,x\}\) are non-decreasing in \(x\), and \(z_r-z_e=z'_r-z'_e\), \cref{eq_dual_index_sample_path_order} follows directly from \cref{eq_dual_index_sample_path_order_sales_relation}. Therefore, the
induction holds, and the sample-path information order follows as in the proof of~\Cref{ic_lemma1}.
\end{rproof}

\subsection{M/M/1/L Queuing Model with Service Rate Control (\cref{case_study:queuing})}
\label{appendix_case_study_queueing}

\subsubsection{Information Order and Counterfactual Estimation}
\label{queuing_policy_order_definition}
{In this case, we modify the information‐order definition from \Cref{definition_general_order}, as we do not use empirical estimators (sample average); instead, we directly estimate the long‐term average cost \(g_{\theta'}\) for each \(\theta'\).
} Thus for any two policies $\theta$ and $\theta'$, we write $\pi_{\theta'} \preceq \pi_{\theta}$ if one can construct $\CGbarprime(\H_{\theta}^T)$ such that with probability at least $1 - \delta$,
\[
|\CGbarprime(\H_{\theta}^T) - g_{\theta'}| \leq O(H \sqrt{1 / \alpha T}).
\]
Using this definition of the information order, one can simply replace \cref{main_lem1} in \cref{thm:regret_bound} with this result to obtain the same regret bound. In this system, we exploit the fact that the performance of any policy is exactly characterized by $\lambda$ and $\mu$.  Hence, we use a ``plug-in'' approach to calculate our estimators $\CGbarprime(\H_{\theta}^T)$.  This contrasts with the empirical average cost from a trajectory sampled under a policy, as was our default estimator to establish an information order in \Cref{definition_general_order}.  In other words, this case study exemplifies that the information order in \Cref{definition_general_order} does not explicitly rely on the empirical average cost, and extends to any well-defined estimator.

\QueuingEstimator*
\begin{rproof}
We first note that, up to time horizon $T$, the numbers of times the system departs from $s=0$ and $s=L$ are standard counting processes. Concentration inequalities for such processes are well established in the literature~\citep{hsu2015mixing}. 
In particular, there exists an absolute constant $c_1>0$ such that, for all sufficiently large $T$ (see Lemmas~1 and~3 of~\citet{hsu2015mixing} for the corresponding requirements on $T$, after plugging in the relevant values of $\lambda$, $\mu$, and $U$), and for any $\delta>0$, with probability at least $1-2\delta$,
\begin{align}
    \sum_{i=0}^{N_\theta(T) -1 }\Ind{s_{\tau_{\theta, i}}=0} &\ge c_1 m_{\theta}(0)\lambda_0\frac{T}{U}, \label{eq_counting_1}\\
    \sum_{i=0}^{N_\theta(T) -1 }\Ind{s_{\tau_{\theta, i}}=L} &\ge c_1 m_{\theta}(L)(L\mu + \theta_L)\frac{T}{U}.
    \label{eq_counting_2}
\end{align}
Here the factor \(2\delta\) follows from a union bound over ~\cref{eq_counting_1} and~\cref{eq_counting_2}.
Intuitively, over a horizon $T$, the chain spends approximately $m_\theta(0)T$ and $m_\theta(L)T$ steps in states $0$ and $L$, with departure probabilities $\lambda_0/U$ and $(L\mu+\theta_L)/U$, respectively.
Thus the expected numbers of observed boundary departures are proportional to \(m_\theta(0)\lambda_0T/U\) and \(m_\theta(L)(L\mu+\theta_L)T/U\).
We refer to \citet{hsu2015mixing,cui2022tutorial,dzhaparidze2001bernstein,lezaud2001chernoff} for concentration inequalities for Markov chain occupation and counting processes. 
Note that the condition on $T$ in Lemma~1 of~\citet{hsu2015mixing} depends on the corresponding spectral gap. For completeness and future reference, we provide below a characterization of this gap using the birth--death spectral-gap bound of~\citet{chen2013mixing}.

\begin{lemma}[Restatement of Theorem~1.2 of \citet{chen2013mixing}]
\label{lemma_spectral_gap_0}
Consider an irreducible discrete-time birth and death chain on ${0,1,\ldots,L}$ with birth and death rates, $\lambda_s,\mu_s$. Let $\pi$ and $\Lambda$ be the stationary distribution and spectral gap and set
\[
\ell=\max\left\{\max_{j:j<i_0}\sum_{s=j}^{i_0-1}
\frac{\pi([0,j])}{\pi(s)\lambda_s},\max_{j:j>i_0}\sum_{s=i_0+1}^{j} \frac{\pi([j,L])}{\pi(s)\mu_s} \right\},
\]
where $\pi([a,b])$ refers to the stationary measure of $[a,b]$, and $i_0$ is a state such that $\pi([0,i_0])\ge 1/2$ and $\pi([i_0,L])\ge 1/2$. Then,
\[
\frac{1}{4\ell}\le \Lambda \le \frac{2}{\ell}.
\]
\end{lemma}
We apply~\Cref{lemma_spectral_gap_0} to the uniformized chain, whose birth and death probabilities are $\lambda_s/U$ and $\mu_s/U$, respectively.
\Cref{lemma_spectral_gap_0} immediately implies the following lemma.
\begin{lemma}
\label{lemma_spectral_gap_1}
    The spectral gap of the M/M/1/L model in~\Cref{case_study:queuing} is bounded below by $\hat{\Lambda}/U$, where $\hat{\Lambda}>0$ is independent of $L$ and $\theta\in\Theta$.
\end{lemma}
Note that~\Cref{lemma_spectral_gap_1} implies that $\hat{\Lambda}$ is a lower bound on the spectral gap of the corresponding continuous-time chain, uniformly over $L$ and $\theta$.
\begin{rproof}
It suffices to show that, for any $L$ and $\theta$, the quantity $\ell$ in~\Cref{lemma_spectral_gap_0} (denoted by $\ell_{L,\theta}$ for clarity) is bounded above by a constant multiple of $U$, where the constant is independent of $L$ and $\theta$.
This yields the desired uniform lower bound \(\Lambda\) on the spectral gap.

Under policy \(\theta\), the stationary distribution \(m_\theta\) of the M/M/1/L system satisfies the detailed balance equations
\[
m_\theta(s)\mu_s=m_\theta(s-1)\lambda_{s-1}, \qquad s=1,\ldots,L.
\]
In our model, $\lambda_s=\lambda(1 - \frac{s}{L})$ and $\mu_s=s\mu+\theta_s$. Let $\rho=\lambda/\mu$. Therefore, for all $s \ge 1$,
\begin{align}
m_\theta(s) = m_\theta(0) \prod_{r=1}^{s}
\frac{\lambda_{r-1}}{\mu_r} \le m_\theta(0) \prod_{r=1}^{s} \frac{\lambda}{r\mu+\theta_r} \le \prod_{r=1}^{s} \frac{\lambda}{r\mu} = \frac{\rho^s}{s!},
\label{eq_spectral_gap_2}
\end{align}
where the last inequality is from that $\theta_r\ge 0, \forall r \in \S$ and $m_\theta(0) \le 1$, for any policy $\theta \in \Theta$.
Following the definition of $i_0$ in~\Cref{lemma_spectral_gap_0}, we now choose an integer $M=M(\rho)$ such that
\(
\sum_{k=M+1}^{\infty} \frac{\rho^k}{k!} \le \frac{1}{2}.
\)
Then for any $L$, 
\(
m_\theta([M+1,L]) \le \sum_{k=M+1}^{\infty} \frac{\rho^k}{k!} \le \frac{1}{2}.
\)
Hence $m_\theta([0,M])\ge 1/2$. Now we let
\[
i_0=\min\{i:m_\theta([0,i])\ge 1/2\}.
\]
Then $i_0\le M$. Thus this $i_0$ is valid to apply Lemma~\ref{lemma_spectral_gap_0}.

Next, define
\(
R=\max\{M+1,\lceil 2\rho\rceil\}.
\)
We claim that $m_\theta(s)$ is uniformly bounded below for all $0\le s\le R$. Indeed, from~\cref{eq_spectral_gap_2},
\begin{align}
\frac{1}{m_\theta(0)} = 1+ \sum_{s=1}^{L} \prod_{r=1}^{s} \frac{\lambda_{r-1}}{r\mu+\theta_r} \le \sum_{s=0}^{\infty} \frac{\rho^s}{s!} = e^\rho.
\label{eq_queuing_proof_4}
\end{align}
Thus $m_\theta(0)\ge e^{-\rho}$. 
Without loss of generality, we assume $L \ge 2R$; the complementary case \(L<2R\) is absorbed into the constant.
Note that $\theta_s\le A_{\max}, \forall s \in \S$. For every $0\le s\le R$, we have that
\begin{align}
m_\theta(s) \ge e^{-\rho} \prod_{r=1}^{s} \frac{(1 - \frac{R-1}{L})\lambda}{r\mu+A_{\max}} \ge e^{-\rho} \prod_{r=1}^{s} \frac{(1 - \frac{R-1}{2R})\lambda}{r\mu+A_{\max}} = e^{-\rho} \prod_{r=1}^{s} \frac{\lambda}{2(r\mu+A_{\max})}.
\label{eq_queuing_proof_3}
\end{align}
Let
\begin{align}
c_R = e^{-\rho} \min\left\{1, \min_{1\le s\le R} \prod_{r=1}^{s} \frac{\lambda}{2(r\mu+A_{\max})}\right\}.
\label{eq_spectral_gap_4}
\end{align}
Then since $R$ depends on $\rho$ and $M$, and $M$ only depends on $\rho$, $c_R>0$ depends only on $\lambda,\mu,A_{\max}$, and correspondingly,
\(
\min_{0\le s\le R}m_\theta(s)\ge c_R.
\)

We first bound the left term in $\ell_{L, \theta}$ (see~\Cref{lemma_spectral_gap_0}). Since $i_0\le M\le R$ and $m_\theta([0,j])\le 1$ for all $j \in \S$,
\begin{align}
\label{eq_spectral_gap_6}
\max_{j:j<i_0} \sum_{s=j}^{i_0-1} \frac{m_\theta([0,j])}{m_\theta(s)\lambda_s} \le \sum_{s=0}^{M-1} \frac{2U}{m_\theta(s)\lambda} \le \frac{2M U}{\lambda c_R}.
\end{align}

We now bound the right term in $\ell_{L, \theta}$ in~\Cref{lemma_spectral_gap_0}. For every $s\ge R$, since $R \ge \lceil 2\rho\rceil$,
\begin{align}
\frac{m_\theta(s+1)}{m_\theta(s)} = \frac{\lambda_s}{\mu_{s+1}} \le \frac{\lambda}{(s+1)\mu+\theta_{s+1}} \le \frac{\lambda}{(s+1)\mu} = \frac{\rho}{s+1} \le \frac{1}{2}.
\label{eq_spectral_gap_9}
\end{align}
Therefore, for every $j'\ge R$,
\begin{align}
m_\theta([j',L]) = \sum_{s=j'}^{L}m_\theta(s) \le 2m_\theta(j').
\label{eq_spectral_gap_5}
\end{align}
We first consider the case where the maximum of the right term in~\Cref{lemma_spectral_gap_0} (i.e., $\max_{j:j>i_0}\sum_{k=i_0+1}^{j} \frac{\pi([j,L])}{\pi(k)\mu_k}$) is attained for some $j \ge R$. In this case, we have
\begin{align}
\sum_{s=i_0+1}^{j} \frac{m_\theta([j,L])}{m_\theta(s)\mu_s} &\le
\frac{2U}{\mu} \sum_{s=1}^{j} \frac{m_\theta(j)}{s m_\theta(s)} \nonumber \\
&= \frac{2U}{\mu} \sum_{s=1}^{R} \frac{m_\theta(j)}{s m_\theta(s)} + \frac{2U}{\mu} \sum_{s=R+1}^{j} \frac{m_\theta(j)}{s m_\theta(s)} \nonumber\\
&\le \frac{2U}{\mu} \frac{R}{c_R} + \frac{2U}{\mu},
\label{eq_spectral_gap_7}
\end{align}
where the first inequality is from~\cref{eq_spectral_gap_5}, and the second inequality is from~\cref{eq_spectral_gap_4} and~\cref{eq_spectral_gap_9}.
We then consider the complementary case where the maximum is attained for some $j<R$. Thus,
\begin{align}
\sum_{k=i_0+1}^{j} \frac{m_\theta([j,L])}{m_\theta(k)\mu_k} \le
\sum_{k=1}^{R} \frac{U}{m_\theta(k)k\mu} \le \frac{RU}{\mu c_R}.
\label{eq_spectral_gap_8}
\end{align}

Combining all above gives
\begin{align}
\ell_{L, \theta} \le \max\left\{ \frac{2MU}{\lambda c_R}, \frac{2U}{\mu}\left(\frac{R}{c_R}+1\right), \frac{RU}{\mu c_R} \right\},
\label{eq_queuing_proof_8}
\end{align}
where the first term is from~\cref{eq_spectral_gap_6}, the second term is from~\cref{eq_spectral_gap_7}, and the third term is from~\cref{eq_spectral_gap_8}.
Thus, by letting $\hat{\Lambda} = \frac{1}{4}\left(\max\left\{ \frac{2M}{\lambda c_R}, \frac{2}{\mu}\left(\frac{R}{c_R}+1\right), \frac{R}{\mu c_R} \right\}\right)^{-1}$, we have that $\hat{\Lambda}$ depends only on $\lambda,\mu,A_{\max}$. From~\Cref{lemma_spectral_gap_0}, $\Lambda \ge \hat{\Lambda}/U$ for all $L$ and $\theta \in \Theta$, which completes the proof.
\end{rproof}

We next show the concentration of the estimators \(\hat{\lambda}\) and \(\hat{\mu}\).
For a geometric random variable $H_i$ with success probability \(q\), the corresponding holding interval has mean \(q^{-1}\) and $\psi_1$ norm of order \(q^{-1}\)~\citep{vershynin2018high}. Therefore, for \(n\) such intervals, Bernstein inequality (see Theorem 2.8.1 of \citep{vershynin2018high}) gives
\[
\Pr\left(\left| \frac1n\sum_{i=1}^n H_i-q^{-1} \right| \ge \epsilon \right) \le 2\exp\left( -c_0 n\min\{q^2\epsilon^2,q\epsilon\}\right),
\]
for some constant $c_0$. In our setting, the effective numbers of intervals used to estimate \(\lambda\) and \(\mu\) are lower bounded by \cref{eq_counting_1} and \cref{eq_counting_2}. Note that the departure probability at $s = 0$ is $\lambda_0/U = \lambda/U$ and the departure probability at $s = L$ is $\mu_L/U = (L\mu + \theta_L)/U$. Thus, there exist constants
\(c_2,c_3>0\) such that, letting
\(
    \epsilon=\sqrt{\frac{1}{T}\log\left(\frac{1}{\delta}\right)},
\)
for sufficiently large $T$, with probability at least \(1-4\delta\), we have \begin{align}
    |\hat\lambda^{-1}-\lambda^{-1}|
    \le c_2\epsilon, 
    \qquad
    |(L\hat\mu+\theta_L)^{-1}-(L\mu+\theta_L)^{-1}| \le c_3\epsilon,
    \label{eq_queuing_proof_1}
\end{align}
where
\begin{align}
    c_2 = O\left(\sqrt{\frac{U}{c_1m_\theta(0)\lambda^3}}\right) = O\left( \sqrt{\frac{U\exp(\lambda/\mu)}{c_1\lambda^3}} \right),
    \label{eq_queuing_proof_5}
\end{align}
and
\begin{align}
    c_3 = O\left(\sqrt{\frac{U}{c_1m_\theta(L)(L\mu+\theta_L)^3}}\right) = O\left(
    \sqrt{\frac{U}{c_1m_\theta(L)(L\mu)^3}}\right).
    \label{eq_queuing_proof_6}
\end{align}
Here the \(O(\cdot)\) notation hides only absolute constants. The second equality in~\cref{eq_queuing_proof_5} follows from \cref{eq_queuing_proof_4}.
Thus, since $\lambda$ is bounded by $\lambda_{\max}$, \cref{eq_queuing_proof_1} implies that, for sufficiently large $T$, $\hat{\lambda}\le 2\lambda_{\max}$ and $(L\hat\mu+\theta_L) \le 2(L\mu+\theta_L)$. 
If we define
\begin{align*}
    \E_q = \left\{ |\hat\lambda-\lambda| \le 2c_2\lambda_{\max}^2\epsilon, |\hat{\mu} - \mu| \le 2c_3\frac{(\mu L + \theta_L)^2}{L}\epsilon\right\},
\end{align*}
then $\E_q$ occurs with probability at least $1 - 4\delta$.

For the remainder of the proof we condition on the event $\E_q$.
For any policy $\theta' \in \Theta$, based on $\hat{\lambda}$ and $\hat{\mu}$, we can derive a unique stationary distribution $\hat{m}_{\theta'}(s)$~\citep{puterman2014markov}.
In the true underlying continuous time Markov chain, denote by $Q(\lambda, \mu, \theta')$ as the $Q$ generator matrix under rates $\lambda$ and $\mu$ following policy $\pi_{\theta'}$.  Then we see that $\Delta Q=Q(\hat\lambda,\hat\mu, \theta')-Q(\lambda,\mu, \theta')$  has at most three nonzero entries per row. Using this we have:
\begin{align*}
\|\Delta Q\|_{\infty}
&=\max_i\sum_j|\Delta Q_{ij}|
\le2|\hat\lambda-\lambda|+2\max_s |s\hat{\mu} + {\theta'}_s - s\mu - {\theta'}_s | = O(c_2\lambda_{\max}^2\epsilon + c_3(\mu L + \theta_L)^2\epsilon),
\end{align*}
where $O(\cdot)$ only contains absolute constants.
We next bound $d_{\mathrm{TV}}\bigl(m_{\theta'},\hat m_{\theta'}\bigr)$.
Let $P_{\theta', u}(t)$ denote the state distribution at time $t$ under initial state distribution $u$ and policy $\theta'$. 
Then, by Theorem 2.1 of \citet{mitrophanov2003stability}, if one can show that for any initial state distributions $u_1$ and $u_2$, for all $t > 0$,
\begin{equation}
     d_{\mathrm{TV}}(P_{\theta', u_1}(t), P_{\theta', u_2}(t)) \le ce^{-bt},
     \label{eq_queuing_proof_2}
\end{equation}
then 
\begin{equation}
\label{eq_A.2}
    d_{\mathrm{TV}} \bigl(m_{\theta'},\hat m_{\theta'}\bigr) 
    \le \frac{\log(\frac{c}{2}) + 1}{b}\|\Delta Q\|_{\infty} 
    = O\left(\left(\frac{\log(\frac{c}{2}) + 1}{b}\right)\left(c_2\lambda_{\max}^2\epsilon + c_3(\mu L + \theta_L)^2\epsilon\right)\right).
\end{equation}

\cref{eq_queuing_proof_2} is a direct consequence of Eq.(20.21) of~\citet{levin2026markov}, with $1/c \ge O(\min_{s \in \S} m_{\theta'}(s))$ and $b$ being the (continuous chain) spectral gap under $\pi_{\theta'}$. We omit the details here for simplicity. From~\Cref{lemma_spectral_gap_1}, it is valid to substitute $b$ by $\hat{\Lambda}$.
We now bound the term $c \le O(\frac{1}{\min_{s \in \S} m_{\theta'}(s)})$. Note that from~\cref{eq_queuing_proof_4}, for any $s \in \S$, for any policy $\theta' \in \Theta$,
$$
m_{\theta'}(s) \ge e^{-\lambda/\mu} \prod_{r=1}^{s} \frac{(1 - \frac{s-1}{L})\lambda}{r\mu+A_{\max}} \ge e^{-\lambda/\mu} \left(\frac{\lambda}{L(L\mu+A_{\max})}\right)^s.
$$
Thus 
$$
    \log(\frac{c}{2}) \le O\left(\log(\frac{e^{\lambda/\mu}}{2}) + L\log\left(\max\left\{1, \frac{L(L\mu + A_{\max})}{\lambda}\right\}\right)\right) = O(L\log(L)).
$$
Thus we have 
$$
 d_{\mathrm{TV}}\bigl(m_{\theta'},\hat m_{\theta'}\bigr) \le O\left(\frac{1}{\hat{\Lambda}}L\log(L)\left(c_2\lambda_{\max}^2\epsilon + c_3(\mu L + \theta_L)^2\epsilon\right)\right).
$$
As a consequence, by the definition of $\CGbarprime(\H_{\theta}^T)$:
\begin{align}
    &|\CGbarprime(\mathcal{H}_{\theta}^T) - g_{\theta'}|  =  |\CGbarprime(\mathcal{H}_{\theta}^T) - \sum_{s=0}^L \hat{m}_{\theta'}(s) \Exp{C(s, \pi_{\theta'}(s))} + \sum_{s=0}^L \hat{m}_{\theta'}(s) \Exp{C(s, \pi_{\theta'}(s))} - g_{\theta'}| \nonumber \\
    & \qquad \leq \abs{\CGbarprime(\mathcal{H}_{\theta}^T) - \sum_{s=0}^L \hat{m}_{\theta'}(s) \Exp{C(s, \pi_{\theta'}(s))}} + \abs{\sum_{s=0}^L \hat{m}_{\theta'}(s) \Exp{C(s, \pi_{\theta'}(s))} - g_{\theta'}} \nonumber \\
    &\qquad = \abs{\sum_{s=0}^L \hat{m}_{\theta'}(s) \Exp{\hat{C}(s, \pi_{\theta'}(s))} - \sum_{s=0}^L \hat{m}_{\theta'}(s) \Exp{C(s, \pi_{\theta'}(s))}} \nonumber \\
    &\qquad + \abs{\sum_{s=0}^L \hat{m}_{\theta'}(s) \Exp{C(s, \pi_{\theta'}(s))} - \sum_{s=0}^L m_{\theta'}(s) \Exp{C(s, \pi_{\theta'}(s))}} \nonumber \\
    &\qquad \leq \sum_{s=0}^L \hat{m}_{\theta'}(s) \left|\Exp{\hat{C}(s, \pi_{\theta'}(s))} - \Exp{C(s, \pi_{\theta'}(s))}\right| + \left(\sup_{\theta'' \in \Theta} \sup_{s \in \S} \Exp{C(s, \pi_{\theta''}(s))}\right) \sum_{s=0}^L \abs{\hat{m}_{\theta'}(s) - m_{\theta'}(s)} \nonumber \\
    &\qquad \leq 2C \frac{L}{U} \abs{\mu - \hat{\mu}}+ 2\frac{C_{\max}}{U} d_{\mathrm{TV}}(m_{\theta'}, \hat{m}_{\theta'})
    \leq O\left(\frac{C_{\max}}{U}\frac{1}{\hat{\Lambda}}L\log(L)\left(c_2\lambda_{\max}^2\epsilon + c_3(\mu L + \theta_L)^2\epsilon\right)\right) \nonumber \\
    &\qquad = O\left(\frac{w(A_{\max}) + CL\mu_{\max}}{U}\frac{1}{\hat{\Lambda}}L\log(L)\left(c_2\lambda_{\max}^2\epsilon + c_3(\mu L + A_{\max})^2\epsilon\right)\right).
    \label{eq_queuing_proof_7}
\end{align}
The second inequality is due to Holder's inequality, and $C_{\max}/U = (w(A_{\max}) + CL\mu_{\max})/U$ is an upper bound on the expected cost per timestep. Since $U = \Theta(L)$, we have that under event $\E_q$,
$$
    |\CGbarprime(\mathcal{H}_{\theta}^T) - g_{\theta'}| \le \tilde{O}\left( L^2\sqrt{\frac{1}{m_{\theta}(L)}} \sqrt{\frac{\log(1 / \delta)}{T}}\right).
$$
\end{rproof}

\subsection{Inventory Control with Returns~\citep{chen2025managing}}
\label{case_study_return}

In this subsection we show that our information-order framework,~\Cref{alg_policy_elimination}, and~\Cref{thm:regret_bound} apply in the inventory model with returns introduced in~\citet{chen2025managing}.

\subsubsection{Model}

The model largely follows the inventory model studied in~\Cref{case_study:inventory} with zero lead time, except that sold units may be returned within a return window of length \(N\). Concretely, at each time step \(t \in [T]\), the system evolves as follows. First, the manager observes the on-hand inventory \(I_t\), as well as the outstanding sales \(Q_{t, j}\) for \(j \in [N]\), where \(Q_{t, j}\) denotes the number of units sold at time \(t-j\) that have not yet been returned by time \(t\). Units that were sold at time \(t-j\) but have already been returned are therefore not included in \(Q_{t, j}\).
The manager chooses an ordering quantity \(O_t\), and the order arrives immediately. After the order arrives, returns are realized. For each \(j \in [N]\), each outstanding unit sold \(j\) periods ago is returned at time \(t\) with probability \(\lambda^{(j)}\). We assume that returns are independent across units and across \(j \in [N]\), and that \(\{\lambda^{(j)}:j \in [N]\}\) are known~\citep{chen2025managing}. 
Then, the discrete demand \(D_t \in \{0,1,\ldots,\bar{D}\}\) is realized from an unknown demand distribution, where \(\{D_t:t\in[T]\}\) are i.i.d. We assume that $\Ex[D_t] \ge \underline{d} > 0$. Demand can be satisfied using the on-hand inventory \(I_t\), the newly arrived order \(O_t\), and the units returned at time \(t\). Any unsatisfied demand is lost. As in~\Cref{case_study:inventory}, demand is censored; the manager observes only the sales quantity when inventory is insufficient. After these steps, the system transitions to time \(t+1\).

\myparagraph{MDP Formulation.}  
To formulate this problem as an MDP, define the state space as
\(
    \S = \{0,\ldots,N\bar{D} + \bar{D}\}\times\{0,\ldots,\bar{D}\}^N,
\)
where the first component \(\{0,\ldots,N\bar{D}+\bar{D}\}\) represents the on-hand inventory, and the second component \(\{0,\ldots,\bar{D}\}^N\) represents the outstanding sales. The upper bound \(N\bar{D}+\bar{D}\) on the on-hand inventory follows from the fact that at most \(N\bar{D}\) units can be outstanding and later returned, while the maximum demand in each time step is bounded by \(\bar{D}\).
The action space is
\(
    \A = \{0,\ldots,\bar{D}\}.
\)
At time \(t\), for each outstanding-sales component \(Q_{t,j}\), let \(R_{t,j}\) denote the number of returned units. Note that
\(
    R_{t,j}\le Q_{t,j}
\)
a.s. The transition dynamics and one-step cost are given by
\begin{align}
    S_{t+1}
    &= (I_{t+1}, Q_{t+1,N}, \ldots, Q_{t+1,1}) \nonumber\\
    &= \left(\left(I_t + O_t + \sum_{j=1}^N R_{t,j} - D_t\right)^+, Q_{t,N-1} - R_{t,N-1}, \ldots, \min\left\{I_t + O_t + \sum_{j=1}^N R_{t,j}, D_t\right\} \right), \nonumber\\
    C_t(S_t)
    &= h\left(I_t + O_t + \sum_{j=1}^N R_{t,j} - D_t\right)^+ -
    p\min\left\{I_t + O_t + \sum_{j=1}^N R_{t,j}, D_t\right\}, \label{eq_return_pseduo_cost}
\end{align}
where \(h\) and \(p\) are the holding-cost and shortage-penalty parameters, respectively. Note that~\cref{eq_return_pseduo_cost} is already after the pseudo-cost transformation, which guarantees cost observability, analogous to the pseudo-cost transformation in~\Cref{case_study:inventory}; see~\cref{eq_pseudo_cost}. For more details on the pseudo-cost transformation in this model, we refer the reader to Section~3.3 of~\citet{chen2025managing}. 
We assume a fixed starting state $(0, \ldots, 0)$.

\subsubsection{Policies and Information Order}
As in~\citet{chen2025managing}, we consider the order-up-to-level policy class. Concretely, let
\(
    \Theta = \{0,\ldots,\bar{D}\}.
\)
For each \(\theta \in \Theta\), given state
\(
    S_t = (I_t, Q_{t,N}, \ldots, Q_{t,1}),
\)
policy \(\pi_\theta\) induces the ordering quantity
\[
    \pi_{\theta}(S_t) = (\theta - I_t)^+.
\]
The asymptotic optimality of the best policy in \(\Theta\) is provided in Lemma~1 of~\citet{chen2025managing}.

We now argue that this policy class satisfies a sample-path information order.
\begin{lemma}
    \(\Theta\) satisfies the sample-path information order of width one, where \(\pi_{\theta'} \preceq \pi_\theta\) whenever \(\theta' \leq \theta\).
\end{lemma}

\begin{rproof}
For any \(\theta' \le \theta\), starting from the initial state \(s_1=(0,\ldots,0)\), the on-hand inventory under \(\pi_{\theta'}\) is always upper bounded by the on-hand inventory under \(\pi_\theta\), in the sample-path sense. Similarly, for any \(t \in [T]\) and any \(j \in [N]\), the outstanding-sales component \(Q_{t,j}\) under \(\pi_{\theta'}\) is upper bounded by its counterpart under \(\pi_\theta\). This follows directly from the order-up-to structure of \(\Theta\); the sample-path coupling details are standard and can be found in~\citet{chen2025managing}. The sample-path information order then follows as in the proof of~\Cref{ic_lemma1}.
\end{rproof}

\subsubsection{Performance Guarantee}
The regret bound obtained by applying the information-order framework to this model follows immediately. As in~\citet{chen2025managing}, define
\[
    \bar\lambda := \prod_{j=1}^{N}(1-\lambda^{(j)})
\]
as the probability that a sold unit is never returned during the return window.

\begin{corollary}
    \label{corollary_case_study_return}
    Applying~\Cref{alg_policy_elimination} to this inventory model with returns yields
    \(
        \Regret(T) = \tilde{O}\left(\max\{h,p\} \frac{N^3\bar{D}^3}{\underline{d}\bar{\lambda}^2}\sqrt{\log(\bar{D}T^{3/2})T}\right).
    \)
\end{corollary}

\begin{remark}
    The leading constant in~\Cref{corollary_case_study_return} is theoretically weaker than that in~\citet{chen2025managing}. This difference is because \citet{chen2025managing} uses a more problem-specific analysis.  When analyzing the concentration of their estimators, they first pay a transient mixing cost and then estimate the long-run cost using finite-length samples under the stationary distribution.  This allows them to avoid a uniform bias-span bound $H$ in the leading term; see Lemma~12 therein.
    In contrast, our information-order framework applies to more general settings, including those in which \(H\) is bounded but the mixing time need not be; for examples of such settings, see the related discussion in~\Cref{section_modelling_assumptions}.
\end{remark}

\begin{rproof}
With the sample-path order established above, to apply our information-order framework in this model, it suffices to show that \Cref{assumption_continuity,assumption_restart,assumption_MRP} are satisfied. Following the model above and similarly to~\Cref{case_study:inventory}, \Cref{assumption_continuity} and~\Cref{assumption_restart} are satisfied directly. The Lipschitzness is trivial since the model is discrete. Thus, it suffices to show the uniformity of \(g_\theta\) and the bounded span of bias \(H\).

\begin{lemma}
\label{lemma_bounded_span_return}
    For any \(\theta \in \Theta\), we have that \(g_\theta\) is uniform, and that
    \(
        H \le \frac{48\Cupperbound}{\bar{\lambda}^2}\frac{N^2\bar{D}^2}{\underline{d}}.
    \)
\end{lemma}

\begin{rproof}
For a fixed policy \(\theta\), for any \(s_1, s_2 \in S_\theta\), and for any given \(T\), define
\[
    \tau_{s_1,s_2,\theta}=\left(\inf\left\{t\in[T]:S^{\theta}_{t \mid s_1}=S^{\theta}_{t \mid s_2}\right\}\right)\wedge T,
\]
where the infimum over the empty set is understood as \(\infty\), and \(S^{\theta}_{t\mid s'}\) refers to the state at time \(t\) under policy \(\theta\) with initial state \(s'\). We construct the two chains \(S^\theta_{t\mid s_1}\) and \(S^\theta_{t\mid s_2}\) on the same probability space by using the same realized demand sequence and the same item-level return random variables. We refer the reader to~\citet{chen2025managing} for the details of this sample-path coupling.

Let
\[
    S^\theta_{t\mid s_i} = \left(I_t^{(i)},Q_{t,N}^{(i)},\ldots,Q_{t,1}^{(i)}\right)
\]
for \(i \in \{1,2\}\) (named as system $1$ and $2$ hereafter), where $I_t^{(i)}$ and $Q_{t, j}^{(i)}, j \in [N]$ refer to the on-hand inventory and the outstanding sales of time $t - j$ with initial state $s_i$.
Define the discrepancy process
\[
    \Delta_t = \left|I_t^{(1)}-I_t^{(2)}\right| + \sum_{j=1}^{N} \left|Q_{t, j}^{(1)}-Q_{t, j}^{(2)}\right|.
\]
The goal is to bound the time required for \(\Delta_t\) to reach \(0\). Once \(\Delta_t=0\), the two coupled chains are in the same state and therefore evolve identically thereafter; hence this hitting time is a sample-path upper bound on \(\tau_{s_1,s_2,\theta}\).
We first show that \(\Delta_t\) is non-increasing over \(t \in [T]\). In each time step \(t\), there are three operations: (i) order placement, (ii) realization of returns and demand, and (iii) transition to \(t + 1\). First, order placement does not increase \(\Delta_t\). This is obvious, since if \(I_t^{(1)} \le I_t^{(2)}\), the order-up-to-level policy described above implies that \(q_t^{(1)} \ge q_t^{(2)}\). Here \(q_t^{(i)}\), \(i \in \{1,2\}\) denotes the ordering quantity at time \(t\) with initial state \(s_i\).  As a consequence, after order placement,
\(
    \left|I_t^{(1)}-I_t^{(2)} + q_t^{(1)} - q_t^{(2)}\right|
\)
does not increase. Also, the realization of returns does not increase \(\Delta_t\), since for any \(j \in [N]\), the realization of returns cannot increase \(\left|Q_{t, j}^{(1)}-Q_{t, j}^{(2)}\right|\), and each unit decrease of \(\left|Q_{t, j}^{(1)}-Q_{t, j}^{(2)}\right|\) can incur at most one unit increase in the on-hand inventory discrepancy. 
The realization of demand does not increase the discrepancy.
The transition does not increase the discrepancy as well, which is immediate. Thus, for any \(t \in [T - 1]\),
\(
    \Delta_t \ge \Delta_{t+1}
\)
a.s.

Next, we bound the expected time to eliminate the initial discrepancy \(\Delta_1\). Since \(s_1\) and \(s_2\) are fixed, \(\Delta_1\) is deterministic. Recall that $\bar\lambda$ is the probability that a sold item is never returned during the return window. Define the stopping times
\[
    \rho_0=1,\qquad
    \rho_{i} =\inf\{t\ge \rho_{i-1}:\Delta_t\le \Delta_1-i\},
    \qquad i=1,\ldots,\Delta_1,
\]
and let
\(
    \tau_i:=\rho_{i}-\rho_{i-1}.
\)
We now provide a uniform upper bound on \(\mathbb E[\tau_i]\) over all possible \(i\). Consider any time \(\rho_{i-1}\) with \(\Delta_{\rho_{i-1}}>0\). Select one unit that contributes to the discrepancy. Note that such a unit either belongs to the outstanding sales or remains in on-hand inventory. Define \(t_i\) as the number of periods from \(\rho_{i-1}\) until this selected unit is no longer active in the system, i.e., until this unit is sold and cannot be returned anymore. We directly obtain that \(\tau_i \le t_i\) a.s. (for more details see the coupling in~\citet{chen2025managing}). Thus, to bound \(\mathbb E[\tau_i]\), it suffices to bound \(\Ex[t_i]\).

Let \(\E_{\Delta_i}\) denote the event that this unit is not returned after time step \(\rho_{i-1}\), and let \(\E^C_{\Delta_i}\) denote its complement. Let \(L_{\mathrm{on}}\) denote the stochastic time required to deplete \((N+1)\bar{D}\) units of on-hand inventory. Note that for any on-hand unit, the time required to deplete \((N+1)\bar{D}\) units of on-hand inventory is a sample-path upper bound on the time required to deplete that unit.
We also note that it is sufficient to ignore new orders in this depletion argument. Indeed, suppose the selected unit in system \(1\) creates a positive on-hand discrepancy, so that \(I_t^{(1)} > I_t^{(2)}\). If system \(1\) places a new order at time \(t\), then, under the order-up-to policy, system \(2\) must receive at least as many units through new orders and returns. In that case, the on-hand discrepancy caused by the selected unit is immediately offset. Therefore, new orders can only help eliminate the selected discrepancy, and ignoring them gives a valid sample-path upper bound on the depletion time.
Then we have
\begin{align*}
    \Ex[t_i] &= \Ex[t_i \mid \E_{\Delta_i}] \Pr(\E_{\Delta_i}) + \Ex[t_i \mid \E^C_{\Delta_i}] \Pr(\E^C_{\Delta_i}) \\
    &\le \Ex[t_i \mid \E_{\Delta_i}] + \Ex[t_i \mid \E^C_{\Delta_i}] (1 - \bar{\lambda}) \\
    &\le N + \Ex[L_{\text{on}}\mid \E_{\Delta_i}] + (1 - \bar{\lambda})[N + \Ex[L_{\text{on}}\mid \E^C_{\Delta_i}] + \Ex[t_i]] \\
    &\le N + \Ex[L_{\text{on}}] + (1 - \bar{\lambda})[N + \Ex[L_{\text{on}}]] + (1 - \bar{\lambda})\Ex[t_i],
\end{align*}
where the second line follows from the fact that, regardless of the position of the item, the probability that this item is returned is upper bounded by \(1 - \bar{\lambda}\), and the last line follows from the fact that \(L_{\text{on}}\) is independent of \(\E_{\Delta_i}\). Note that
\(
    \Ex[L_{\text{on}}] \le \frac{N\bar{D} + 2\bar{D}}{\bar{\lambda}\underline{d}},
\)
which follows from Wald's identity, since the on-hand inventory is upper bounded by \(N\bar{D}+\bar{D}\), the overshoot is upper bounded by $\bar{D}$, and the expected number of non-returned sales per time step is lower bounded by \(\bar{\lambda}\underline{d}\). With some arithmetic,
\[
    \Ex[t_i] \le \frac{2}{\bar{\lambda}}\left(N + \frac{N\bar{D} + 2\bar{D}}{\bar{\lambda}\underline{d}}\right) \le \frac{8}{\bar{\lambda}^2}\frac{N\bar{D}}{\underline{d}},
\]
where the last inequality follows from that \(\bar{D} \ge \underline{d}\) and \(N \ge 1\). Thus,
\begin{align}
    \Ex[\tau_{s_1,s_2,\theta}]
    \le \Ex\left[\sum_{i=1}^{\Delta_1}\tau_i\right]
    \le \Ex\left[\sum_{i=1}^{\Delta_1}t_i\right]
    \le (2N\bar{D} + \bar{D}) \frac{8}{\bar{\lambda}^2}\frac{N\bar{D}}{\underline{d}}
    \le \frac{24}{\bar{\lambda}^2}\frac{N^2\bar{D}^2}{\underline{d}}.\label{diameter_return}
\end{align}

With the bound on $\Ex[\tau_{s_1,s_2,\theta}]$, we now show that \(g_{\theta}\) is uniform and that \(H\) is bounded. Similar to~\Cref{lem:dual_index_bias},
\[
\left| g_\theta(s_1) - g_\theta(s_2) \right| = \left|\lim_{T \to \infty} \frac{1}{T}  \mathbb{E}\left[TG_{\theta}(\H_{\theta}^T)(s_1) - TG_{\theta}(\H_{\theta}^T)(s_2)\right] \right| \le \lim_{T \to \infty} 2\frac{\Cupperbound}{T} \Ex[\tau_{s_1,s_2,\theta}] = 0,
\]
and
\begin{align*}
    \left| v_\theta(s_1) - v_\theta(s_2) \right| 
    = \left| \lim_{T \to \infty} \Ex[ TG_{\theta}(\H_{\theta}^T)(s_1) - Tg_\theta - TG_{\theta}(\H_{\theta}^T)(s_2) + Tg_\theta ] \right|
    \le 2\Cupperbound \Ex[\tau_{s_1,s_2,\theta}]
    \le \frac{48\Cupperbound}{\bar{\lambda}^2}\frac{N^2\bar{D}^2}{\underline{d}}.
\end{align*}
\end{rproof}

The proof is completed by noting $U = \bar{D}, H \le \frac{48\Cupperbound}{\bar{\lambda}^2}\frac{N^2\bar{D}^2}{\underline{d}}, \alpha = w = d = 1, \Cupperbound = \max\{h,p\}(N\bar{D} + \bar{D})$. A bound of \(D_\Theta\) also follows from~\cref{diameter_return}, which provides a bound on the diameter.
\end{rproof}

\subsection{Inventory Model with Multiple Products}
\label{case_study_multiple_products}

A natural question is whether an information order must always relate two individual policies; that is, whether all information required to evaluate a policy \(\pi_{\theta'}\) must come from implementing a single policy \(\pi_{\theta}\). This need not be the case. For example, implementing one policy may reveal part of the information required to evaluate \(\pi_{\theta'}\), while implementing another policy may reveal the remainder. The combined information from these policies may therefore be sufficient for counterfactual evaluation. In this section, we illustrate this possibility through a simple extension of the inventory model in~\Cref{case_study:inventory}.

Consider an inventory model with \(d \ge 1\) products, each with a lead time $L$. At each time \(t\), the retailer chooses ordering quantities \(Q_{1,t},\ldots,Q_{d,t}\), and the dynamics of each product are identical to those in~\Cref{case_study:inventory}. For each product \(i\in[d]\), the retailer observes the inventory state \(I_{i,t},Q_{i,t-L},\ldots,Q_{i,t-1}\), where \(I_{i,t}\) is the on-hand inventory at the beginning of period \(t\), and \(Q_{i,t-L},\ldots,Q_{i,t-1}\) are the outstanding orders in the pipeline.
For each \(i\in[d]\), the demands \(\{D_{i,t}\}_{t\ge1}\) are i.i.d.\ over time with an unknown distribution \(\F_i\) supported on \([0,U]\). The demand processes are mutually independent across products (the results are readily extendable to dependent demands). As in~\Cref{case_study:inventory}, we assume that $\Pr(D_{i,t}=0)>0$ for all $i\in[d]$. Demand is censored, and unmet demand is lost.

The transition and cost dynamics for each product are identical to those in~\Cref{case_study:inventory}, and the total one-step cost is the sum of the costs across products. The corresponding pseudo-cost transformation is also identical and is therefore omitted.
We consider product-specific order-up-to policies. For
\(
\theta=(\theta_1,\ldots,\theta_d)\in\Theta=[0,U]^d,
\)
the ordering decision for product \(i\in[d]\) is
\begin{equation*}
\pi_{\theta_i} \bigl(\Inventory_{i,t}, \Order_{i,t-L}, \ldots, \Order_{i,t-1} \bigr) = \Order_{i,t} = (\theta_i - \Inventory_{i,t} - \tsum_{j\in[L]}\Order_{i,t-j})^+.
\end{equation*}
Here, with a slight abuse of notation, \(\pi_{\theta_i}\) denotes the \(i\)-th coordinate of the action selected by policy \(\pi_{\theta}\), namely the ordering quantity for product \(i\). Note that as in~\Cref{case_study:inventory}, the state space, action space, and policy class are compact.

Under the sample-path order introduced in~\Cref{section_policy_ordering}, the width of the discretized policy class scales as \(\Theta\left(\left(\frac{U}{r}\right)^{d-1}\right)\), since $\theta'\preceq\theta$ if and only if $\theta_i'\le\theta_i$ for all $i\in[d]$, which is the standard coordinatewise partial order~\citep{srinivasan2011symmetric}. Thus, the optimal discretization radius is \(r=\Theta\left(T^{-\frac{1}{d+1}}\right)\), leading to the $T^{\frac{d}{d+1}}$ scaling regret.

We now show that, following the intuition introduced at the beginning of the section, combining information from multiple implemented policies yields an information order with constant width $w \le d$, and hence a $\sqrt{T}$ scaling regret bound by~\Cref{thm:regret_bound}.
Fix \(\theta'=(\theta_1',\ldots,\theta_d')\in\Theta\).
By~\Cref{ic_lemma1}, for each product \(i\in[d]\), the counterfactual sample corresponding to order-up-to level \(\theta_i'\) can be exactly recovered from a trajectory generated by any policy \(\theta^{(i)}\in\Theta\) satisfying \(\theta_i'\le\theta_i^{(i)}\).
Therefore, for any active policy set \(\Theta_k\) in~\Cref{alg_policy_elimination}, it is sufficient to implement at most \(d\) distinct policies \(\theta^{(1)},\ldots,\theta^{(d)}\), where
\(
\theta^{(i)} \in \arg\max_{\theta\in\Theta_k}\theta_i
\)
for all $i \in [d]$.
For each product \(i\), the trajectory generated by \(\theta^{(i)}\) yields the corresponding counterfactual product-level trajectory under \(\theta_i'\). Because the demand processes are independent across products, these recovered product-level trajectories can be combined to construct a random variable with the same distribution as
\(
G_{\theta'}\bigl(\mathcal H_{\theta'}^T\bigr)
\)
for any $T$. This induces a distributional information order with \(\alpha=1\) and \(\Th(\delta)=1\) for every \(\delta\in(0,1)\). Although this relation slightly extends the information order defined in~\Cref{section_policy_ordering}, we continue to refer to it as an information order; the definition of width and the remainder of~\Cref{alg_policy_elimination} apply analogously.

\begin{lemma}
\label{lemma_multiple_products}
For the inventory model with \(d\) products, there exists an information order with width \(w\le d\).
\end{lemma}

\Cref{assumption_continuity,assumption_restart,assumption_MRP} follow directly from the corresponding results in~\Cref{case_study:inventory}. In particular, the bound on $H$ becomes $H = 36\max\{h,p\}LdU$, instead of \(36\max\{h,p\}LU\) in~\Cref{case_study:inventory}, because the total cost is the sum of the costs across \(d\) products. The constants \(L_{\Theta}\), \(D_{\Theta}\), and \(\Cupperbound\) change accordingly. The regret guarantee then follows directly from~\Cref{thm:regret_bound}.

\begin{corollary}
\label{cor_inventory_multiple_product_regret}
Applying~\Cref{alg_policy_elimination} to the inventory model with \(d\) products and lead time $L$ yields
\[
\Regret(T) = \tilde O\left(Ld^2\max\{h,p\}U\sqrt{\log(UT^{3/2})T}\right).
\]
\end{corollary}

 \section{Numerical Simulations}
\subsection{Simulation Information for~\Cref{sec:simulations}} \label{sec:simulation_information}
\begin{table}[t]
  \centering
  \small  
  \caption{Parameter specifications for the numerical simulations.}
  \label{tab:configs-by-block}
  \begin{tabular}{|l|c|c|}
    \hline
      & \textbf{Small-scale} & \textbf{Large-scale} \\ \hline

    %—— Block 1 — Inventory Control ——
    \multicolumn{3}{|c|}{\bfseries Inventory Control} \\ \hline
    Lead Time ($L$)             & 2   & 6     \\ \hline
    Holding Cost ($h$)          & 1   & 1     \\ \hline
    Lost-sales Penalty ($p$)      & 10   & 10     \\ \hline
    $\Pr(D_t = 0)$ ($\gamma$)      & 0.3   & 0.3     \\ \hline
    Demand Support ($d$)        & $[0,3]$   & $[0,40]$    \\ \hline
    \multirow{3}{*}{Distributions}
                                & Exp($\lambda=1$)   & Exp($\lambda=\frac{40}{3}$)    \\
                                & Normal($1.5,0.5$)      & Normal($20,\frac{20}{3}$)       \\
                                & Uniform($[0,3]$)& Uniform($[0,40]$) \\ \hline

    %—— Block 2 — Dual Sourcing ——
    \multicolumn{3}{|c|}{\bfseries Dual Sourcing} \\ \hline
    Lead Time ($L_r,L_e$)
                                & (1,0)  & (5,0)    \\ \hline
    Holding Cost ($h$)          & 1    & 1      \\ \hline
    Shortage Penalty ($p$)            & 10    & 10      \\ \hline
    Purchasing Cost ($c_r, c_e$)
                                & (0, 0.5)  & (0, 0.5)    \\ \hline
    $\Pr(D_t = 0)$ ($\gamma$)      & 0.3   & 0.3     \\ \hline
    Demand Support ($d$)        & $[0,3]$   & $[0,40]$    \\ \hline
    \multirow{3}{*}{Distributions}
                                & Exp($\lambda=1$)   & Exp($\lambda=\frac{40}{3}$)    \\
                                & Normal($1.5,0.5$)      & Normal($20,\frac{20}{3}$)       \\
                                & Uniform($[0,3]$)& Uniform($[0,40]$) \\ \hline

    %—— Block 3 — Queuing ——
    \multicolumn{3}{|c|}{\bfseries Queuing Model}  \\ \hline
    Buffer Size ($L$)        & -- & $2$    \\ \hline
    Maximum Service Rate ($A_{\max}$)        & -- & $3$    \\ \hline
    Arrival Rate ($\lambda, \lambda_{\max}$)        & -- & $(6, 10)$    \\ \hline
    Service Rate ($\mu, \mu_{\max}$)        & --  & $(3,10)$    \\ \hline
    Power Cost Function ($w(a)$)          & --    & $w(a) = a^2$      \\ \hline
    Deadline Missing Penalty ($C$)          & --    & 100      \\ \hline

  \end{tabular}
\end{table}

\paragraph{Computing Infrastructure.} The experiments were conducted on a server with an Intel i7-14700K 20 Core Processor and 32GB of RAM. No GPUs were needed for the experiments.  Each simulation (evaluating all algorithms included in the figures) took approximately 60 hours.

\paragraph{Experiment Setup.}\label{experiment_settings}
Each experiment was repeated \(20\) times, and all plots and metrics report averages over these runs.  Each policy was evaluated on trajectories of length \(10^5\).  Hyperparameters for all algorithms were set as follows: \ALG uses only the constants specified in the main paper (e.g.\ \(H\), \(\alpha\), \(r\)) with no additional tuning; \textsf{PPO}'s constant learning rate (chosen from $[10^{-5}, 10^{-4}, 10^{-3}, 10^{-2}]$), 
batch size ($[16, 32, 64, 128]$), and discount factor ($[0.9, 0.99, 0.999, 0.9999]$) were selected via grid search. 
\textsf{SGD}'s constant learning rate (ranging between $10^{-2}$ and $4 \times 10^{4}$), 
sample size (chosen from $[10^{0}, 10^{1}, 10^{2}, 10^{3}, 10^{4}]$), 
and the gap width (ranging between $10^{-2}$ and $10^{0}$) used in the finite-difference estimator 
were also tuned via grid search. 
\textsf{Feedback Graph}'s~\citep{dann2020reinforcement} exploration bonus scale was selected via grid search over $[10^{-4}, 10^{4}]$. 
The convexity-based algorithm of \citet{agrawal2019learning} and \textsf{BASA}~\citep{chen2024tailored} involve no tuning. 
\textsf{UCRL2}'s~\citep{anselmi2022reinforcement} exploration bonuses (fixed constants $7$ and $14$) 
are taken directly from the original paper, 
while its confidence parameter $\delta$ was tuned by grid search over $[0.01, 0.05, 0.1]$.

To identify the optimal policy $\theta^*$ in each setting, we discretized the policy class with a radius $r = 0.1$ (for the algorithms that need discretization, except \ALG, which is determined by $T^{-1/2}$). 

In the inventory and dual-sourcing simulations, we train the algorithms on pseudocost $\tilde{C}$ but report performance on $C$ (as discussed in \Cref{case_study:inventory} and \ref{case_study:dual_index_policy}).  Additionally, for the dual-sourcing simulations, we assume $c_r = 0$ and $L_e = 0$, i.e., the purchasing cost in the regular channel is omitted, and we assume the lead time of the expedited channel is $0$. This is to have a fair comparison with~\citet{chen2024tailored}.

In \cref{tab_simulation_all} we report the performance of the M/M/1/L queueing model under exponential distributions, and leave the other two blocks empty. All other parameter specifications for the numerical simulations are in~\cref{tab:configs-by-block}.

% We also observe that in both the decaying‐rate and fixed‐rate scenarios, the {\textsf Problem‐Specific} algorithm (i.e., {\textsf UCRL2}) underperforms even the {\textsf Random} policy.  This is because the state and action spaces are relatively small (\(L=2, A_{\max}=3\)), where random actions are not far away from reasonable solutions.  Consequently, when a specialized algorithm fails to exploit structure effectively (by $T = 3\times10^5$), its performance may be worse than the \textsf{Random} policy.  We limit our experiments to this scale because {\textsf UCRL2} requires comparing multiple MDP models at each epoch, which becomes prohibitively expensive in larger scale problems.

\subsection{Additional Empirical Simulations}

\subsubsection{Numerical Assessment of the Theoretical Conditions in the Dual-Index Case Study}
\label{appendix_case_study_dual_index_empirical_evidence}

The theoretical guarantees for the dual-index case study rely on several conservative sufficient conditions, including the $\gamma^{-L_r}$-type dependence in the bounds on $H$ and $\alpha$, the Lipschitz assumption over the policy class, and a potentially long burn-in period before the regret guarantee becomes informative. In this section, we investigate whether these features reflect the practical behavior of the model or are primarily artifacts of the analysis.

\begin{figure}[!t]
    \centering
    \captionsetup[subfigure]{font=footnotesize}
    
    \begin{subfigure}[t]{0.32\textwidth}
        \centering
        \includegraphics[width=\linewidth]{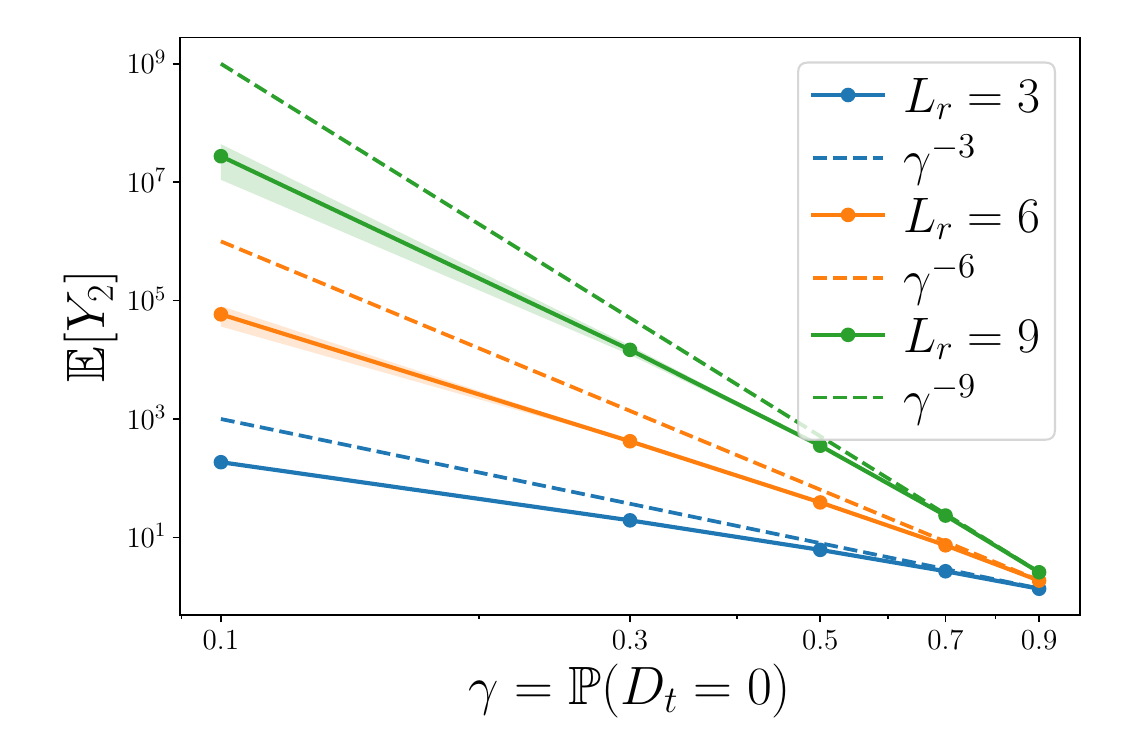}
        \caption{Empirical scaling of \(\Ex[Y_2]\).}
        \label{fig_2a}
    \end{subfigure}
    \hfill
    \begin{subfigure}[t]{0.32\textwidth}
        \centering
        \includegraphics[width=\linewidth]{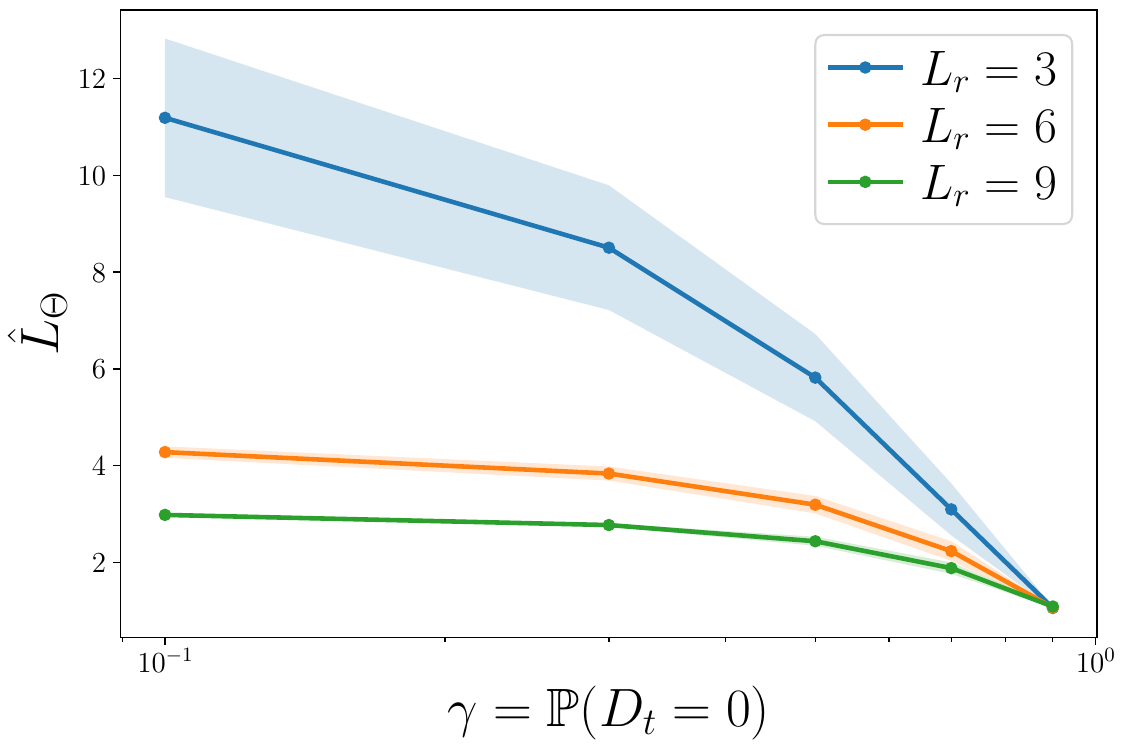}
        \caption{Empirical Lipschitz constant.}
        \label{fig_2b}
    \end{subfigure}
    \hfill
    \begin{subfigure}[t]{0.32\textwidth}
        \centering
        \includegraphics[width=\linewidth]{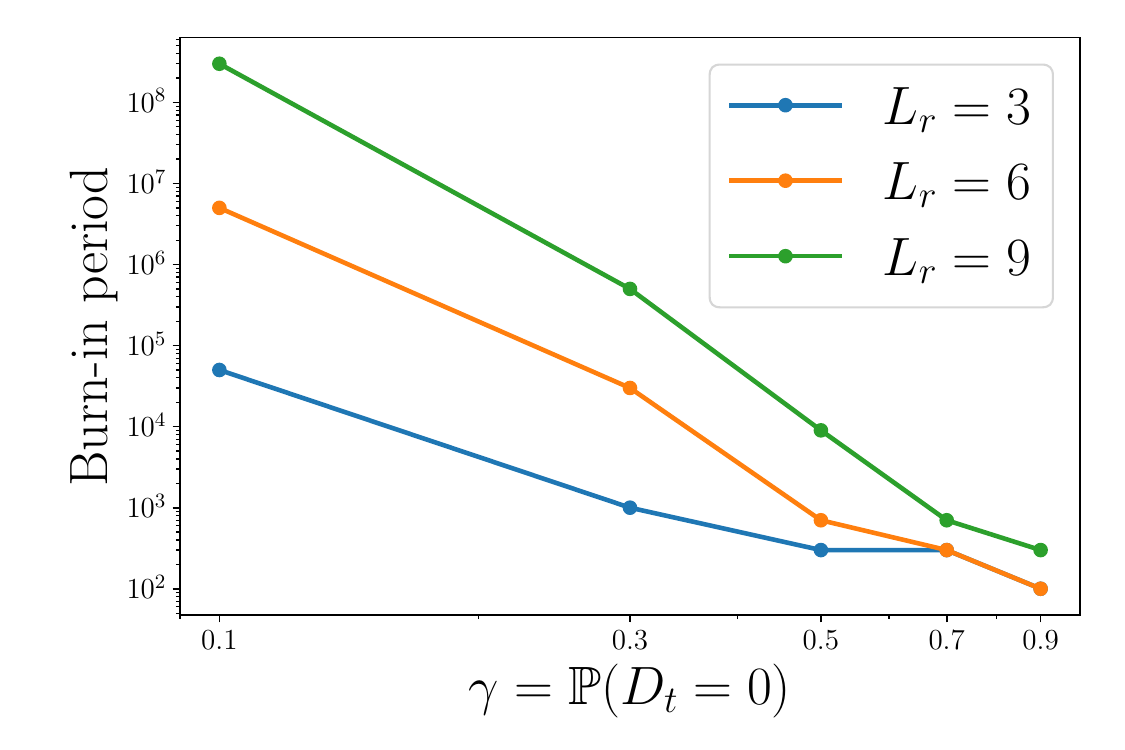}
        \caption{Empirical burn-in period.}
        \label{fig_2c}
    \end{subfigure}

    \vspace{0.5em}
    \caption{Numerical evidence on the applicability of~\ALG in the dual-index case study. \Cref{fig_2a} examines the effective quantity governing \(H\) and \(\alpha\), namely \(\Ex[Y_2]\), which can replace \(\gamma^{-L_r}\) in the regret bound in~\Cref{cor:dual_index_regret}. \Cref{fig_2b} reports the empirical Lipschitz constant \(\hat{L}_{\Theta}\), and \Cref{fig_2c} reports the end-to-end burn-in period. \Cref{fig_2a} and~\Cref{fig_2c} are plotted on log--log scales, while \Cref{fig_2b} uses a logarithmic scale only for the \(x\)-axis.
    }
    \label{fig_2}
\end{figure}

\paragraph{\(H\) and \(1/\alpha\) May Not Scale with \(\gamma^{-L_r}\).}
For the dual-index case study, we have \(\alpha = \frac{1-\gamma}{2}\gamma^{L_r}\) and \(H \le \frac{2\Cupperbound}{(1-\gamma)\gamma^{L_r}}\) (see~\Cref{lem:dual_index_partial_orders} and~\Cref{lem:dual_index_bias}). Thus, the current bounds on \(H\) and \(\alpha\) involve a \(\gamma^{-L_r}\)-type dependence, which can be large for realistic lead times \(L_r\) and moderate values of \(\gamma\).

We first note that such exponential dependence is common in the related literature; for example, see~\citet{zhang2020closing, huh2009adaptive, lyu2024ucb}. We discuss~\citet{zhang2020closing} in detail for comparison. \citet{zhang2020closing} study a lost-sales inventory model with a single retailer under censored demand, which is closely related to the setting we study in~\Cref{case_study:inventory}. In their Theorem~1, they show that the expected regret up to time \(T\) is bounded by \(K\sqrt{T}\), where \(K\) is proportional to \((1/c_1)^{2L}\). They assume that the support of demand \(D_t\) is lower bounded by some \(\underline D\), that \(\Pr(D_t=\underline D)=c_1\), and that \(L\) is the lead time. Consequently, the dependence of their constant \(K\) on \((1/c_1)^{2L}\) is analogous to the \(\gamma^{-3L_r/2}(1-\gamma)^{-3/2}\) dependence appearing in the regret bound in our~\Cref{cor:dual_index_regret}, although their exponent is even larger (their exponent coefficient is \(2\), whereas ours is \(3/2\)). Importantly, the source of these exponential-type constants is similar in the two works. Both analyses use such quantities to control the probability of a demand event, namely ``\(L_r\) consecutive zero demands'' in our work and ``\(L\) consecutive \(\underline D\) demands'' in~\citet{zhang2020closing}; see also~\Cref{lem_consecutive_zero}. This event is closely connected to regret in both analyses, in our work through the sample-reuse ratio \(\alpha\), and in~\citet{zhang2020closing} through the number of times the algorithm can execute its subgradient-based iteration.

Second, the \(\gamma^{-L_r}\)-type constant is not intrinsically necessary for our regret bound in~\Cref{cor:dual_index_regret}. Instead of analyzing the conservative event ``\(L_r\) consecutive zero demands'', whose probability scales as \(\gamma^{L_r}\), one can directly analyze the event that the on-hand inventory (after order arrivals) reaches $z_{r}^{\theta}$ under any policy $\theta$; see the proof of~\Cref{lem:recursive_inventory_bound} and~\Cref{dip_lem4}. Let $Y_2$ denote the first hitting time of this event, maximized over all states $s\in \S_\theta$. Then,
$H=O(\Ex[Y_2])$ and $\frac{1}{\alpha}=O(\Ex[Y_2])$, analogous to the relation between \(H\), \(\alpha\), and \(\gamma^{L_r}\) in the current proof. However, \(\Ex[Y_2]\) need not scale as \(\gamma^{-L_r}\). In~\Cref{fig_2a}, we numerically estimate how \(\Ex[Y_2]\) scales with \(\gamma\) and \(L_r\), using the \textsf{Optimal In-Class Policy} (see~\Cref{sec:simulations} for the definition) as a representative policy. The observed scale is significantly milder than \(\gamma^{-L_r}\). For example, when \(\gamma=0.1\) and \(L_r=6\), we have \(\gamma^{-L_r}\approx 10^{6}\), while the empirical value of \(\Ex[Y_2]\) lies between \(10^{4}\) and \(10^{5}\). 
Moreover, \Cref{fig_2a} suggests that the improvement is not merely in the multiplicative constant. If \(\Ex[Y_2]\) scaled as \(\gamma^{-L_r}\), then the log--log plot would be approximately linear with slope \(L_r\), as indicated by the dashed lines. Instead, all three curves are visibly nonlinear.
These numerical observations suggest that a sharper analysis on $\Ex[Y_2]$ could yield improved bounds on \(H\) and \(\alpha\) in~\Cref{cor:dual_index_regret}, avoiding the conservative \(\gamma^{-L_r}\)-type dependence. 

\paragraph{Lipschitzness May Not Require Large \(\gamma\).}
\Cref{fig_2b} provides empirical evidence of Lipschitz-like behavior across the tested values of \(\gamma\) and \(L_r\). Even for small values of \(\gamma\), such as \(\gamma = 0.1\), our simulations still give \(\hat{L}_{\Theta}\le 20\) (the definition of \(\hat{L}_{\Theta}\) is provided below in the simulation setup).

Interestingly, \Cref{fig_2b} suggests that the estimated Lipschitz constant decreases as \(L_r\) increases. One possible conjecture is that, for larger \(L_r\), the regular channel has a weaker immediate effect on realized costs, making the system closer, from the perspective of policy sensitivity, to a single-sourcing problem driven primarily by the short-lead channel. The single-sourcing model is indeed Lipschitz continuous~\citep{agrawal2019learning}. We leave a formal investigation of this phenomenon open. Nevertheless, \Cref{fig_2b} is sufficient for our purpose, i.e., in the simulation, the empirical Lipschitz constants remain mild, staying below \(20\).

\paragraph{The Burn-In Period Can Be Practical}
Based on the observations above, one may expect~\ALG to exhibit \(\sqrt{T}\)-type behavior on regret in the dual-index case study beyond the parameter regime covered by our current proof, and with a much milder effective dependence on \(\gamma\) and \(L_r\) than the conservative \(\gamma^{-L_r}\)-type bound suggests. In~\Cref{fig_2c}, we provide numerical evidence for this point by estimating the burn-in period. 
Concretely, for each parameter setting, we repeat the experiment over \(20\) independent runs and define the burn-in period as the smallest training horizon \(T\) at which the learned policy achieves lower long-run cost than the \textsf{Random} policy (see~\Cref{sec:simulations}) in more than \(80\%\) of the runs.

The numerical results in \Cref{fig_2c} suggest that our model has practical relevance beyond the conservative worst-case bound. For example, many dual-sourcing problems in reality involve a trade-off between an international sourcing channel, with a lead time of several months, and a domestic sourcing channel, with a lead time of one or two months~\citep{allon2010global}. Thus, setting \(L_r = 6\) can be viewed as a representative case when one period corresponds to one month and international sourcing takes six months. The choice \(\gamma = 0.3\) is also plausible for intermittent or slow-moving demand; as noted in the main text, slow-moving items account for a large fraction of modern supply chains. Under this setting, \Cref{fig_2c} shows that the observed burn-in period is roughly \(10^5\).
Unlike a hopelessly large burn-in period, such as one on the order of \(10^9\) or larger, a burn-in period of \(10^5\) remains practically meaningful. In real inventory-learning problems, effective samples need not come from a single isolated product trajectory; they can often be accumulated across similar products, stores, and regions. We therefore view this burn-in period as large, but still realistic for learning in this setting.

Finally, the shape of the curves in \Cref{fig_2c} largely mirrors that in \Cref{fig_2a}. This is expected; the two figures use largely the same simulation settings, while \Cref{fig_2a} measures the time required to obtain one ``effective sample'' for counterfactual estimation and \Cref{fig_2c} reflects the time required to collect enough such effective samples for learning.

\paragraph{Simulation Setup}
Lastly, we provide the details of the simulation setup for~\Cref{fig_2}. The three simulations in \Cref{fig_2a}, \Cref{fig_2b}, and \Cref{fig_2c} share the same model hyperparameters unless otherwise specified below. Following~\Cref{sec:simulations}, we set the holding cost \(h=1\), the shortage penalty \(p=10\), and the purchasing costs \((c_r,c_e)=(0,0.5)\). The demand support is \(D_t\in[0,40]\). To focus on the effects of \(\gamma=\Pr(D_t=0)\) and \(L_r\), we use a binary demand distribution $\Pr(D_t=0)=\gamma, \Pr(D_t=40)=1-\gamma$.
We set \(L_e=2\), noting that \(L_e\) does not appear in the regret bound in~\Cref{cor:dual_index_regret}. Across the three simulations, we consider $L_r\in\{3,6,9\}$ and $\gamma\in\{0.1,0.3,0.5,0.7,0.9\}$.

For \Cref{fig_2a}, we estimate \(Y_2\), defined as the maximum, over all initial states, of the first hitting time for the on-hand inventory (after order arrivals) to reach \(z_r^\theta\). For each setting \((L_r,\gamma)\), we simulate trajectories of length \(10^9\) under the \textsf{Optimal In-Class Policy}, as in~\Cref{sec:simulations}.
The reported value of \(\Ex[Y_2]\) is the empirical average over \(20\) independent runs. The dashed lines correspond to the conservative benchmark \(\gamma^{-L_r}\), which is linear on the log--log scale. 

For \Cref{fig_2b}, the empirical Lipschitz constant is computed as
\(
    \hat{L}_\Theta = \max_{\theta_1,\theta_2\in\Theta_1}\frac{\left|G_{\theta_1}(\H_{\theta_1}^T) - G_{\theta_2}(\H_{\theta_2}^T)\right|}{\|\theta_1-\theta_2\|_\infty},
\)
where \(G_{\theta_i}(\H_{\theta_i}^T)\) is the empirical average cost under policy \(\theta_i\), computed over a trajectory of length \(T\). We take \(T=10^5\) and use the grid
\(
    \Theta_1=\{0,5,10,\ldots,400\}\times\{0,5,10,\ldots,400\}.
\)
The upper bound is selected as \(400\) because \(D_t\in[0,40]\) and \(L_r\le 9\), so it is large enough to contain the relevant ordering quantities. Each independent run produces one value of \(\hat L_\Theta\); the reported curve and error bars in \Cref{fig_2b} are computed over \(20\) independent runs.

For \Cref{fig_2c}, we use the same simulation environment as in~\Cref{sec:simulations}. Apart from the components specified above, the remaining details of the simulation setup, including the discretization radius used to construct \(\Theta_1\), follow~\Cref{sec:simulations}.
To estimate the burn-in period, we consider the training budgets
\[
    \{a\times 10^k: a\in\{1,3,5,7,9\},\ k\in\{2,\ldots,8\}\}.
\]
As introduced above, the burn-in period is the smallest training budget in this set for which the learned policy achieves lower (empirical) long-run cost than \textsf{Random} in more than \(80\%\) of the \(20\) independent runs.

The error bars in \Cref{fig_2a} and \Cref{fig_2b} correspond to \(95\%\) confidence intervals. \Cref{fig_2c} does not include error bars, since the burn-in period is defined through an \(80\%\) success threshold across independent runs.

\begin{table}[h]
\centering
\caption{Queuing Performance under Decaying vs.\ Fixed Arrival Rates}
\label{tab:queuing-comparison}
\begin{tabular}{lcc}
\toprule
{\textbf Algorithm} & \multicolumn{2}{c}{Queuing} \\
\cmidrule(lr){2-3}
          & Decaying Arrival Rate & Fixed Arrival Rate \\
\midrule
\textsf{Optimal} $(g_{\theta^*})$      & $9.5$            & $11.6$            \\
\textsf{ERM} \citep{sinclair2023hindsight}               & $9.6(1\%)$     & $11.6(0\%)$     \\
\ALG                        & $^\star10.5(11\%)$ & $^\star13.1(13\%)$ \\
\textsf{PPO}~\citep{schulman2017proximal}              & $^{\star}9.8(3\%)$  & $^\star11.8(2\%)$  \\
\textsf{Feedback Graph}~\citep{dann2020reinforcement}     & $11.5(21\%)$   & $13.2(13\%)$               \\
\textsf{Problem-Specific}              & $11.3(19\%)$   & $13.9(20\%)$               \\
\textsf{Random}                         & $11.1(17\%)$   & $13.5(16\%)$               \\
\bottomrule
\end{tabular}

\end{table}

\subsubsection{Queuing Model with Fixed Arrival Rate}
\label{appendix_c2}

In the queuing case study of \Cref{case_study:queuing,sec:simulations}, we adopted the decaying arrival‐rate setting from \citet{anselmi2022reinforcement} for both theoretical analysis and numerical experiments. In this appendix, we instead examine the classic M/M/1/L model with a fixed (constant) arrival rate, $\lambda$, and present the corresponding simulation results for various algorithms, including \textsf{IOPEA}. These results are summarized in \Cref{tab:queuing-comparison}. Aside from replacing the time‐varying rates $\{\lambda_s\}$ of \Cref{case_study:queuing} with the constant rate $\lambda$ (see \Cref{tab:configs-by-block}), the experimental setup is identical to that of \cref{sec:simulations}. We observe that nearly all methods exhibit performance comparable to the decaying‐rate case, and thus the main conclusions of \cref{sec:simulations} remain unchanged under the fixed‐rate setting.

We also observe that in both the decaying-rate and fixed-rate scenarios, the {\textsf Problem-Specific} algorithm ({\textsf UCRL2}) underperforms the \textsf{Random} baseline.  This is because the problem is relatively small (so \textsf{Random}  performs reasonably) and the {\textsf UCRL2} algorithm after $T = 3 \times 10^5$ is not long enough for the algorithm to approximate the optimal policy.  We limit our experiments to this scale because {\textsf UCRL2} requires comparing multiple MDP models at each epoch, which becomes prohibitively expensive in larger scale problems.
\end{APPENDICES}

% Acknowledgments here
% \ACKNOWLEDGMENT{We would like to express our sincere gratitude to Connor Lawless for their technical assistance in the numerical simulations.}
%contributions to this research. We are also grateful to [mention any additional acknowledgements, such as technical assistance, data providers, or colleagues] for their support and assistance throughout the course of this work.}

% References here (outcomment the appropriate case)

% CASE 1: BiBTeX used to constantly update the references
%   (while the paper is being written).
%\bibliographystyle{informs2014} % outcomment this and next line in Case 1
%\bibliography{<your bib file(s)>} % if more than one, comma separated

% CASE 2: BiBTeX used to generate mypaper.bbl (to be further fine tuned)
%\input{mypaper.bbl} % outcomment this line in Case 2

%If you don't use BiBTex, you can manually itemize references as shown below.

%\bibliographystyle{nonumber}

%%%%%%%%%%%%%%%%%
\end{document}
%%%%%%%%%%%%%%%%%